\documentclass[mnsc,nonblindrev]{informs3a}

\OneAndAHalfSpacedXI % 11-point type with 1.5 line spacing for the public version

\RequirePackage{tgtermes}
\RequirePackage{newtxtext}
\RequirePackage{newtxmath}
\RequirePackage{bm}
\RequirePackage{endnotes}

\usepackage{makecell}
\usepackage{natbib, multirow, multicol, xspace, enumitem, pifont, algorithm, algpseudocode, subcaption}
\usepackage{accents}
\usepackage{placeins}
\usepackage{mdframed}
\bibpunct[, ]{(}{)}{,}{a}{}{,}%
\def\bibfont{\small}%
\usepackage[dvipsnames]{xcolor}
\usepackage[all]{xy}
\usepackage{hyperref}
\hypersetup{
    pdftitle={AdaSwitch: An Adaptive Switching Meta-Algorithm for Learning-Augmented Bounded-Influence Problems},
    pdfauthor={Xi Chen, Yuze Chen, Shibo Dai, Yuan Zhou},
    colorlinks=true,
    linkcolor=blue,
    filecolor=magenta,      
    urlcolor=blue,
    citecolor=blue
    }
\usepackage{framed}

\newcommand{\defeq}{\overset{\text{\tiny def}}{=}}
\theoremstyle{TH}%
\newtheorem{observation}{Observation}

\TheoremsNumberedThrough     % Preferred (Theorem 1, Lemma 1, Theorem 2)
\ECRepeatTheorems

\EquationsNumberedThrough    % Default: (1), (2), ...
\MANUSCRIPTNO{}

\begin{document}
%%%%%%%%%%%%%%%%

\RUNAUTHOR{Chen, Chen, Dai, and Zhou}
% The electronic companion uses the same public author running head.
\ECRUNAUTHOR{Chen, Chen, Dai, and Zhou}

\RUNTITLE{Adaptive Switching for Learning-Augmented Bounded-Influence Problems}

\TITLE{AdaSwitch: An Adaptive Switching Meta-Algorithm for Learning-Augmented Bounded-Influence Problems}

\ARTICLEAUTHORS{
 \AUTHOR{Xi Chen}
 \AFF{Leonard N. Stern School of Business, New York University, New York, NY 10012, USA, \EMAIL{xc13@stern.nyu.edu}}
 \AUTHOR{Yuze Chen}
 \AFF{Qiuzhen College, Tsinghua University, Beijing 100084, China, \EMAIL{yz-chen21@mails.tsinghua.edu.cn}}
 \AUTHOR{Shibo Dai}
 \AFF{Qiuzhen College, Tsinghua University, Beijing 100084, China, \EMAIL{dsb24@mails.tsinghua.edu.cn}}
 \AUTHOR{Yuan Zhou}
 \AFF{Yau Mathematical Sciences Center \& Department of Mathematical Sciences, Tsinghua University, Beijing 100084, China, \\ Beijing Institute of Mathematical Sciences and Applications, Beijing 101408, China, \EMAIL{yuan-zhou@tsinghua.edu.cn}}
}

\ABSTRACT{We study history-dependent online problems with a possibly inaccurate  prediction of the future request sequence. Motivated by several real-world applications, we introduce a \emph{bounded-influence} framework in which past decisions and requests affect the future optimal value by only a bounded amount. Within this framework, we develop AdaSwitch, a meta-algorithm that adaptively switches between suitable offline and online oracles. AdaSwitch provides explicit guarantees on expected performance that tighten as prediction error decreases or the offline optimum increases. With perfect predictions, its guarantee approaches the offline oracle's guarantee as the offline optimum grows. It also retains a worst-case guarantee close to that of the online oracle under arbitrary predictions. Applications to online lead-time quotation, $k$-server and caching, and online reusable resource allocation demonstrate the framework's applicability to both reward maximization and cost minimization.}


\maketitle

\section{Introduction}

Learning-augmented algorithms combine online decision rules with predictions or advice to improve performance while maintaining guarantees when predictions are inaccurate~\citep{purohit2018improving}. The challenge is to make effective use of predictive information whose reliability is not known in advance. This approach has found applications in staffing~\citep{feng2025robust}, scheduling~\citep{lattanzi2020online,li2021online}, energy systems~\citep{lechowicz2023online}, facility location~\citep{azar2022online}, and online allocation~\citep{mahdian2007allocating}, where predictions can guide decisions despite uncertainty about future demand.

In this paper, we study learning-augmented algorithms for multi-period online decision-making problems. Before the process begins, the decision maker receives a prediction of the future request sequence. This prediction remains fixed and may be inaccurate. In each period, after observing the realized request, the decision maker takes an irrevocable action that yields a reward or incurs a cost and may affect future opportunities. The objective is to maximize total reward or minimize total cost. We evaluate performance relative to the \emph{offline optimum}: the maximum total reward or minimum total cost achievable with full knowledge of future requests. The central challenge is to balance \emph{consistency}, the performance guarantee when predictions are perfect (identical to the actual requests), and \emph{robustness}, the guarantee that holds for arbitrary predictions. We seek to benefit from accurate predictions while retaining robustness comparable to that of classical online algorithms that do not use predictions.

One strand of learning-augmented research adapts classical online algorithms by incorporating predictions into problem-specific thresholds and decision rules~\citep{purohit2018improving,lykouris2021competitive}. We identify a structural property that supports a common approach across several history-dependent problems: changing past requests and actions can alter the best achievable future reward or cost by only a bounded amount. For example, earlier bookings occupy hotel rooms for a bounded time, while cache contents can be changed with a bounded number of page loads, consequently, the total influence of past requests and actions remains bounded. In Section~\ref{sec:framework-formulation}, we formalize this property as \emph{bounded influence}, together with conditions that control how discrepancies in requests affect rewards or costs. These properties provide a basis for switching between algorithms while accounting for the influence of earlier decisions and inaccurate predictions.

Within this framework, we propose the Adaptive Switching (AdaSwitch) meta-algorithm, a general procedure that leverages suitable existing offline and online algorithms. AdaSwitch maintains robustness close to the online algorithm's guarantee and explicitly quantifies how prediction accuracy can improve its guarantee. The same switching principle applies to reward maximization and cost minimization, using either exact or approximate offline algorithms. We summarize the algorithmic results and their applications below.

\subsection{Our Contributions}

\noindent{\bf The Adaptive Switching Meta-Algorithm.}
In Section~\ref{sec:adaswitch-framework}, we introduce the Adaptive Switching (AdaSwitch) meta-algorithm for bounded-influence problems with predictions. The meta-algorithm relies on two oracles: a \emph{$\gamma$-offline oracle} that outputs an action sequence achieving $\gamma$-approximation of the optimal solution for a supplied request sequence; and an \emph{$\eta$-online oracle} that guarantees an $\eta$-competitive ratio without prior knowledge of the request sequence. AdaSwitch provides explicit performance guarantees in terms of prediction error and the offline optimum, which improves as prediction error decreases or the offline optimum grows. Under perfect predictions, the bound can be arbitrarily close to \(\gamma\) once the offline optimum is sufficiently large, achieving near-$\gamma$ consistency. At the same time, AdaSwitch retains worst-case guarantee close to $\eta$ for arbitrary predictions, achieving near-$\eta$ robustness.

\begin{figure}[htbp]
    \centering
    \includegraphics[width=0.95\linewidth]{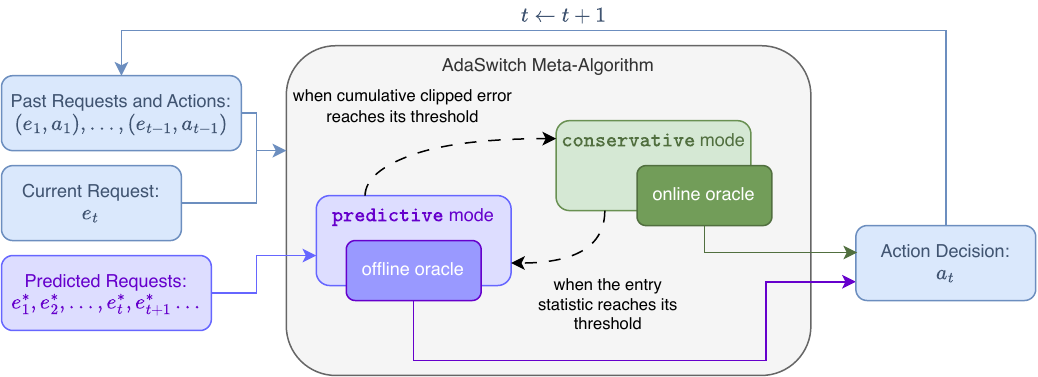}
    \caption{Information flow and mode transitions in AdaSwitch.}
    \label{fig:adaswitch}
\end{figure}

As illustrated in Figure~\ref{fig:adaswitch}, AdaSwitch adaptively alternates between conservative and predictive modes. In conservative mode, it follows the online oracle, relying on its guarantee to protect against inaccurate predictions. Once a switching condition based on the requests observed so far is met, AdaSwitch enters predictive mode, where it uses the offline oracle to construct actions from the current request and predicted future requests. During predictive operation, the algorithm continuously monitors cumulative prediction error; if this error reaches a specified threshold, it returns to conservative mode. This process allows AdaSwitch to dynamically balance between robustness and benefiting from predictions. The key insight is that, due to the bounded-influence property of the underlying problem, the losses incurred during state transitions can be effectively controlled. By carefully designing the switching conditions, we establish performance guarantees that explicitly depend on the quality of the predicted request sequence. 

\medskip
\noindent{\bf Application I: Online Lead-Time Quotation with Predictions (OLTQwP).}
To illustrate the power of our bounded-influence framework and the AdaSwitch meta-algorithm, in Section~\ref{sec:lead-time-quotation}, we investigate online lead-time quotation (OLTQ) and its version with predictions (OLTQwP). In the OLTQ model, multiple requests may arrive in each period, but the processor can serve only one. Each arriving request must immediately and irrevocably be assigned to a processing period or rejected. The time between arrival and processing is the quoted lead time; shorter lead times earn larger rewards, and rewards become zero once the lead time reaches or exceeds a fixed patience limit. The objective is to maximize total reward. This model captures quotation decisions in customized manufacturing and time-sensitive services, where limited capacity and uncertain demand make these commitments consequential~\citep{keskinocak2001scheduling}. The central challenge is to balance service speed against revenue:  accepting more requests reserves capacity that later requests may use more profitably, while delaying service reduces the reward earned. 

We instantiate AdaSwitch with an exact offline oracle and an online oracle, both adapted from~\cite{keskinocak2001scheduling}, and compare its guarantees with those of the learning-augmented algorithm Q-FRACwP proposed by \citet{huo2026online}. Q-FRACwP provides fixed consistency and robustness guarantees; however, these guarantees do not distinguish between different levels of prediction quality. By contrast, AdaSwitch provides a competitive-ratio bound that explicitly depends on the prediction error and tightens as the error decreases. Moreover, the bound of AdaSwitch is instance-dependent: under matched robustness guarantees, AdaSwitch can achieve near-perfect consistency as the offline optimal reward grows, yielding a strictly better consistency guarantee than Q-FRACwP.

\medskip
\noindent{\bf Application II: The $k$-Server Problem with Predictions ($k$SEwP).}
In Section~\ref{sec:k-server}, we demonstrate the second application of our AdaSwitch meta-algorithm: the $k$SEwP problem. In this problem, a sequence of requests arrives sequentially, and the decision maker must select one of the $k$ servers to serve the request by moving it to the request's location, which incurs a traveling cost. The objective is to minimize the total cost. In inventory and logistics management, the $k$-server problem naturally models situations where a limited number of mobile agents must respond to sequential requests dispersed over a metric space. A classical example of the $k$-server problem is caching, where servers correspond to cache slots and requests correspond to data items. Serving a request may require replacing a cached item, which incurs a cost analogous to moving a server.

Our bounded-influence framework and AdaSwitch also apply to cost minimization, including $k$SEwP and its special case, caching with predictions (CAwP), on bounded metric spaces. A closely related work is~\cite{lykouris2021competitive}, which studies learning-augmented caching with a next-arrival-time predictor. Although the different prediction forms prevent direct comparison, AdaSwitch achieves near-perfect consistency, whereas their approach still has a competitive ratio of at least 2 even under perfect predictions. The Blum--Burch combiner~\citep{wei2020better,antoniadis2023online} and DART~\citep{christianson2023optimal} also switch between policies with and without predictions, using switching mechanisms that differ from AdaSwitch's. For a fair comparison, we evaluate AdaSwitch and these combiners under a common caching setup. AdaSwitch offers two clear advantages. First, it matches Blum--Burch's leading robustness factor without an additive term and achieves a substantially smaller robustness factor than DART. Second, its consistency guarantee approaches 1 as the offline optimum grows, yielding a tighter guarantee than both combiners.

\medskip
\noindent{\bf Application III: Online Reusable Resource Allocation with Predictions (ORRAwP).}
In Section~\ref{sec:ORRA}, we study online reusable resource allocation (ORRA) with predictions, building on the online bipartite matching model of \citet{delong2024online}. In ORRA, each arriving request specifies a set of eligible resources, and the decision maker either assigns an available resource from that set or rejects the request. Each resource can serve at most one request at a time. Once assigned, it remains occupied for $D-1$ subsequent periods before becoming available again, with the same duration $D$ for all assignments. The objective is to maximize the number of served requests. A motivating example is a cloud platform offering fixed-duration computing-resource reservations, where allocating a compatible resource now may prevent a later job from being served~\citep{babaioff2017era}.

For a fixed reuse duration $D$, we construct a polynomial-time exact offline oracle and integrate it into AdaSwitch, achieving near-perfect consistency. For general $D$, however, obtaining an efficient exact oracle remains an open challenge, making a more efficient $\gamma$-approximate offline oracle particularly appealing. Given such an oracle, AdaSwitch can also incorporate it to achieve near-$\gamma$ consistency. And in both cases, AdaSwitch preserves the competitive guarantees of~\cite{delong2024online} under arbitrary predictions, showing that its robustness does not rely on an exact oracle. This demonstrates AdaSwitch's flexibility in using either exact or approximate offline oracles. Such flexibility is particularly valuable because reusable resources are central to many practical systems. Moreover, to the best of our knowledge, this is the first study of learning-augmented algorithms for this model using predictions of the future request sequence.

\medskip
\noindent{\bf Experiments.}
Synthetic experiments on OLTQ and caching compare AdaSwitch against classical online algorithms and algorithms that use predictions. In OLTQ, under perfect predictions, AdaSwitch earns a larger fraction of the optimal reward than Q-FRACwP on average over most shared robustness targets. In caching, AdaSwitch likewise incurs lower average cost relative to the offline optimum than Marking and Predictive Marker~\citep{lykouris2021competitive} throughout the tested robustness sweep.

\section{Related Work}\label{sec:related-works}

\noindent{\bf Learning-Augmented Problems.}
Learning-augmented research spans online assignment~\citep{vee2010optimal}, online matching~\citep{li2023learning}, load balancing and resource allocation~\citep{mahdian2012online}, revenue management~\citep{balseiro2026single}, and concave packing and convex covering~\citep{grigorescu2025learning}. One strand adapts the decision rules of classical online algorithms to incorporate predictions. For example, \citet{purohit2018improving} make the ski-rental purchase threshold depend on the prediction, while \citet{lykouris2021competitive} modify the Marking caching algorithm through prediction-guided evictions with a randomized fallback. 

Other approaches combine online algorithms with plans computed from predictions in different ways. \citet{azar2022online} and \citet{jabalAmeli2025covering} run online algorithms to serve arriving requests and use predictions to purchase additional resources. \citet{balkanski2023energy} initially schedule online, then assign correctly predicted future jobs to an offline schedule and the remaining work to an online algorithm. AdaSwitch instead takes each action from either a plan computed using predictions or its online oracle.

Switching approaches differ in their prediction inputs and whether they allow repeated switching. \citet{rutten2023smoothed} use predictions of optimal actions and allow repeated switching in smoothed online optimization. The interval-scheduling frameworks of \citet{antoniadis2026switching} receive recommendations to accept or reject each interval. Once they switch to an algorithm that does not use predictions, they do not return to using predictions. AdaSwitch uses a fixed request-sequence prediction and permits repeated switching. 

A further distinction concerns how policy states are handled when switching. Blum--Burch~\citep{wei2020better,antoniadis2023online} and DART~\citep{christianson2023optimal} track the costs of separately simulated candidate policies and use a metric on their states to account for additional cost of switching. AdaSwitch constructs plans from predicted requests and restarts its online oracle from the realized history. Its analysis uses bounded influence to control differences in optimal continuation value across histories, without requiring a metric on policy states.

\medskip
\noindent{\bf Lead-Time Quotation.}
Lead time is an important performance metric in operations and supply chain management. Prior work examines how lead-time variability affects supply-chain performance~\citep{kim2006quantifying,heydari2009study}. A related stream treats lead time as a decision, studying lead-time reduction~\citep{hill1992models}, quotation policies~\citep{duenyas1995quoting}, and interactions between pricing and lead-time decisions~\citep{liu2007pricing,ray2004customer}. 

In the online setting, \citet{keskinocak2001scheduling} study quotation decisions made when requests arrive. \citet{huo2026online} incorporate demand predictions through dynamic reward thresholds and analyze the trade-off between consistency and robustness. We complement their study with guarantees that depend explicitly on prediction error and the offline optimum. For fixed problem parameters and robustness target, our bound under perfect predictions approaches $1$ as the offline optimum grows.

\medskip
\noindent{\bf The $k$-Server Problem and Caching.}
In the $k$-server problem, a server must be at the requested location to serve a request, and the objective is to minimize total movement cost~\citep{manasse1988competitive}. Unit-cost caching is its uniform-metric special case. Classical algorithms include the work function algorithm (WFA)~\citep{koutsoupias1995k} for $k$-server and randomized Marking~\citep{fiat1991competitive} for caching.

In the learning-augmented setting, \citet{antoniadis2023mixing} develop algorithms that combine multiple predictors for the more general \emph{metrical task systems (MTS)} problem, which includes $k$-server and caching. For MTS, Blum--Burch~\citep{antoniadis2023online} and DART~\citep{christianson2023optimal} combine a prediction-informed policy with a robust baseline. For caching specifically, Predictive Marker and subsequent caching algorithms commonly use predictions of when a requested page will next appear and provide guarantees that depend on prediction error~\citep{lykouris2021competitive,rohatgi2020near,wei2020better}. \cite{bansal2022learning} study weighted caching with this type of prediction, while~\cite{im2022parsimonious} develop algorithms that reduce the number of prediction queries. Request-sequence predictions have also been studied in the related page-migration problem~\citep{indyk2022online}, which permits remote service at a distance-dependent cost, whereas $k$-server requires a server at the requested location.

We apply AdaSwitch to $k$-server on bounded metric spaces and caching using a fixed request-sequence prediction supplied in advance and online oracles constructed from WFA and Marking, respectively.  Sections~\ref{sec:caching} and~\ref{sec:ec-related-comparison} compare the theoretical guarantees of AdaSwitch, Blum--Burch, and DART under a common caching setup. Under matched parameter choices, AdaSwitch matches the leading robustness factor of Blum--Burch without an additive term and achieves a substantially smaller leading factor than DART; its consistency guarantee approaches $1$ as the offline optimum grows, eventually becoming tighter than both combiners.

\medskip
\noindent{\bf Reusable Resource Allocation.}
Online reusable resource allocation assigns available resources to arriving requests, with each assignment temporarily preventing that resource from serving another request. Applications include ride-sharing~\citep{dickerson2021allocation}, cloud job reservations~\citep{babaioff2017era}, and healthcare appointment scheduling~\citep{zhao2022dynamic}. Reusable resources have also been studied in assortment optimization~\citep{gong2022online} and revenue management~\citep{levi2010provably}.

Learning-augmented resource allocation has been studied for non-reusable resources using predicted capacity shadow prices~\citep{an2024best} or convex sets describing possible total demand~\citep{golrezaei2026online}. For reusable resources, we study a unit-reward version of the matching model of \citet{delong2024online} with request-sequence predictions (Section~\ref{sec:ORRA}). We adapt their Periodic Reranking algorithm to obtain an online oracle that can be used in AdaSwitch.

\section{The Bounded-Influence Framework} \label{sec:framework-formulation}
In period $t$, a request $e_t\in E_t$ arrives, the decision maker chooses $a_t\in A_t$, and earns a reward that may depend on the entire realized history. Let $\mathcal P=\{\{E_t\}_{t\geq1},\{A_t\}_{t\geq1},\{R_t\}_{t\geq1}\}$ denote this multi-period reward-maximization problem, with request spaces $E_t$, action spaces $A_t$, and reward functions $R_t$ fixed and known before period~$1$. The cost-minimization interpretation uses the same primitives and is specified in Section~\ref{sec:adaswitch-cost-minimization}.

Action spaces are common to all histories. OLTQ and ORRA encode capacity and eligibility constraints by interpreting conflicting, unavailable, or ineligible assignments as unsuccessful attempts. This lets the same action sequence be evaluated under different histories and requests, as required by the comparisons below.

Write $\bm e_{i:j}$ and $\bm a_{i:j}$ for inclusive request and action slices and $\bm E_{i:j}$ and $\bm A_{i:j}$ for their product spaces, and let $\circ$ denote sequence concatenation. Throughout, $\mathbb Z_+=\{1,2,\ldots\}$; for $q\in\mathbb Z_{\geq0}$, write $[q]\defeq\{1,\ldots,q\}$, with $[0]\defeq\varnothing$; set $[x]_+\defeq\max\{x,0\}$; and let $\mathbb I(\cdot)$ denote the indicator of an event. Empty slices and products have their usual conventions.

\noindent\textbf{Regularity assumptions.}
Fix $0<L<\infty$ and assume that each reward is deterministic and measurable:
\begin{align*}
R_t:\bm E_{1:t}\times\bm A_{1:t}\longrightarrow[0,L].
\end{align*}
\phantomsection\label{para:request-prediction-information}
Policies are nonanticipating and may be randomized. Each admissible realized-sequence/prediction pair $(\bm e_{1:\infty},\bm e^*_{1:\infty})$ is fixed before algorithmic randomization. The prediction is revealed before period~$1$, whereas realized requests are revealed sequentially. All theoretical guarantees are pointwise in this pair; expectations are over algorithmic randomness only. Every displayed optimization is assumed to have a nonempty feasible set, an attained optimum, and a measurable value map; optimizers follow a fixed measurable selection rule. Section~\ref{sec:ec-framework-conventions} gives the measure-theoretic formulation.

A finite effective horizon is the last period in which some action sequence can earn a nonzero reward, or zero if none can. It may extend beyond the last request: in OLTQ, an earlier arrival can earn reward when processed later.
\begin{definition}
\label{def:admissible-request-sequence}
With $\min\varnothing\defeq+\infty$, the \emph{effective horizon} of a sequence $\bm e_{1:\infty}\in\bm E_{1:\infty}$ is
\[
M(\bm e_{1:\infty})
\defeq
\min\left\{
m\in\mathbb Z_{\geq0}:
R_t(\bm e_{1:t},\bm a_{1:t})=0
\text{ for every }t\geq m+1
\text{ and every }\bm a_{1:t}\in\bm A_{1:t}
\right\}.
\]
The sequence is \emph{admissible} if, for every $n\in\mathbb Z_{\geq0}$ and every replacement prefix $\widehat{\bm e}_{1:n}\in\bm E_{1:n}$, we have
$
M(\widehat{\bm e}_{1:n}\circ\bm e_{n+1:\infty})<\infty.
$
\end{definition}
Unless stated otherwise, all realized and predicted request sequences are assumed to be admissible. All finite prefixes and infinite continuations considered are taken from admissible request sequences.

For $n\in\mathbb Z_{\geq0}\cup\{\infty\}$, define
\begin{align*}
\mathrm{Val}(\mathcal P,\bm e_{1:n},\bm a_{1:n})
\defeq\sum_{t=1}^{n}R_t(\bm e_{1:t},\bm a_{1:t})
\quad \mathrm{and} \quad
\mathrm{Val}(\mathcal P,\bm e_{1:n},\bm\pi)
\defeq\mathbb E_{\bm\pi}\!\left[
\mathrm{Val}(\mathcal P,\bm e_{1:n},\bm a_{1:n})
\right].
\end{align*}
The sums have finitely many nonzero terms. Define
\begin{align*}
\mathrm{Opt}(\mathcal P,\bm e_{1:n})
\defeq
\max_{\bm a_{1:n}\in\bm A_{1:n}}
\mathrm{Val}(\mathcal P,\bm e_{1:n},\bm a_{1:n}).
\end{align*}

For a fixed history $\mathcal I=(\bm e_{1:h},\bm a_{1:h})$ of length $h\in\mathbb Z_{\geq0}$, define its continuation problem $\mathcal P^{\mathcal I}$, whose rewards, indexed from the start of the continuation, are
\begin{align*}
R_r^{\mathcal I}(\bm e_{h+1:h+r},\bm a_{h+1:h+r})
\defeq
R_{h+r}(\bm e_{1:h}\circ\bm e_{h+1:h+r},
\bm a_{1:h}\circ\bm a_{h+1:h+r}),
\end{align*}
and
$\mathcal P^{\mathcal I}
\defeq\{\{E_{h+r}\}_{r\geq1},\{A_{h+r}\}_{r\geq1},
\{R_r^{\mathcal I}\}_{r\geq1}\}$.
Continuation requests and actions retain their calendar indices; $R_r^{\mathcal I}$ uses the local index $r=t-h$. For $N\in\{h,h+1,\ldots\}\cup\{\infty\}$,
\begin{align*}
\mathrm{Val}(\mathcal P^{\mathcal I},\bm e_{h+1:N},\bm a_{h+1:N})
\defeq
\sum_{t=h+1}^{N}R_{t-h}^{\mathcal I}(\bm e_{h+1:t},\bm a_{h+1:t}).
\end{align*}
We define policy values by taking expectations and define $\mathrm{Opt}$ by maximizing action-sequence values over $\bm A_{h+1:N}$. When $N=h$, the continuation is empty, so its action-sequence value, policy value, and optimum are all zero.

\medskip
\noindent\textbf{Bounded-Influence and Lipschitz Assumptions.}
Bounded influence bounds differences in optimal continuation values between histories, whereas the Lipschitz conditions below control losses caused by request perturbations.
\begin{definition}[$f$-bounded-influence]
\label{def:bounded-influence}
For $f\in\mathbb R_{\geq0}$, a problem $\mathcal P$ is \emph{$f$-bounded-influence} if, for every $m\in\mathbb Z_+$ and $n\in\mathbb Z_+\cup\{\infty\}$ with $m\leq n$, histories
$\mathcal I=(\bm e_{1:m-1},\bm a_{1:m-1})$ and
$\mathcal I'=(\bm e'_{1:m-1},\bm a'_{1:m-1})$, and common continuation $\bm e_{m:n}\in\bm E_{m:n}$, we have
$
\left|
\mathrm{Opt}(\mathcal P^{\mathcal I},\bm e_{m:n})
-\mathrm{Opt}(\mathcal P^{\mathcal I'},\bm e_{m:n})
\right|\leq fL.
$
\end{definition}

A sufficient condition is that, for some uniform $r_{\mathrm{mem}}\in\mathbb Z_{\geq0}$, under every common request continuation and fixed continuation action sequence, changing the preceding history can alter only rewards in the first $r_{\mathrm{mem}}$ periods. Fixed-action continuation values then differ by at most $r_{\mathrm{mem}}L$, and optimizing shows that the problem is $r_{\mathrm{mem}}$-bounded-influence.

Let $d:\mathcal E\times\mathcal E\to\mathbb R_{\geq0}$ be a metric on an ambient space $\mathcal E\supseteq\bigcup_{t\geq1}E_t$, and assume that $d|_{E_t\times E_t}$ is measurable for every $t$.
\begin{definition}[$(u,v)$-Lipschitz]
\label{def:lipschitz}
For $u,v\in\mathbb R_{\geq0}$, a problem $\mathcal P$ is \emph{$(u,v)$-Lipschitz} if, for every $m,n\in\mathbb Z_+$ with $m\leq n$, history $\mathcal I=(\bm e_{1:m-1},\bm a_{1:m-1})$, requests $e_m,e'_m\in E_m$, and common suffix $\bm e_{m+1:n}\in\bm E_{m+1:n}$,
\[
\left|
\mathrm{Opt}(\mathcal P^{\mathcal I},e_m\circ\bm e_{m+1:n})
-\mathrm{Opt}(\mathcal P^{\mathcal I},e'_m\circ\bm e_{m+1:n})
\right|
\leq L\min\{u\,d(e_m,e'_m),v\}.
\]
\end{definition}

\begin{definition}[$(u,v)$-strongly Lipschitz]
\label{def:strongly-lipschitz}
For $u,v\in\mathbb R_{\geq0}$, a problem $\mathcal P$ is \emph{$(u,v)$-strongly Lipschitz} if, under the same conditions, every fixed $\bm a_{m:n}\in\bm A_{m:n}$ satisfies
\[
\left|
\mathrm{Val}(\mathcal P^{\mathcal I},e_m\circ\bm e_{m+1:n},\bm a_{m:n})
-\mathrm{Val}(\mathcal P^{\mathcal I},e'_m\circ\bm e_{m+1:n},\bm a_{m:n})
\right|
\leq L\min\{u\,d(e_m,e'_m),v\}.
\]
\end{definition}
The $(u,v)$-strongly Lipschitz condition implies the $(u,v)$-Lipschitz condition by applying the fixed-action inequality to an optimizer in each direction.

Under the measurable discrete metric \(d_{\mathrm{disc}}(e,e')=\mathbb I(e\ne e')\), \(f\)-bounded influence and the one-period reward bound imply the \((f+1,f+1)\)-Lipschitz condition. To see this, fix the same action in the period whose request is changed. The immediate rewards differ by at most \(L\), and the optimal values for the remaining periods differ by at most \(fL\). Maximizing over the action in that period on each side preserves the bound \((f+1)L\) on the difference in optimal values. The strongly Lipschitz condition remains an additional assumption because it requires a bound on the value difference for every fixed action sequence.

\medskip
\noindent\textbf{Predictions and Effective-Horizon Estimation.}
The prediction $\bm e^*_{1:\infty}$ is \emph{perfect} when it equals the realized sequence $\bm e_{1:\infty}$. For a learning-augmented policy $\bm\pi$, let $\bm\pi(\bm e^*_{1:\infty})$ denote the policy instantiated with prediction $\bm e^*_{1:\infty}$. Whenever $\mathrm{Opt}(\mathcal P,\bm e_{1:\infty})>0$, define its \emph{instancewise performance ratio} for this realized-sequence/prediction pair by
\begin{align*}
\mathrm{Comp}(\mathcal P,\bm e_{1:\infty},\bm e^*_{1:\infty},\bm\pi)
\defeq
\frac{\mathrm{Val}(\mathcal P,\bm e_{1:\infty},
\bm\pi(\bm e^*_{1:\infty}))}
{\mathrm{Opt}(\mathcal P,\bm e_{1:\infty})}.
\end{align*}
Inputs with zero offline optimum remain admissible to the underlying problem and oracle guarantees. For reward maximization, a policy is \(x\)-consistent if this ratio is at least \(x\) for every admissible realized sequence with positive offline optimum under a perfect prediction, and \(y\)-robust if its ratio is at least \(y\) for every such sequence and every admissible prediction.

\phantomsection\label{para:effective-horizon-estimator}
We assume access to an \emph{effective-horizon estimator}. Given the observed requests and the prediction, it returns a finite upper bound on the effective horizon of the sequence formed by appending the predicted future requests to the observed prefix. Specifically, for every admissible realized sequence $\bm e_{1:\infty}$, admissible prediction $\bm e^*_{1:\infty}$, and $i\in\mathbb Z_{\geq0}$, it returns a finite integer
$
\widehat M_i
\defeq\mathrm{EstimateM}(\bm e_{1:i},\bm e^*_{1:\infty})
\geq\max\!\left\{i,
M(\bm e_{1:i}\circ\bm e^*_{i+1:\infty})\right\}.
$
For each fixed prediction and index $i$, the estimator is measurable as a function of the observed prefix $\bm e_{1:i}$.

\medskip
\noindent\textbf{Offline and Online Oracles.}
Assume access to the fixed-history offline and restartable online oracles defined below.

\begin{definition}[$\gamma$-approximate offline oracle]
\label{def:gamma-offline-oracle}
Fix $\gamma\in(0,1]$. For every $h\in\mathbb Z_{\geq0}$, endpoint $N\in\mathbb Z_+\cup\{\infty\}$ with $N\geq h+1$, history $\mathcal I_h=(\bm e_{1:h},\bm a_{1:h})$, and continuation $\bm e_{h+1:N}\in\bm E_{h+1:N}$, a \emph{$\gamma$-approximate offline oracle} returns $\bm a_{h+1:N}\in\bm A_{h+1:N}$ such that
\[
\mathrm{Val}(\mathcal P^{\mathcal I_h},\bm e_{h+1:N},\bm a_{h+1:N})
\geq\gamma\,\mathrm{Opt}(\mathcal P^{\mathcal I_h},\bm e_{h+1:N}).
\]
For $N=\infty$, let $T=M(\bm e_{1:h}\circ\bm e_{h+1:\infty})<\infty$. The oracle solves through $\max\{h,T\}$ (an empty continuation when $T\leq h$) and pads later actions according to the standing measurable selection rule.
\end{definition}

When $\gamma=1$, we call this an \emph{exact offline oracle}; the same terminology applies to the cost-minimization definition.

\begin{definition}[$\eta$-online oracle]
\label{def:eta-online-oracle}
Fix $f\in\mathbb R_{\geq0}$ such that $\mathcal P$ is $f$-bounded-influence, and fix $\eta\in(0,1]$. A family of randomized online policies
$\Pi=\{\bm\pi_h\}_{h\geq0}$, where
$\bm\pi_h=(\pi_{h,h+1},\pi_{h,h+2},\ldots)$, is an \emph{$\eta$-online oracle} if both of the following hold:
\begin{itemize}
\item For every $N\in\mathbb Z_+\cup\{\infty\}$ and $\bm e_{1:N}\in\bm E_{1:N}$, we have
$
\mathrm{Val}(\mathcal P,\bm e_{1:N},\bm\pi_0)
\geq\eta\,\mathrm{Opt}(\mathcal P,\bm e_{1:N}).
$
\item For every $h\in\mathbb Z_{\geq0}$, endpoint $N\in\mathbb Z_+\cup\{\infty\}$ with $N\geq h+1$, history $\mathcal I_h=(\bm e_{1:h},\bm a_{1:h})$, and continuation $\bm e_{h+1:N}\in\bm E_{h+1:N}$, we have
$
\mathrm{Val}(\mathcal P^{\mathcal I_h},\bm e_{h+1:N},
\bm\pi_h^{\mathcal I_h})
\geq\eta\,\mathrm{Opt}(\mathcal P^{\mathcal I_h},\bm e_{h+1:N})-fL.
$
Here $\bm\pi_h^{\mathcal I_h}$ denotes $\bm\pi_h$ restarted after the fixed history $\mathcal I_h$; the restriction of its policy kernels to the fixed history is stated in Section~\ref{sec:ec-framework-conventions}.
\end{itemize}
\end{definition}

\noindent\emph{Resettable-simulation assumption.}
Approximate-oracle variants additionally require a measurable resettable simulator. For any fixed history and finite continuation, conditional on the query data and all previously revealed randomness, it produces independent and identically distributed (i.i.d.) trajectories with the restarted online-oracle law. The simulator uses fresh randomness independent of the random seeds used by every online-oracle copy that generates executed actions (a \emph{live copy}). Simulation changes neither the state nor the random seed of any live copy. Section~\ref{sec:ec-framework-conventions} gives the formal details.

Table~\ref{tab:adaswitch-requirements} lists problem assumptions, planning inputs, and required oracles.

\begin{table}[htbp]
\centering
\caption{AdaSwitch requirements by offline-oracle regime.}
\label{tab:adaswitch-requirements}
\footnotesize
\setlength{\tabcolsep}{4pt}
\renewcommand{\arraystretch}{1.15}
\begin{tabular}{@{}
>{\raggedright\arraybackslash}m{2.4cm}
>{\raggedright\arraybackslash}m{6.2cm}
>{\raggedright\arraybackslash}m{6.2cm}
@{}}
\hline
Requirement & Exact-oracle variants & Approximate-oracle variants \\
\hline
Problem and sequence assumptions
& Finite effective horizons under prefix replacement (Definition~\ref{def:admissible-request-sequence}); $f$-bounded influence (Definition~\ref{def:bounded-influence}); $(u,v)$-Lipschitz condition (Definition~\ref{def:lipschitz}).
& Finite effective horizons under prefix replacement (Definition~\ref{def:admissible-request-sequence}); $f$-bounded influence (Definition~\ref{def:bounded-influence}); $(u,v)$-strongly Lipschitz condition (Definition~\ref{def:strongly-lipschitz}). \\
\hline
Prediction and planning endpoint
& One request-sequence prediction revealed before period~$1$ (p.~\pageref{para:request-prediction-information}); effective-horizon estimator (p.~\pageref{para:effective-horizon-estimator}).
& One request-sequence prediction revealed before period~$1$ (p.~\pageref{para:request-prediction-information}); effective-horizon estimator (p.~\pageref{para:effective-horizon-estimator}). \\
\hline
Oracle and simulation access
& Exact fixed-history offline oracle (Definition~\ref{def:gamma-offline-oracle}, $\gamma=1$); restartable online-oracle family (Definition~\ref{def:eta-online-oracle}).
& $\gamma$-approximate fixed-history offline oracle (Definition~\ref{def:gamma-offline-oracle}); restartable online-oracle family (Definition~\ref{def:eta-online-oracle}); resettable simulator (Section~\ref{sec:ec-framework-conventions}). \\
\hline
\end{tabular}
\par\smallskip
\begin{minipage}{\linewidth}
\footnotesize
Cost oracles: Definitions~\ref{def:gamma-offline-oracle-cost}--\ref{def:eta-online-oracle-cost}. Both cost variants permit a positive initial additive term $\beta$ (Section~\ref{sec:adaswitch-cost-minimization}).
\end{minipage}
\end{table}

\section{The Adaptive Switching Meta-Algorithm}
\label{sec:adaswitch-framework}
AdaSwitch begins with the online oracle and switches to an offline-oracle plan when the variant's entry conditions are met. The parameter $\epsilon$ sets the robustness slack and enters the switching thresholds, while $b$ and $c$ scale error clipping and those thresholds. In variants with an approximate offline oracle, $\alpha$ controls the batch-value threshold and the approximation-factor gap, and $H$ is the Monte Carlo sampling scale.

The algorithm's \emph{mode} is its conservative or predictive flag, an \emph{epoch} is a maximal consecutive run in one mode, and a \emph{batch} is a consecutive executed interval within a predictive epoch governed by one fixed action plan. For $b,c>0$ and $q\in\mathbb Z_{\geq0}$, define
$
\hat d_{b,c}(e,e')\defeq\min\{d(e,e'),c/b\},
\varphi_q^*(b,c)\defeq\sum_{t=1}^{q}\hat d_{b,c}(e_t,e_t^*).
$
Once $b,c$ are fixed, we suppress this dependence and write $\hat d$ and $\varphi_q^*$.

\subsection{AdaSwitch with an Exact Offline Oracle}
\label{sec:Adaptive-Switching-Algorithm-for-1-oracle}
Assume access to an exact offline oracle $\mathcal A$ and an $\eta$-online oracle $\Pi$. In a conservative epoch starting at $\tau$, the entry statistic at period $t$ is $\mathrm{Opt}(\mathcal P^{\mathcal I(\tau)},\bm e_{\tau:t})$, computed from the fixed pre-epoch history $\mathcal I(\tau)=(\bm e_{1:\tau-1},\bm a_{1:\tau-1})$. The entry statistic determines when conservative operation provides enough slack in the reward bound to cover losses from the following predictive epoch and any subsequent restart. Its value depends only on the fixed pre-epoch history and realized requests, actions sampled within the epoch do not affect it. AdaSwitch enters \(\mathtt{predictive}\) mode when this statistic reaches \(10cL/\epsilon\). In each predictive period, AdaSwitch reoptimizes its plan using the observed request and the predicted suffix. It exits predictive mode when the cumulative clipped error within the current epoch reaches $2c/(\eta b)$. 

In a predictive period $t$, the oracle is called at the finite endpoint $\widehat M_t=\mathrm{EstimateM}(\bm e_{1:t},\bm e^*_{1:\infty})$. The estimator makes optimization through $\widehat M_t$ equivalent to optimization on the full hybrid continuation.
\algtext*{EndWhile}% Remove "end while" text
\algtext*{EndIf}% Remove "end if" text
\algtext*{EndFor}% Remove "end for" text

\begin{algorithm}[h]
\caption{AdaSwitch with an Exact Offline Oracle}
\label{alg:Adaptive-Switching-Algorithm-for-1-oracle}
\begin{algorithmic}[1]
\State \textbf{Oracles:} the exact offline oracle $\mathcal A$ and the restartable online oracle $\Pi$, with known factor $\eta$.
\State \textbf{Input:} request prediction $\bm{e}^*_{1:\infty}$, effective-horizon estimator $\mathrm{EstimateM}$, slack parameter $\epsilon>0$, and threshold parameters $b,c>0$.
\State \textbf{Initialize:} $\mathrm{mode} \gets \mathtt{conservative}$ and the conservative-epoch start $\tau \gets 1$.

\For{$t = 1$ to $\infty$} 
    \State Observe request $e_t$. 
    \If{$\mathrm{mode} = \mathtt{conservative}$}
        \State Sample $a_t\sim\pi_{\tau-1,t}(\bm{e}_{1:t},\bm{a}_{1:t-1})$ and execute $a_t$.
        \State With $\mathcal I(\tau)=(\bm e_{1:\tau-1},\bm a_{1:\tau-1})$, compute $s=\mathrm{Opt}(\mathcal P^{\mathcal I(\tau)},\bm e_{\tau:t})$ using $\mathcal A$.
        \State \textbf{if} {$s \geq \frac{10cL}{\epsilon}$} \textbf{then} set $\mathrm{mode}\gets\mathtt{predictive}$ and initialize the next predictive epoch's cumulative clipped prediction error $\varphi\gets0$.
    \Else \Comment{$\mathrm{mode} = \mathtt{predictive}$}
        \State Let $\mathcal{I}(t)=(\bm{e}_{1:t-1}, \bm{a}_{1:t-1})$, set $\widehat M_t\gets\mathrm{EstimateM}(\bm e_{1:t},\bm e^*_{1:\infty})$, and invoke $\mathcal{A}$ to select
        \begin{align}
            \bm{a}^{*,(t)}_{t:\widehat M_t} \in \arg\max_{\bm{a}''_{t:\widehat M_t}\in\bm A_{t:\widehat M_t}} \mathrm{Val}\left(\mathcal{P}^{\mathcal{I}(t)}, e_t\circ\bm{e}^*_{t+1:\widehat M_t}, \bm{a}''_{t:\widehat M_t}\right). \label{eq:adaswitch-1-predictive-follow}
        \end{align}
        \State Execute $a_t = a^{*,(t)}_{t}$ and set $\varphi \gets \varphi + \hat d(e_t,e_t^*)$.
        \State \textbf{if} {$\varphi \geq  \frac{2  c}{\eta  b}$} \textbf{then} $\mathrm{mode} \gets \mathtt{conservative}$ and $\tau \gets t + 1$.
    \EndIf
\EndFor
\end{algorithmic}
\end{algorithm}

\begin{theorem}
\label{thm:ASA-error-dependent-1-oracle}
Under the standing admissibility convention, let $\mathcal{P}$ be $(u,v)$-Lipschitz and $f$-bounded-influence. Fix an online-oracle factor $0<\eta\leq1$, a slack parameter $\epsilon \in (0,\eta)$, a realized request sequence $\bm{e}_{1:\infty}$ with $\mathrm{Opt}(\mathcal{P},\bm{e}_{1:\infty})>0$, and a prediction $\bm{e}^*_{1:\infty}$. Choose threshold parameters $b,c$ such that $b>0$, $b\geq u$, and $c\geq\max(v,f,1)$. Let $M\defeq M(\bm e_{1:\infty})$. Then AdaSwitch with an exact offline oracle and an $\eta$-online oracle achieves
\begin{align} \label{eq:ASA-error-dependent-1-oracle}
\mathrm{Comp}(\mathcal{P},\bm{e}_{1:\infty},\bm{e}^*_{1:\infty},\mathrm{AdaSwitch})
\geq \max\left\{\eta-\epsilon,1-\frac{L(10c+7b\eta\varphi_M^*)}
{\epsilon\cdot\mathrm{Opt}(\mathcal{P},\bm{e}_{1:\infty})}\right\}.
\end{align}
\end{theorem}

\noindent\emph{Proof idea.}
Lemma~\ref{lem:predictive-following-finite-reward} guarantees reward on any finite predictive interval of at least its fixed-history optimum minus $2bL D+cL$, where $D$ is the interval's clipped error. Lipschitz sensitivity charges request discrepancies; bounded influence compares histories on the common predicted tail. Thus only errors within the interval enter the bound. Entry thresholds amortize additive losses, and completed predictive epochs consume disjoint portions of clipped error. Summing over epochs truncated at $M$ gives the guarantee using $\varphi_M^*$, even when plans extend beyond $M$. Section~\ref{sec:proof-thm-ASA-error-dependent-1-oracle} gives the full proof.

Under perfect predictions, the error-dependent bound approaches \(1\) as the optimum grows relative to the additive loss. The following corollary gives a corresponding bound in terms of the predicted optimum. Its proof appears in Section~\ref{sec:proof-cor-ASA-error-dependent-1-oracle-Rpre}.

\begin{corollary}[Loss bound expressed using the predicted optimum]
\label{cor:ASA-error-dependent-1-oracle-Rpre}
Under the assumptions of Theorem~\ref{thm:ASA-error-dependent-1-oracle}, let
$ M\defeq M(\bm e_{1:\infty}), M^*\defeq M(\bm e^*_{1:\infty}), M^\vee\defeq\max\{M,M^*\}$, and $\varphi_\vee^*\defeq\varphi_{M^\vee}^*$.
If $\mathrm{Opt}(\mathcal P,\bm e^*_{1:\infty})>0$, then
\begin{align}
\mathrm{Comp}(\mathcal{P},\bm{e}_{1:\infty},\bm{e}^*_{1:\infty},\mathrm{AdaSwitch})
\geq \max\left\{\eta-\epsilon,1-\frac{L(10c+8b\eta\varphi_\vee^*)}
{\epsilon\cdot\mathrm{Opt}(\mathcal{P},\bm{e}^*_{1:\infty})}\right\}.
\label{eq:ASA-error-dependent-1-oracle-Rpre}
\end{align}
\end{corollary}

The entry and exit thresholds in Algorithm~\ref{alg:Adaptive-Switching-Algorithm-for-1-oracle} can also be adjusted while retaining a chosen robustness target. Proposition~\ref{prop:ASA-adjustable-thresholds-1-oracle} gives a sufficient condition on these thresholds and shows how their values affect the error-dependent bound.

\subsection{\texorpdfstring{AdaSwitch with a $\gamma$-Approximate Offline Oracle}{AdaSwitch with a gamma-Approximate Offline Oracle}}
\label{sec:ASA-for-gamma-oracle}
The offline oracle’s approximation guarantee applies to the returned plan as a whole. Repeatedly reoptimizing and executing only the first action need not preserve this guarantee. Algorithm~\ref{alg:Adaptive-Switching-Algorithm-for-gamma-oracle} therefore executes fixed plans in batches, with the strongly Lipschitz condition bounding the effect of prediction errors on their rewards. 
In conservative mode, the entry test uses resettable simulations to estimate the expected reward of running the online oracle from the epoch's starting history through the current period. Conditional on that history, the simulations use randomness independent of the live copy that executes actions. Basing the entry test on the live copy's realized reward could make the switching time depend on its random actions, so the oracle's guarantee for a fixed endpoint would not apply directly.

For a batch beginning at $q$, define the hybrid prefix
\[
\widetilde{\bm e}^{(q)}_{q:r}\defeq
\begin{cases}
e_q\circ\bm e^*_{q+1:r}, & r\geq q,\\
\varnothing, & r<q.
\end{cases}
\]
For parameters $0<\epsilon<\eta$ and $\alpha,b,c>0$, define $\rho,\delta$ and the entry and exit thresholds $\Theta_{\mathrm c},\Theta_{\mathrm p}$ by
\[
\rho\defeq\frac{\gamma\alpha}{\alpha+\gamma},\qquad
\delta\defeq\eta-\frac{15\epsilon}{16},\qquad
\Theta_{\mathrm c}\defeq\frac{16\eta\alpha cL}{\epsilon},\qquad
\Theta_{\mathrm p}\defeq\frac{5\rho\alpha c}{\delta b}.
\]
Here $\gamma-\rho=\gamma^2/(\alpha+\gamma)$ is the reward-side approximation-factor gap; increasing $\alpha$ reduces this gap but enlarges the additive loss term.

For a conservative epoch beginning at $\tau$, write
$\mathcal I(\tau)\defeq(\bm e_{1:\tau-1},\bm a_{1:\tau-1})$ and let
$R^{\mathrm{on}}_{\tau,t}\defeq
\mathrm{Val}(\mathcal P^{\mathcal I(\tau)},\bm e_{\tau:t},
\bm\pi_{\tau-1}^{\mathcal I(\tau)})$
be the expected reward of an uninterrupted online-oracle continuation from $\mathcal I(\tau)$ through $t$.
Set the Monte Carlo error tolerance $\lambda\defeq\min\{1,L\}$. For a sampling scale $H>0$, set $n_{\tau,t}=\lceil H(t-\tau+1)^2\log(\mathrm e\,t)\rceil$, let
$Z^{(1)}_{\tau,t},\ldots,Z^{(n_{\tau,t})}_{\tau,t}$ be the total rewards of independent resettable simulations on $\bm e_{\tau:t}$ under $\bm\pi_{\tau-1}^{\mathcal I(\tau)}$, and define
$
\widehat R_{\tau,t}\defeq
\frac{1}{n_{\tau,t}}\sum_{\ell=1}^{n_{\tau,t}}Z^{(\ell)}_{\tau,t}.
$
Each estimate $\widehat R_{\tau,t}$ has conditional mean $R^{\mathrm{on}}_{\tau,t}$ given the query data and all randomness revealed before the call.

Fix a measurable fallback action sequence $\bm a^{\mathrm{fb}}_{1:\infty}\in\bm A_{1:\infty}$ in advance. If the candidate through $\widehat M_q$ has value below $\alpha cL$, execute $\bm a^{\mathrm{fb}}_{q:\infty}$ as the \emph{terminal fallback} until clipped error triggers a return to conservative mode. The approximation guarantee bounds the remaining hybrid optimum by $\alpha cL/\gamma$, which enters the additive loss.

\begin{algorithm}
\caption{AdaSwitch with a $\gamma$-Approximate Offline Oracle}
\label{alg:Adaptive-Switching-Algorithm-for-gamma-oracle}
\begin{algorithmic}[1]
\State \textbf{Oracle access:} a $\gamma$-approximate offline oracle $\mathcal A$, a restartable $\eta$-online oracle $\Pi$, and a resettable simulator for $\Pi$; $\gamma,\eta$ are known.
\State \textbf{Input:} prediction $\bm e^*_{1:\infty}$, effective-horizon estimator $\mathrm{EstimateM}$, and $0<\epsilon<\eta$, $\alpha,b,c>0$.
\State \textbf{Thresholds:} $\Theta_{\mathrm c},\Theta_{\mathrm p}$ as defined above. Fix $H\geq \frac{2L^2}{\lambda^2}\max\!\left\{3,\log\!\left(\frac{7680\eta\alpha cL}{\epsilon^2\lambda}\right)\right\}$.
\State \textbf{Initialize:} $\mathrm{mode}\gets\mathtt{conservative}$ and $\tau\gets1$.

\For{$t = 1$ to $\infty$} 
    \State Observe request $e_t$. 
    \If{$\mathrm{mode} = \mathtt{conservative}$}
        \State Sample $a_t\sim\pi_{\tau-1,t}(\bm{e}_{1:t},\bm{a}_{1:t-1})$ and execute it.
        \State Compute $\widehat R_{\tau,t}$ from $n_{\tau,t}$ fresh resettable simulations. \label{line:adaswitch-gamma-monte-carlo}
        \If{$\widehat R_{\tau,t}\geq\Theta_{\mathrm c}$} set $\mathrm{mode}\gets\mathtt{predictive}$, $\tau_p\gets t+1$, and $\varphi\gets0$.
        \EndIf
    \Else \Comment{$\mathrm{mode} = \mathtt{predictive}$}
        \If{$t=\tau_p$} \Comment{A new batch starts}
            \State $q\gets t$, $\mathcal I(q)\gets(\bm e_{1:q-1},\bm a_{1:q-1})$, and $\widehat M_q\gets\mathrm{EstimateM}(\bm e_{1:q},\bm e^*_{1:\infty})$.
            \State Initialize $r\gets q-1$ and $\bm a_{q:r}\gets\varnothing$.
            \While{$r<\widehat M_q$ \textbf{and} $\mathrm{Val}(\mathcal P^{\mathcal I(q)},\widetilde{\bm e}^{(q)}_{q:r},\bm a_{q:r})<\alpha cL$}
                \State Set $r\gets r+1$ and query $\mathcal A$ for a new candidate plan $\bm a_{q:r}$ satisfying
                \begin{align*}
                \mathrm{Val}(\mathcal P^{\mathcal I(q)},\widetilde{\bm e}^{(q)}_{q:r},\bm a_{q:r})
                \geq\gamma\,\mathrm{Opt}(\mathcal P^{\mathcal I(q)},\widetilde{\bm e}^{(q)}_{q:r}).
                \end{align*}
            \EndWhile
            \If{$\mathrm{Val}(\mathcal P^{\mathcal I(q)},\widetilde{\bm e}^{(q)}_{q:r},\bm a_{q:r})<\alpha cL$} set $\tau_p\gets\infty$ and $\bm a_{q:\infty}\gets\bm a^{\mathrm{fb}}_{q:\infty}$. \label{line:adaswitch-gamma-predictive-if-end}
            \Else{} set $\tau_p\gets r+1$.
            \EndIf
        \EndIf
        \State Execute $a_t$ from the current plan (or terminal fallback); set $\varphi\gets\varphi+\hat d(e_t,e_t^*)$.
        \State \textbf{if} $\varphi\geq\Theta_{\mathrm p}$ \textbf{then} $\tau\gets t+1$, $\mathrm{mode}\gets\mathtt{conservative}$. \label{line:adaswitch-gamma-predictive-stop}
    \EndIf
\EndFor
\end{algorithmic}
\end{algorithm}
Because $\widehat M_q\geq q$ is finite and $r$ increases by one before every oracle call, each batch-construction loop terminates after at most $\widehat M_q-q+1$ calls.

\begin{theorem}
\label{thm:ASA-error-dependent-gamma-oracle}
Under the standing admissibility convention, let $\mathcal P$ be $(u,v)$-strongly Lipschitz and $f$-bounded-influence. Fix $0<\eta\leq\gamma\leq1$, $\epsilon\in(0,\eta)$, a realized sequence $\bm e_{1:\infty}$ with $\mathrm{Opt}(\mathcal P,\bm e_{1:\infty})>0$, and a prediction $\bm e^*_{1:\infty}$. Choose $b,c,\alpha$ such that $b>0$, $b\geq u$, $c\geq\max\{v,f,1\}$, $\alpha\geq3$, and $\rho\geq\delta$. Let $M\defeq M(\bm e_{1:\infty})$. Then AdaSwitch with a $\gamma$-approximate offline oracle and an $\eta$-online oracle achieves
\begin{align*}
\mathrm{Comp}(\mathcal P,\bm e_{1:\infty},\bm e^*_{1:\infty},\mathrm{AdaSwitch})
\geq\max\left\{\eta-\frac{19}{20}\epsilon,
\rho-\frac{L}{\epsilon\,\mathrm{Opt}(\mathcal P,\bm e_{1:\infty})}
\left(17\alpha c+\frac{17}{5}b\eta\varphi_M^*\right)\right\}.
\end{align*}
\end{theorem}

To express this guarantee using the predicted optimum, use the common-horizon quantities $M^\vee,\varphi_\vee^*$ from Corollary~\ref{cor:ASA-error-dependent-1-oracle-Rpre}. When $\mathrm{Opt}(\mathcal P,\bm e^*_{1:\infty})>0$, the resulting bound is
\begin{align}
\mathrm{Comp}(\mathcal P,\bm e_{1:\infty},\bm e^*_{1:\infty},\mathrm{AdaSwitch})
\geq\max\left\{\eta-\frac{19}{20}\epsilon,
\rho-\frac{L}{\epsilon\,\mathrm{Opt}(\mathcal P,\bm e^*_{1:\infty})}
\left(17\alpha c+\frac{22}{5}b\eta\varphi_\vee^*\right)\right\}.
\label{eq:ASA-error-dependent-gamma-oracle-Rpre}
\end{align}
The theorem and this predicted-optimum bound are proved in Sections~\ref{sec:proof-of-theorem-ASA-error-dependent-gamma} and~\ref{sec:proof-gamma-oracle-Rpre}, respectively.

\subsection{AdaSwitch for Online Cost Minimization with Predictions}\label{sec:adaswitch-cost-minimization}
The cost variants retain the reward algorithms' two-mode structure, with guarantees expressed as upper bounds on the performance ratio. Both variants permit an initial additive term $\beta$ in the online-oracle guarantee.

Take
$
\mathcal P=\{\{E_t\}_{t\geq1},\{A_t\}_{t\geq1},\{C_t\}_{t\geq1}\}, 
C_t:\bm E_{1:t}\times\bm A_{1:t}\to[0,L],
$
where each stage cost $C_t$ is deterministic and measurable. Define values, continuation problems, effective horizons, and admissibility as in Section~\ref{sec:framework-formulation}, replacing rewards by costs. Define $\mathrm{Opt}$ by minimization:
\begin{align*}
\mathrm{Opt}(\mathcal{P},\bm{e}_{1:n}) \defeq \min_{\bm{a}_{1:n}\in\bm A_{1:n}}\mathrm{Val}(\mathcal{P},\bm{e}_{1:n},\bm{a}_{1:n}).
\end{align*}
The bounded-influence and Lipschitz definitions use these cost values and optima.
For cost minimization, a policy is $x$-consistent if its instancewise performance ratio satisfies $\mathrm{Comp}\leq x$ for every admissible realized sequence with positive offline optimum under a perfect prediction, and $y$-robust if $\mathrm{Comp}\leq y$ for every such sequence and every admissible prediction. 

\begin{definition}[$\gamma$-approximate offline oracle for cost minimization]\label{def:gamma-offline-oracle-cost}
Use Definition~\ref{def:gamma-offline-oracle}, but take $\gamma\geq1$ and replace the guarantee by
$
\mathrm{Val}(\mathcal P^{\mathcal I_h},\bm e_{h+1:N},\bm a_{h+1:N})
\leq\gamma\,\mathrm{Opt}(\mathcal P^{\mathcal I_h},\bm e_{h+1:N}).
$
\end{definition}
\begin{definition}[$(\eta,\beta)$-online oracle for cost minimization] \label{def:eta-online-oracle-cost}
Use Definition~\ref{def:eta-online-oracle}, but require the restarted guarantee only after positive-length histories $h\in\mathbb Z_+$. Take $\eta\geq1$ and a fixed $\beta\geq0$ independent of the request sequence and endpoint, and replace the two guarantees, respectively, by
\[
\mathrm{Val}(\mathcal P,\bm e_{1:N},\bm\pi_0)
\leq\eta\,\mathrm{Opt}(\mathcal P,\bm e_{1:N})+\beta
\quad
\mathrm{and}
\quad
\mathrm{Val}(\mathcal P^{\mathcal I_h},\bm e_{h+1:N},\bm\pi_h^{\mathcal I_h})
\leq\eta\left(\mathrm{Opt}(\mathcal P^{\mathcal I_h},\bm e_{h+1:N})+fL\right).
\]
An $(\eta,0)$-online oracle is abbreviated as an $\eta$-online oracle.
\end{definition}

The exact-oracle cost variant replaces maximization with minimization in predictive reoptimization. It uses an entry threshold of \(10(\eta+\epsilon)cL/\epsilon\) and a clipped-error exit threshold of \(2(\eta+\epsilon)c/b\). The approximate-oracle cost variant enters predictive mode when both the estimated online cost and the returned offline plan's cost reach their respective thresholds. Both costs are evaluated from the epoch's starting history using the requests observed so far. During batch construction, if the candidate plan's cost remains below the batch threshold at the estimated effective horizon, the algorithm retains that plan. Algorithms~\ref{alg:Adaptive-Switching-Algorithm-for-1-oracle-c} and~\ref{alg:Adaptive-Switching-Algorithm-for-gamma-oracle-c} in the companion give the full modifications for cost minimization.

The following setup applies to both results. Under the standing admissibility convention, fix a cost-minimization problem $\mathcal P$, structural constants $u,v,f\geq0$, a slack parameter $\epsilon>0$, a realized sequence $\bm e_{1:\infty}$ with $\mathrm{Opt}(\mathcal P,\bm e_{1:\infty})>0$, and any admissible prediction $\bm e^*_{1:\infty}$. Choose $b,c$ with $b>0$, $b\geq u$, and $c\geq\max\{v,f,1\}$, and set $M\defeq M(\bm e_{1:\infty})$.
\begin{theorem}
\label{thm:cost-version-ASA-error-dependent-1-oracle}
Under the common setup above, if $\mathcal P$ is $(u,v)$-Lipschitz and $f$-bounded-influence, then AdaSwitch with an exact offline oracle and an $(\eta,\beta)$-online oracle with $\eta\geq1$ and $\beta\geq0$ (Algorithm~\ref{alg:Adaptive-Switching-Algorithm-for-1-oracle-c}) achieves
\[
\begin{aligned}
\mathrm{Comp}(\mathcal{P},\bm{e}_{1:\infty},\bm{e}^*_{1:\infty},\mathrm{AdaSwitch}) 
& \leq \frac{\beta}{\mathrm{Opt}(\mathcal{P},\bm{e}_{1:\infty})}
+ \min\Bigg\{\eta+\epsilon, \\
& 1+\frac{L}{\epsilon \cdot \mathrm{Opt}(\mathcal{P},\bm{e}_{1:\infty})}
   \Bigg( 
   \left[2\epsilon+(\eta-1)(10\eta+11\epsilon)\right]c
   +\left[5(\eta-1)+4\epsilon\right]b\varphi_M^*
   \Bigg)\Bigg\}.
\end{aligned}
\]
The first branch bounds expected cost by $(\eta+\epsilon)\mathrm{Opt}+\beta$; for $\beta=0$, it gives a purely multiplicative robustness factor $\eta+\epsilon$.
\end{theorem}
\begin{theorem}
\label{thm:cost-version-ASA-error-dependent-gamma-oracle}
Under the common setup above, if $\mathcal P$ is $(u,v)$-strongly Lipschitz and $f$-bounded-influence, $1\leq\gamma\leq\eta$, and $\alpha\geq \max(16\gamma,\gamma+\frac{2\gamma^2}{\epsilon})$, then AdaSwitch with a $\gamma$-approximate offline oracle and an $(\eta,\beta)$-online oracle with $\beta\geq0$ (Algorithm~\ref{alg:Adaptive-Switching-Algorithm-for-gamma-oracle-c}) achieves
\[
\begin{aligned}
\mathrm{Comp}(\mathcal{P},\bm{e}_{1:\infty},\bm{e}^*_{1:\infty},\mathrm{AdaSwitch})
&\leq\frac{\beta}{\mathrm{Opt}(\mathcal{P},\bm{e}_{1:\infty})} +\min\Bigg\{\eta+\frac{19\epsilon}{20}
+\frac{15\epsilon\beta}{16\eta\,\mathrm{Opt}(\mathcal{P},\bm{e}_{1:\infty})},\,\\
&\frac{\gamma\alpha}{\alpha-\gamma}
+\frac{L}{\epsilon\cdot\mathrm{Opt}(\mathcal{P},\bm{e}_{1:\infty})}
\Bigg(
\gamma\alpha\eta(20\eta+21\epsilon)c
+\gamma\left(4\eta+\frac{8}{3}\epsilon\right)b\varphi_M^*
\Bigg)\Bigg\}.
\end{aligned}
\]
The first branch bounds expected cost by $(\eta+19\epsilon/20)\mathrm{Opt}+(1+15\epsilon/(16\eta))\beta$; for $\beta=0$, it gives a purely multiplicative robustness factor $\eta+19\epsilon/20$.
\end{theorem}
The proofs are in Sections~\ref{sec:proof-thm-cost-version-ASA-error-dependent-1-oracle} and~\ref{sec:proof-thm-cost-version-ASA-error-dependent-gamma-oracle}. The different denominators reflect the one-sided batching bounds: the reward analysis lower-bounds collected value and yields $\alpha+\gamma$, whereas the cost analysis upper-bounds accumulated cost and yields $\alpha-\gamma$, with $\alpha>\gamma$ required by the stated parameter conditions.

\noindent\emph{Computational scope.}
Let $M=M(\bm e_{1:\infty})$ denote the realized effective horizon, and let $\overline M\defeq\max_{0\leq i\leq M}\widehat M_i$ be the largest estimated planning endpoint through period \(M\). Since \(M\) and each estimate are finite, \(\overline M<\infty\). All planning queries during periods \(1,\ldots,M\) use only predicted requests through period \(\overline M\). For \(M\geq1\), we count oracle calls and simulations only during these periods. The exact variants make at most \(M\) offline-oracle calls. The approximate variants make at most \(M\overline M\) offline-oracle calls during predictive batch construction, with at most \(M\) additional calls for the conservative entry test in cost minimization. Their sampling schedule generates \(O(M+HM^3\log(\mathrm e M))\) simulated trajectories, comprising \(O(M^2+HM^4\log(\mathrm e M))\) simulated periods in total. These counts exclude the running time of individual oracle calls.

\section{Application I: Online Lead-Time Quotation with Predictions}
\label{sec:lead-time-quotation}
We apply the bounded-influence framework to \emph{online lead-time quotation (OLTQ)}~\citep{keskinocak2001scheduling} and its learning-augmented version, \emph{OLTQ with predictions (OLTQwP)}. The decision maker irrevocably assigns each arriving request to a processing period. Shorter quoted lead times earn larger rewards.

\subsection{Problem Setting}

The decision maker operates a single processor that can process one request per period $t\in\mathbb Z_+$. At the start of period $t$, the decision maker observes a raw count $\widetilde e_t\in\mathbb Z_{\geq0}$ of new requests. Fix a patience limit $\ell\in\mathbb Z_+$. Because the requests are identical and only periods $t,\ldots,t+\ell-1$ can yield positive reward, copies beyond the first $\ell$ may be rejected without loss. We therefore use the capped count $e_t\defeq\min\{\widetilde e_t,\ell\}$ and cap prediction counts identically.

For each retained request copy $i\in[e_t]$, the decision maker immediately and irrevocably chooses a processing period $a_t(i)\geq t$, thereby quoting lead time $d_t(i)\defeq a_t(i)-t$. The request is served only if $d_t(i)<\ell$ and its processing period is not already assigned to an earlier-arriving request or a smaller-indexed request from the same period; otherwise it is rejected. A served request with quoted lead time $d_t(i)$ yields
\begin{equation*}
    r_\ell(d_t(i)) =
    \begin{cases}
        (\ell - d_t(i)) \cdot \omega, & \text{if } d_t(i) < \ell, \\
        0, & \text{otherwise},
    \end{cases}
\end{equation*}
where $\omega>0$ is the per-unit revenue rate. We normalize $\omega=1$. Accordingly, we take
$
E_t\defeq\{0,1,\ldots,\ell\},
A_t\defeq\{t,t+1,\ldots,t+\ell-1,\infty\}^{\ell}.
$
Here \(a_t(i)=\infty\) represents rejection, and coordinates \(i>e_t\) are ignored. For a realized action sequence, the reward credited in processing period \(t\) is
\begin{align*}
R_t^{\mathrm{OLTQ}} (\bm{e}_{1:t},\bm{a}_{1:t})=\sum_{s=1}^t\sum_{i=1}^{e_s}[s+\ell-a_s(i)]_+\cdot\mathbb I(a_s(i)=t)\cdot\prod_{j<s}\prod_{k=1}^{e_j}\mathbb I(a_j(k)\neq t)\cdot\prod_{k=1}^{i-1}\mathbb I(a_s(k)\neq t).
\end{align*}
We use the convention \([s+\ell-\infty]_+=0\). The indicator products implement arrival-time and within-period priority, so at most one summand is nonzero and the framework reward bound is $L=\ell$. For a finite continuation, we count only rewards credited during that continuation, including fixed rewards from inherited reservations. For an arrival horizon \(T\), we extend the evaluation through processing period \(T+\ell-1\) to capture the full reward.

We restrict both the realized and predicted request sequences to have finitely many arrivals in total and otherwise use the framework definitions of \(\mathrm{Val}\), \(\mathrm{Opt}\), and \(\mathrm{Comp}\).

\subsection{Applying AdaSwitch}
We use the natural metric $d(e_t,e'_t)\defeq |e_t-e'_t|$. Past decisions affect the future only through reserved slots in the $\ell$-period lead-time window, while perturbing one arrival count changes only a bounded amount of feasible reward. The following lemma (proved in Section~\ref{sec:proof-p-lead-bounded-influence}) gives the resulting bounded-influence and Lipschitz properties.
\begin{lemma}
\label{lem:p-lead-bounded-influence}
    The $\mathrm{OLTQ}$ problem is $(\ell-1)$-bounded-influence and $\left(1,\frac{\ell+1}{2}\right)$-strongly Lipschitz.
\end{lemma}

\medskip
\noindent\underline{The effective-horizon estimator.}
For a sequence \(\bm e'_{1:\infty}\) with finitely many arrivals in total, let
$
\tau(\bm e'_{1:\infty})
\defeq\max(\{t:e'_t>0\}\cup\{0\}).
$
Assume that the last positive-arrival period \(\tau(\bm e^*_{1:\infty})\) is supplied with the prediction. Rewards vanish after period \(\tau(\bm e'_{1:\infty})+\ell-1\), so \(M(\bm e'_{1:\infty})\leq\tau(\bm e'_{1:\infty})+\ell-1\); if the offline optimum is positive, equality holds because one last-arriving request can be served in its latest profitable slot. Hence
$
\mathrm{EstimateM}(\bm e_{1:i},\bm e^*_{1:\infty})
=\max\{i,\tau(\bm e^*_{1:\infty})\}+\ell-1
$
is a valid estimator, because the last arrival in \(\bm e_{1:i}\circ\bm e^*_{i+1:\infty}\) occurs no later than \(\max\{i,\tau(\bm e^*_{1:\infty})\}\).

\medskip
\noindent\underline{The exact offline oracle.}
For a fixed history $\mathcal I=(\bm e_{1:m},\bm a_{1:m})$ and a continuation $\bm e_{m+1:m+n}$ of length $n\in\mathbb Z_{\geq0}\cup\{\infty\}$, the exact offline oracle solves
\begin{align}
\arg\max_{\bm{a}'_{m+1:m+n}\in\bm A_{m+1:m+n}}\mathrm{Val}(\mathrm{OLTQ}^\mathcal{I},\bm{e}_{m+1:m+n},\bm{a}'_{m+1:m+n}).    \label{eq:lead-time-1-oracle-problem}
\end{align}
The O-HRR scheduling algorithm of \citet{keskinocak2001scheduling} solves the no-history case by scanning slots in order and scheduling the available request with the highest remaining revenue. Its fixed-history extension O-HRR$^*$ skips inherited reservations, whose rewards are fixed. The last processing period examined in Eq.~\eqref{eq:lead-time-1-oracle-problem} is
\[
T_{m,n}^{\mathrm{off}}
\defeq
\begin{cases}
m+n,&n<\infty,\\
M(\bm e_{1:\infty}),&n=\infty.
\end{cases}
\]
Admissibility ensures a finite scan, which is empty when $T_{m,n}^{\mathrm{off}} \le m$.

\begin{framed}
\noindent\textbf{Algorithm O-HRR}$^*$:
\begin{itemize}
    \item Initialize $\bm a_{m+1:m+n}$ to rejection ($\infty$), representing entries beyond the finite scan implicitly.
    \item For each processing period $t=m+1,m+2,\ldots,T_{m,n}^{\mathrm{off}}$, continue to the next period if a request copy in the fixed history reserves slot $t$. Otherwise, define
    \[
    \mathcal U_t=\{i:t\geq i\geq\max(m+1,t-\ell+1)\ \wedge\ \exists 1\leq j\leq e_i\text{ such that }a_i(j)=\infty\}
    \]
    to be the set of arrival periods containing an unscheduled request copy that could earn positive reward in slot $t$.
    If $\mathcal U_t\neq\emptyset$, let $s_t^*=\max\mathcal U_t$ be the latest such arrival period. Select a copy $j\leq e_{s_t^*}$ with $a_{s_t^*}(j)=\infty$ and set $a_{s_t^*}(j)\leftarrow t$.
    \item Return the updated schedule $\bm a_{m+1:m+n}$.
\end{itemize}
\end{framed}

\begin{lemma}
\label{lem:O-HRR-star-offline-oracle}
Under the standing admissibility convention, O-HRR$^*$ is an exact offline oracle for OLTQ.
\end{lemma}
The proof appears in Section~\ref{sec:proof-O-HRR-star-offline-oracle} of the electronic companion.

\medskip
\noindent\underline{The online oracle.}
We extend the Q-FRAC online policy of \citet{keskinocak2001scheduling} to fixed histories and denote the resulting policy by Q-FRAC$^*$. Its fixed integer threshold sets the minimum reward of assignments admitted by its internal schedule. We choose this threshold, $k_\ell^*$, to maximize the resulting competitive factor. For $j\in[\ell]$, define
\[
\begin{aligned}
g_\ell(j)&\defeq
\frac{(\ell+j)(\ell-j+1)}{\ell(\ell+1)},
&
k_\ell^*&\defeq
\min\!\left(\operatorname*{arg\,max}_{j\in[\ell]}
\min\left\{\frac{j}{\ell},g_\ell(j)\right\}\right),\\[-1pt]
\eta^{\mathrm{OLTQ}}
&\defeq\min\left\{\frac{k_\ell^*}{\ell},g_\ell(k_\ell^*)\right\},
&
\theta_\ell^*&\defeq\sqrt{\frac54+\frac1\ell}-\frac12.
\end{aligned}
\]
Equivalently,
\[
\eta^{\mathrm{OLTQ}}
=\max\left\{
\frac{\lfloor\theta_\ell^*\ell\rfloor}{\ell},
g_\ell(\lceil\theta_\ell^*\ell\rceil)
\right\}.
\]

\begin{framed}
\noindent\textbf{Algorithm Q-FRAC$^*$}. For each $m\in\mathbb Z_{\geq0}$, define a base policy $\bm\pi_m$ as follows. After any fixed history $\mathcal I$ of length $m$, its restarted version $\bm\pi_m^{\mathcal I}$ uses a fresh internal schedule.
\begin{itemize}
    \item Initialize the next internally unfilled slot $U_{m+1}\leftarrow m+1$.
    \item For each processing period $t=m+1,m+2,\ldots$, observe the $e_t$ request copies arriving in period $t$. Let the number proposed for internal scheduling be
    $
    N_t\leftarrow\max\{0,\min(e_t,t+\ell-U_t+1-k_\ell^*)\}.
    $
    The policy $\pi_{m,t}$ sets
    \[
    a_t(i)\leftarrow
    \begin{cases}
    U_t+i-1,&\text{if }i\in[N_t],\\
    \infty,&\text{otherwise},
    \end{cases}
    \qquad
    U_{t+1}\leftarrow\max\{t+1,U_t+N_t\}.
    \]
\end{itemize}
\end{framed}
This fresh internal schedule ignores inherited reservations. Conflicting new assignments earn zero, but such conflicts occur in at most $\ell-1$ slots. The resulting reward loss is therefore at most $\ell(\ell-1)=fL$, as permitted by the restarted-oracle condition. The following lemma states the resulting guarantee; its proof appears in Section~\ref{sec:proof-Q-FRAC-eta} of the electronic companion.

\begin{lemma}
\label{lem:Q-FRAC-star-online-oracle}
The restarted Q-FRAC$^*$ policy family is an $\eta^{\mathrm{OLTQ}}$-online oracle for OLTQ.
\end{lemma}

\medskip
We instantiate AdaSwitch with the effective-horizon estimator and the two oracles above, set $b=1$ and $c=\ell$, and call the resulting policy $\mathrm{AdaSwitch}\text{-}\mathrm{OLTQ}$. These choices satisfy $b\geq1$ and $c\geq\max\{(\ell+1)/2,\ell-1,1\}$. Moreover, $c/b=\ell\geq|e_t-e_t^*|$, so clipping is inactive and $\varphi_q^*=\sum_{t=1}^{q}|e_t-e_t^*|$ for every $q\in\mathbb Z_{\geq0}$. Theorem~\ref{thm:ASA-error-dependent-1-oracle} now gives the following guarantee.

\begin{theorem}
\label{thm:competitive-ratio-adaswitch-OLTQ}
    Consider any realized request sequence $\bm{e}_{1:\infty}$ and predicted request sequence $\bm{e}^*_{1:\infty}$ such that $\sum_{t=1}^\infty e_t < \infty$, $\sum_{t=1}^\infty e_t^* < \infty$, and $\mathrm{Opt}(\mathrm{OLTQ},\bm{e}_{1:\infty})>0$. Set $M\defeq M(\bm e_{1:\infty})=\tau(\bm e_{1:\infty})+\ell-1$. Then, for every slack parameter $\epsilon\in(0,\eta^{\mathrm{OLTQ}})$,
	    \[
	    \begin{aligned}
	    &\mathrm{Comp}(\mathrm{OLTQ},\bm{e}_{1:\infty},\bm{e}^*_{1:\infty},\mathrm{AdaSwitch}\text{-}\mathrm{OLTQ})\\
	    &\qquad\geq \max\left\{\eta^{\mathrm{OLTQ}}-\epsilon,1-\frac{\ell}{\epsilon\cdot\mathrm{Opt}(\mathrm{OLTQ},\bm{e}_{1:\infty})}\left(10\ell+7\eta^{\mathrm{OLTQ}}\varphi_M^*\right)\right\}.
	    \end{aligned}
	    \]
\end{theorem}

% The adjustable-threshold analogue follows from Proposition~\ref{prop:ASA-adjustable-thresholds-1-oracle}.

\medskip
\noindent\underline{Comparison with Q-FRACwP.}
For the comparison, fix a known arrival horizon $T\in\mathbb Z_+$, take $e_t=e_t^*=0$ for $t>T$, and retain all rewards earned through $T+\ell-1$. Fix a target robustness \(r\in(0,\eta^{\mathrm{OLTQ}})\). Q-FRACwP is the prediction-informed dynamic-threshold policy of \citet{huo2026online}; Q-FRAC uses the fixed integer threshold defined above. Define
\[
\begin{aligned}
\xi(r)&\defeq \sup\left\{x\in[0,1]:\frac{(\ell+\lceil x\ell\rceil)(\ell-\lceil x\ell\rceil+1)}{(\ell+1)\ell}\geq r\right\},\\
\kappa(r)&\defeq\lceil r\ell\rceil,\\
\beta(r)&\defeq \min\left\{\xi(r),
\frac{(\ell+\kappa(r))(\ell-\kappa(r)+1)}{\ell(\ell+1)}
\right\}.
\end{aligned}
\]
With these parameters, Q-FRACwP is \(r\)-robust and \(\beta(r)\)-consistent~\citep{huo2026online}. Setting $\epsilon=\eta^{\mathrm{OLTQ}}-r$ gives AdaSwitch-OLTQ the same robustness guarantee $r$. Table~\ref{tab:comparison-between-our-results-for-app-I-and-related-work} summarizes the comparison.

\begin{table}[htbp]
  \centering
  \caption{Q-FRACwP~\citep{huo2026online} and AdaSwitch-OLTQ for sequences with a common known arrival horizon.}
  \label{tab:comparison-between-our-results-for-app-I-and-related-work}
  \small
  \setlength{\tabcolsep}{4pt}
  \renewcommand{\arraystretch}{1.55}
  \begin{tabular}{@{}
  >{\raggedright\arraybackslash}m{2.6cm}
  >{\raggedright\arraybackslash}m{5.4cm}
  >{\raggedright\arraybackslash}m{7.4cm}@{}}
  \hline
  Policy &
  \makecell[l]{Perfect-prediction guarantee\\ at robustness $r$} &
  \makecell[l]{Arbitrary-prediction guarantee} \\
  \hline
  Q-FRACwP & $\beta(r)$ & $r$ \\
  \hline
  AdaSwitch-OLTQ &
  \makecell[l]{$\displaystyle\max\!\left\{r, \displaystyle 1-\frac{10\ell^2}{(\eta-r)\mathrm{Opt}_{\mathrm{pred}}}\right\}$} &
  \makecell[l]{$\displaystyle\max\!\left\{r, \displaystyle 1-\frac{\ell(10\ell+7\eta\varphi_M^*)}{(\eta-r)\mathrm{Opt}_{\mathrm{real}}}\right\}$} \\
  \hline
  \multicolumn{3}{@{}p{15.4cm}@{}}{Here $\eta\defeq\eta^{\mathrm{OLTQ}}$, $\mathrm{Opt}_{\mathrm{pred}}\defeq\mathrm{Opt}(\mathrm{OLTQ},\bm e^*_{1:\infty})$, $\mathrm{Opt}_{\mathrm{real}}\defeq\mathrm{Opt}(\mathrm{OLTQ},\bm e_{1:\infty})$, and, because clipping is inactive, $\varphi_M^*=\sum_{t=1}^{M}|e_t-e_t^*|$. The AdaSwitch entries are instance-dependent.} \\
  \end{tabular}
\end{table}

To retain Q-FRACwP's perfect-prediction guarantee while exploiting AdaSwitch's stronger bound on sufficiently large instances, define Strengthened AdaSwitch-OLTQ using the predicted optimum. When $\beta(r)=1$, it uses Q-FRACwP. When $\beta(r)<1$, it selects AdaSwitch-OLTQ if
$
\mathrm{Opt}(\mathrm{OLTQ},\bm e^*_{1:\infty})
\geq
\frac{10\ell^2}{(\eta^{\mathrm{OLTQ}}-r)(1-\beta(r))}.
$
Otherwise, it selects Q-FRACwP. Proposition~\ref{prop:strengthened-adaswitch-oltq} in the electronic companion shows that the policy remains $r$-robust and retains $\beta(r)$ as a  perfect-prediction guarantee, with strict improvement when the displayed inequality is strict.
\FloatBarrier

\section{\texorpdfstring{Application II: The $k$-Server Problem with Predictions}{Application II: The k-Server Problem with Predictions}}
\label{sec:k-server}
We now turn to cost minimization and apply the framework to the classical \emph{$k$-server problem} ($k\mathrm{SE}$) and its learning-augmented version, \emph{$k$-server with predictions} ($k\mathrm{SEwP}$).

\subsection{Problem Setting}
Fix $k\in\mathbb Z_+$ and a metric space $(X,d)$ with $d:X\times X\to[0,1]$. In every period $t$, the request $e_t$ lies in $E_t\defeq X$. The space contains a distinguished no-request element $\bot$ satisfying $d(x,\bot)=1$ for every $x\neq\bot$. For a request sequence, define its last real-request period by
\[
T_{\mathrm{req}}(\bm e_{1:\infty})
\defeq
\sup\bigl(\{t\in\mathbb Z_+:e_t\neq\bot\}\cup\{0\}\bigr)
\in\mathbb Z_{\geq0}\cup\{\infty\}.
\]
The sequence has \emph{finite support} if $T_{\mathrm{req}}(\bm e_{1:\infty})<\infty$. Throughout this section, both realized and predicted request sequences have finite support; internal $\bot$ periods are allowed because they leave the configuration unchanged and incur zero cost.

The decision maker starts from a server configuration $\bm S=(S_1,\ldots,S_k)\in X^k$. Let $\bm S_t=(S_{t,1},\ldots,S_{t,k})$ be the configuration after period $t$, with $\bm S_0=\bm S$. After observing $e_t$, the decision maker chooses $a_t\in A_t\defeq[k]$. If $e_t\neq\bot$, server $a_t$ moves to $e_t$ and incurs its movement distance; if $e_t=\bot$, the action is ignored and the state is unchanged. Thus
\begin{equation}
\label{eq:k-server-cost-function}
\begin{aligned}
C_t^{k\mathrm{SE}_{\bm S}}(\bm e_{1:t},\bm a_{1:t})
&=\mathbb I(e_t\neq\bot)d(S_{t-1,a_t},e_t),\\
S_{t,i}
&=
\begin{cases}
e_t,&e_t\neq\bot\text{ and }i=a_t,\\
S_{t-1,i},&\text{otherwise}.
\end{cases}
\end{aligned}
\end{equation}
Let $k\mathrm{SE}_{\bm S}$ denote this problem. Equation~\eqref{eq:k-server-cost-function} gives the one-period bound $L=1$. We otherwise adopt the policy, value, optimum, and instancewise performance ratio conventions of Sections~\ref{sec:framework-formulation} and~\ref{sec:adaswitch-cost-minimization}.

Some formulations of the \(k\)-server problem also allow servers that do not serve the current request to reposition. Restricting movement to the server serving the current request does not weaken competitive guarantees. To see this, simulate any policy that allows additional movements. Whenever the simulated policy serves a request, move the corresponding server directly to the requested point and leave all other servers fixed. By the triangle inequality, each server's total movement cost is no greater than under the original policy.

\subsection{Applying AdaSwitch}
A fixed history affects a continuation only through the resulting server configuration. The minimum total distance over matchings between any two configurations is at most $k$, while changing one request changes the optimum by at most twice the request distance. Section~\ref{sec:proof-k-server-finite-influence} gives the details.

\begin{lemma}
\label{lem:k-server-finite-influence}
For every $\bm S\in X^k$, the problem $k\mathrm{SE}_{\bm S}$ is $k$-bounded-influence and $(2,2)$-Lipschitz.
\end{lemma}

\medskip
\noindent\underline{The effective-horizon estimator.}
Assume that the last real-request period $T_{\mathrm{req}}(\bm e^*_{1:\infty})$ is supplied with the prediction. Because all later costs vanish and the same bound holds for every realized-prefix/predicted-suffix splice, 
$
\mathrm{EstimateM}(\bm e_{1:i},\bm e^*_{1:\infty})
\defeq\max\{i,T_{\mathrm{req}}(\bm e^*_{1:\infty})\}
$
is a valid estimator, even when the splice contains internal $\bot$ periods.

\medskip
\noindent\underline{The exact offline oracle.}
Assume that each finite oracle input explicitly lists the initial server locations and requests, and that pairwise distances  between locations are provided as binary-encoded rational numbers. After deleting $\bot$ periods, an acyclic minimum-cost-flow formulation solves offline $k\mathrm{SE}$ exactly in time polynomial in the total encoded input size~\citep{chrobak1991new,tarjan1983data}. A history only replaces the initial configuration by its induced configuration $\bm S^{\mathcal I}$, so the same reduction is an exact fixed-history offline oracle; the formal residual-problem observation appears in Section~\ref{sec:proof-k-server-finite-influence}.

\medskip
\noindent\underline{The online oracles.}
For a configuration $\bm C=(C_1,\ldots,C_k)$, write $\mathrm{Disp}(\bm C)\defeq\sum_{i<j}d(C_i,C_j)$. We use two guarantees for a WFA-based oracle: the classical bound $(2k-1)\mathrm{Opt}+2\mathrm{Disp}(\bm C)$ with an initial-dispersion term and the strict bound $(4k-2)\mathrm{Opt}$ with no additive term (Lemma~\ref{lem:wfa-multiset-strict}). Section~\ref{sec:k-server-WFA} constructs the underlying measurable nonanticipating policy $\mathrm{WFA}^{\mathrm{lim}}_{\bm C}$, permitting co-located servers, as a limit of WFA instances in which those servers are separated by vanishing auxiliary distances. Set
\[
(\eta_{\mathrm{add}},\beta_{\mathrm{add}})
\defeq(2k-1,2\mathrm{Disp}(\bm S)),
\qquad
(\eta_{\mathrm{str}},\beta_{\mathrm{str}})
\defeq(4k-2,0).
\]
Because $d\leq1$, every history-induced configuration $\bm C$ satisfies $2\mathrm{Disp}(\bm C)\leq k(k-1)\leq(2k-1)k=\eta_{\mathrm{add}}fL$. The fixed-history construction in Section~\ref{sec:k-server-WFA} therefore gives an $(\eta_{\mathrm{add}},\beta_{\mathrm{add}})$-online oracle and an $\eta_{\mathrm{str}}$-online oracle.

\medskip
For each $z\in\{\mathrm{add},\mathrm{str}\}$, instantiate cost-minimization AdaSwitch using the estimator, the exact offline oracle, and the corresponding online oracle above, with $b=2$, $c=\bar k\defeq\max\{k,2\}$, and $\eta=\eta_z$.
Lemma~\ref{lem:k-server-finite-influence} gives $u=v=2$ and $f=k$, so $b\geq u$ and $c\geq\max\{v,f,1\}$. Moreover, $d(e_t,e_t^*)\leq1\leq c/b$, so clipping is inactive and $\varphi_q^*=\sum_{t=1}^{q}d(e_t,e_t^*)$ for every $q\in\mathbb Z_{\geq0}$. Call the resulting policy $\mathrm{AdaSwitch}\text{-}k\mathrm{SE}^{z}$. Theorem~\ref{thm:cost-version-ASA-error-dependent-1-oracle} gives the following guarantee.

\begin{theorem}
\label{thm:k-server-adaswitch-se}
Consider any initial configuration $\bm S\in X^k$ and finite-support realized and predicted request sequences $\bm e_{1:\infty}$ and $\bm e^*_{1:\infty}$. Suppose $\mathrm{Opt}(k\mathrm{SE}_{\bm S},\bm e_{1:\infty})>0$, and define $M\defeq M(\bm e_{1:\infty})$ and $\bar k\defeq\max\{k,2\}$. For every $\epsilon>0$ and $x\geq1$, let
$
\Delta_k(x)\defeq\frac{1}{\epsilon}\left(
\left[2\epsilon+(x-1)(10x+11\epsilon)\right]\bar k
+\left[10(x-1)+8\epsilon\right]\varphi_M^*
\right).
$
Then, for every $z\in\{\mathrm{add},\mathrm{str}\}$,
\[
\mathrm{Comp}(k\mathrm{SE}_{\bm{S}},\bm{e}_{1:\infty},\bm{e}^*_{1:\infty},
\mathrm{AdaSwitch}\text{-}k\mathrm{SE}^{z})
\leq
\frac{\beta_z}{\mathrm{Opt}(k\mathrm{SE}_{\bm S},\bm e_{1:\infty})}
+\min\left\{\eta_z+\epsilon,\,
1+\frac{\Delta_k(\eta_z)}
{\mathrm{Opt}(k\mathrm{SE}_{\bm S},\bm e_{1:\infty})}\right\}.
\]
\end{theorem}

The variant using the additive WFA guarantee has a leading factor $2k-1+\epsilon$ and additive term $\beta_{\mathrm{add}}\leq k(k-1)$; the variant using the strict WFA guarantee has a purely multiplicative robustness factor $4k-2+\epsilon$. For either variant, the error-dependent bound approaches $1$ when $\beta_z+\Delta_k(\eta_z)=o(\mathrm{Opt})$. If the initial servers are co-located, $\beta_{\mathrm{add}}=0$ and the variant using the additive guarantee has a purely multiplicative robustness factor $2k-1+\epsilon$.

\subsection{The Caching Problem: A Special Case with the Uniform Metric}
\label{sec:caching}
Under the uniform metric $d(e,e')=\mathbb I(e\neq e')$, restricting $k$-server to \emph{lazy policies} gives standard caching: a hit leaves the cache unchanged and costs zero, while a miss loads the requested page at unit cost, evicting a resident page when necessary. Denote this problem by $\mathrm{CA}_{\bm S}$ and caching with predictions by CAwP. When interpreting a server configuration as a cache state, $\bot$ entries and all but one copy of each duplicated real page are treated as empty slots.

We use fresh-start Marking~\citep{fiat1991competitive}, denoted $\mathrm{Mark}^{\mathrm{fs}}$, as the fixed-history online oracle. It keeps all pages unmarked through initial hits, begins its first phase at the first miss, and then follows standard randomized Marking, filling empty slots before evicting pages. The following lemma gives a logarithmic guarantee with no additive term; Section~\ref{sec:proof-marking-strict-online-oracle} provides the full specification and proof.

\begin{lemma}
\label{lem:strict-marking-bound}
Let $H_k\defeq\sum_{j=1}^k 1/j$ be the $k$-th harmonic number. For every initial cache state $\bm S\in X^k$ and every finite request sequence $\bm e_{1:n}$, fresh-start Marking satisfies
$
\mathrm{Val}(\mathrm{CA}_{\bm S},\bm e_{1:n},\mathrm{Mark}^{\mathrm{fs}})
\leq 2H_k\,\mathrm{Opt}(\mathrm{CA}_{\bm S},\bm e_{1:n}).
$
Consequently, restarting this policy from the cache state induced by each fixed history gives a $2H_k$-online oracle with no additive term. For the logarithmic instantiation below, we use the looser factor $\eta^{\mathrm{CA}}\defeq2(\ln k+1)\geq2H_k$.
\end{lemma}

Define $\mathrm{AdaSwitch}\text{-}\mathrm{CA}$ by applying cost-minimization AdaSwitch to the full request sequence, using the exact offline oracle and effective-horizon estimator above, independent fresh-start instances of Marking as the online oracle, and
$
b=2, c=\bar k\defeq\max\{k,2\},
\eta=\eta^{\mathrm{CA}}=2(\ln k+1).
$
The parameter checks from the general $k$-server case remain valid. Because $d(e,e')\in\{0,1\}$ and $c/b\geq1$, clipping is again inactive and $\varphi_q^*=\sum_{t=1}^{q}\mathbb I(e_t\neq e_t^*)$ for every $q\in\mathbb Z_{\geq0}$. Theorem~\ref{thm:cost-version-ASA-error-dependent-1-oracle} gives the following result.
\begin{theorem}
\label{thm:cache-adaswitch-ca}
Let $(X,d)$ be a uniform metric space. Consider any initial cache state $\bm S\in X^k$ and finite-support realized and predicted request sequences $\bm e_{1:\infty}$ and $\bm e^*_{1:\infty}$. Suppose $\mathrm{Opt}(\mathrm{CA}_{\bm S},\bm e_{1:\infty})>0$, and define
$
M\defeq M(\bm e_{1:\infty}),
\bar k\defeq\max\{k,2\}.
$
Then, for every $\epsilon>0$,
\[
\begin{aligned}
&\mathrm{Comp}(\mathrm{CA}_{\bm{S}},\bm{e}_{1:\infty},\bm{e}^*_{1:\infty},
\mathrm{AdaSwitch}\text{-}\mathrm{CA})\\
&\quad\leq
\min\Bigg\{
\eta^{\mathrm{CA}}+\epsilon,\,
1+\frac{
\left[2\epsilon+(\eta^{\mathrm{CA}}-1)(10\eta^{\mathrm{CA}}+11\epsilon)\right]\bar{k}}
{\epsilon\cdot\mathrm{Opt}(\mathrm{CA}_{\bm{S}},\bm{e}_{1:\infty})}
+\frac{
\left[10\eta^{\mathrm{CA}}-10+8\epsilon\right]\varphi_M^*}
{\epsilon\cdot\mathrm{Opt}(\mathrm{CA}_{\bm{S}},\bm{e}_{1:\infty})}
\Bigg\}.
\end{aligned}
\]
\end{theorem}
Choosing $\epsilon=\eta^{\mathrm{CA}}$ and using $\eta^{\mathrm{CA}}\geq2$ gives
\[
\mathrm{Comp}(\mathrm{CA}_{\bm S},\bm e_{1:\infty},\bm e^*_{1:\infty},
\mathrm{AdaSwitch}\text{-}\mathrm{CA})
\leq
\min\left\{2\eta^{\mathrm{CA}},\,
1+\frac{21\eta^{\mathrm{CA}}\bar k+18\varphi_M^*}
{\mathrm{Opt}(\mathrm{CA}_{\bm S},\bm e_{1:\infty})}\right\}.
\]
Under the displayed choice, the error-dependent bound approaches $1$ when $[\bar k(\ln k+1)+\varphi_M^*]/\mathrm{Opt}(\mathrm{CA}_{\bm S},\bm e_{1:\infty}) \to 0$.

\medskip
\noindent\underline{Comparison with policy combiners.}
We compare AdaSwitch-CA with the two policy combiners Blum--Burch~\citep{wei2020better,antoniadis2023online} and DART~\citep{christianson2023optimal} under a common caching setup. All three use the same initial cache and fixed request-sequence prediction, with fresh-start Marking as the baseline. The combiners' advice policies replan each period from their own caches using the same exact offline oracle as AdaSwitch-CA.

Under matched parameter choices, AdaSwitch-CA matches the leading robustness factor of the cited Blum-Burch bound without an additive term. Its robustness factor is also smaller than those in the cited DART bounds. For the same fixed parameter choices, AdaSwitch-CA's expected cost-ratio bound under perfect predictions approaches \(1\) as the offline optimum grows. This bound falls below that of DART for sufficiently large offline optimum. Section~\ref{sec:ec-related-comparison} and Table~\ref{tab:switching-frameworks} give the parameter matching and detailed bounds.

\section{Application III: Online Reusable Resource Allocation with Predictions}
\label{sec:ORRA}
We return to reward maximization with online reusable resource allocation with predictions (ORRAwP), based on the model of \citet{delong2024online}. For fixed reuse duration $D\in\mathbb Z_+$, we construct an exact fixed-history dynamic program and give an additional guarantee conditional on a supplied $\gamma$-approximate oracle.

\subsection{Problem Setting}
An instance has $n$ reusable resources, initially all available. In period $t\in\mathbb Z_+$, a request $e_t=(e_t(1),\ldots,e_t(n))\in E_{n,t}\defeq\{0,1\}^n$ reveals its eligible resources. After observing $e_t$, the decision maker chooses $a_t\in A_{n,t}\defeq[n]\cup\{0\}$, where $0$ rejects and $i\in[n]$ attempts to assign resource $i$. The attempt succeeds exactly when $e_t(i)=1$ and resource $i$ is available; it then earns one reward and makes $i$ unavailable in periods $t+1,\ldots,t+D-1$. Every other action earns zero and leaves each resource's next available period unchanged. We denote this problem by $\mathrm{ORRA}_n$. Section~\ref{sec:proof-ORRA} gives the formal state recursion and reward function.

Let
$
\widetilde M(\bm e_{1:\infty})
\defeq
\min\{i\in\mathbb Z_{\geq0}:e_j=\bm0\text{ for every }j>i\}.
$
We restrict both the realized and predicted sequences to finite support, so $\widetilde M(\bm e_{1:\infty})<\infty$ and $\widetilde M(\bm e^*_{1:\infty})<\infty$. We otherwise use the policy, value, and optimum conventions of Section~\ref{sec:framework-formulation}.

\subsection{Applying AdaSwitch}

Write $\delta_{\mathrm{req}}(e,e')\defeq\mathbb I(e\neq e')$ for the request metric. Rejecting requests for $D-1$ consecutive periods clears inherited assignments. With a fixed continuation action sequence, changing one eligibility vector can affect only the resource attempted in that period. Successes occurring in only one of the two runs alternate between them, so cumulative rewards differ by at most one. The following lemma, proved in Section~\ref{sec:proof-lem-ORRA-bounded-influence-Lipschitz}, gives both structural properties.

\begin{lemma}
\label{lem:ORRA-bounded-Lipschitz}
For every $n\in\mathbb{Z}_+$, $\mathrm{ORRA}_n$ is $(D-1)$-bounded-influence and $(1,1)$-strongly Lipschitz.
\end{lemma}

\noindent\underline{The effective-horizon estimator.}
Assume that the last nonzero-request period $\widetilde M(\bm e^*_{1:\infty})$ is supplied with the prediction. For every finite-support sequence $\bm e'$, the effective horizon equals its last nonzero-request period, and hence $M(\bm e'_{1:\infty})=\widetilde M(\bm e'_{1:\infty})$. Then $\mathrm{EstimateM}(\bm e_{1:i},\bm e^*_{1:\infty}) =\max\{i,\widetilde M(\bm e^*_{1:\infty})\}$ is a valid estimator.

\medskip
\noindent\underline{The offline oracle.}
For a length-$h$ history and length-$T$ continuation, the exact dynamic program in Section~\ref{sec:proof-ORRA-fixed-history-DP} records successful assignments during the preceding $D-1$ periods. It forms its initial state in $O(h+D)$ steps, uses at most $(n+1)^{D-1}$ states per period, and examines $O(T(n+1)^D)$ state--action pairs. It is polynomial in $h,T,n$ for fixed $D$ but exponential in $D$.

\medskip
\noindent\underline{The $\eta$-online oracle.}
In our unit-reward model, \emph{Periodic Reranking} (PRR)~\citep{delong2024online} draws fresh independent resource priorities at the start of each $D$-period block and assigns each request to the highest-priority available eligible resource, or rejects it if none exists. With parameter $\beta_{\mathrm{PRR}}=0.89$, its expected reward is at least $\eta^{\mathrm{ORRA}}\defeq0.589$ times a linear-programming upper bound on the offline optimum. Define the restartable extension PRR$^*$ as follows: $\bm\pi_0$ starts PRR with all resources available; after a history of length $h\geq1$, $\bm\pi_h$ rejects requests for $D-1$ periods and restarts PRR at $h+D$ with fresh independent priorities and all resources available, beginning a local $D$-period block.

\begin{lemma}
\label{lem:ORRA-online-oracle}
PRR$^*$ is an $\eta^{\mathrm{ORRA}}$-online oracle for $\mathrm{ORRA}_n$.
\end{lemma}
Section~\ref{sec:proof-lem-ORRA-online-oracle}  proves Lemma~\ref{lem:ORRA-online-oracle}. PRR$^*$ also supports independent simulations from any fixed history, with each simulated run generating its own fresh random resource priorities. This provides the resettable-simulation access required by batched AdaSwitch with a \(\gamma\)-approximate offline oracle.

With these offline and online oracles in place, we now instantiate AdaSwitch for ORRA. Set $\bar D\defeq\max\{D-1,1\}$. Here $L=1$, and Lemma~\ref{lem:ORRA-bounded-Lipschitz} permits $b=1$ and $c=\bar D$. Because $\delta_{\mathrm{req}}\in\{0,1\}$ and $c/b\geq1$, clipping is inactive, so $\varphi_q^*=\sum_{t=1}^{q}\mathbb I(e_t\neq e_t^*)$ for every $q\in\mathbb Z_{\geq0}$. Consider finite-support realized and predicted sequences with $\mathrm{Opt}(\mathrm{ORRA}_n,\bm e_{1:\infty})>0$, and set
$
M\defeq M(\bm e_{1:\infty})=\widetilde M(\bm e_{1:\infty}).
$
Denote the variants using exact and $\gamma$-approximate fixed-history offline oracles by $\mathrm{AdaSwitch}\text{-}\mathrm{ORRA}^{\mathrm{exact}}$ and $\mathrm{AdaSwitch}\text{-}\mathrm{ORRA}^{\mathrm{batch}}$, respectively. Theorems~\ref{thm:ASA-error-dependent-1-oracle} and~\ref{thm:ASA-error-dependent-gamma-oracle} give the following guarantees.

\begin{corollary}
\label{cor:ORRA-adaswitch-orra-exact}
For every $\epsilon\in(0,\eta^{\mathrm{ORRA}})$, the exact-oracle variant satisfies
\[
\mathrm{Comp}(\mathrm{ORRA}_n,\bm e_{1:\infty},\bm e^*_{1:\infty},
\mathrm{AdaSwitch}\text{-}\mathrm{ORRA}^{\mathrm{exact}})
\geq
\max\left\{
\eta^{\mathrm{ORRA}}-\epsilon,
1-\frac{10\bar D+7\eta^{\mathrm{ORRA}}\varphi_M^*}
{\epsilon\cdot\mathrm{Opt}(\mathrm{ORRA}_n,\bm e_{1:\infty})}
\right\}.
\]
\end{corollary}

\begin{theorem}
\label{thm:ORRA-adaswitch-orra}
Fix $\gamma\in[\eta^{\mathrm{ORRA}},1]$, $\epsilon\in(0,\eta^{\mathrm{ORRA}})$, and $\alpha\geq3$ such that $\frac{\gamma\alpha}{\alpha+\gamma}\geq\eta^{\mathrm{ORRA}}-\frac{15\epsilon}{16}$. Suppose a $\gamma$-approximate fixed-history offline oracle satisfying Definition~\ref{def:gamma-offline-oracle} is supplied. Use PRR$^*$ in conservative mode and its resettable simulator for the entry test, with the sampling scale prescribed in Algorithm~\ref{alg:Adaptive-Switching-Algorithm-for-gamma-oracle}. Then $\mathrm{AdaSwitch}\text{-}\mathrm{ORRA}^{\mathrm{batch}}$ satisfies
\[
\mathrm{Comp}(\mathrm{ORRA}_n,\bm e_{1:\infty},\bm e^*_{1:\infty},\mathrm{AdaSwitch}\text{-}\mathrm{ORRA}^{\mathrm{batch}})
\geq\max\Bigg\{\eta^{\mathrm{ORRA}}-\frac{19}{20}\epsilon,
\frac{\gamma\alpha}{\alpha+\gamma}
-\frac{17\alpha\bar D+\frac{17}{5}\eta^{\mathrm{ORRA}}\varphi_M^*}
{\epsilon\,\mathrm{Opt}(\mathrm{ORRA}_n,\bm e_{1:\infty})}\Bigg\}.
\]
\end{theorem}

\medskip
\noindent\underline{Comparison with Periodic Reranking.}
PRR provides a certified prediction-free factor of $\eta^{\mathrm{ORRA}}=0.589$ against the offline optimum~\citep{delong2024online}. Corollary~\ref{cor:ORRA-adaswitch-orra-exact} retains robustness factor $\eta^{\mathrm{ORRA}}-\epsilon$ for arbitrary admissible predictions, while its error-dependent bound approaches~$1$ when $(\bar D+\eta^{\mathrm{ORRA}}\varphi_M^*)/(\epsilon\mathrm{Opt})\to0$. Under perfect predictions, its guaranteed factor exceeds PRR's certified factor whenever
$
\mathrm{Opt}>\frac{10\bar D}{\epsilon(1-\eta^{\mathrm{ORRA}})}.
$

The error-dependent bound in Theorem~\ref{thm:ORRA-adaswitch-orra} approaches $\gamma\alpha/(\alpha+\gamma)$ for fixed $\alpha$ when $(\alpha\bar D+\eta^{\mathrm{ORRA}}\varphi_M^*)/(\epsilon\mathrm{Opt})\to0$, while retaining the PRR baseline up to $19\epsilon/20$.

\section{Numerical Experiments}
\label{sec:numerical-experiment}

Using exact-oracle AdaSwitch on OLTQwP and CAwP, we examine how the robustness target and effective horizon affect performance under perfect and corrupted predictions.  Full protocols, additional OLTQwP demand and horizon studies, and caching stress tests appear in Sections~\ref{sec:additional-oltq-experiments} and~\ref{sec:additional-caching-experiments}.

\subsection{OLTQwP: Perfect-Prediction Performance under Varying Guaranteed Robustness}\label{sec:experiment-oltq}
We compare AdaSwitch-OLTQ with Q-FRACwP and two prediction-free benchmarks, Q-FRAC and Immediate service, under perfect predictions. Immediate service serves one new request at once in each nonempty arrival period and rejects the others.

We use capped geometric demand that allows periods with no arrivals: $\Pr(G_t=g)=p(1-p)^g$ for $g\in\mathbb Z_{\geq0}$ and $e_t=\min\{G_t,\ell\}$. With $p=0.1$, each period has a $10\%$ probability of no new arrivals. These periods can be used to serve requests accepted earlier. The three settings use patience limits $\ell\in\{20,30,40\}$ at known arrival horizon $T=15000$.

We vary a common guaranteed reward-ratio target $r$ for AdaSwitch-OLTQ and Q-FRACwP. AdaSwitch sets $\epsilon_r=\eta^{\mathrm{OLTQ}}-r$, giving a reward threshold of $10\ell^2/\epsilon_r$ for entering predictive mode. Increasing $r$ strengthens the worst-case requirement, reduces $\epsilon_r$, and raises this threshold. Across 200 paired trials, we report mean ratios to the exact optimum and pointwise $95\%$ confidence intervals. Section~\ref{sec:additional-oltq-experiments} provides the full experimental protocols and additional studies of how arrival-horizon length, demand patterns, and prediction errors affect AdaSwitch-OLTQ's performance.

\begin{figure}[htbp]
    \centering
    \includegraphics[width=0.9\textwidth]{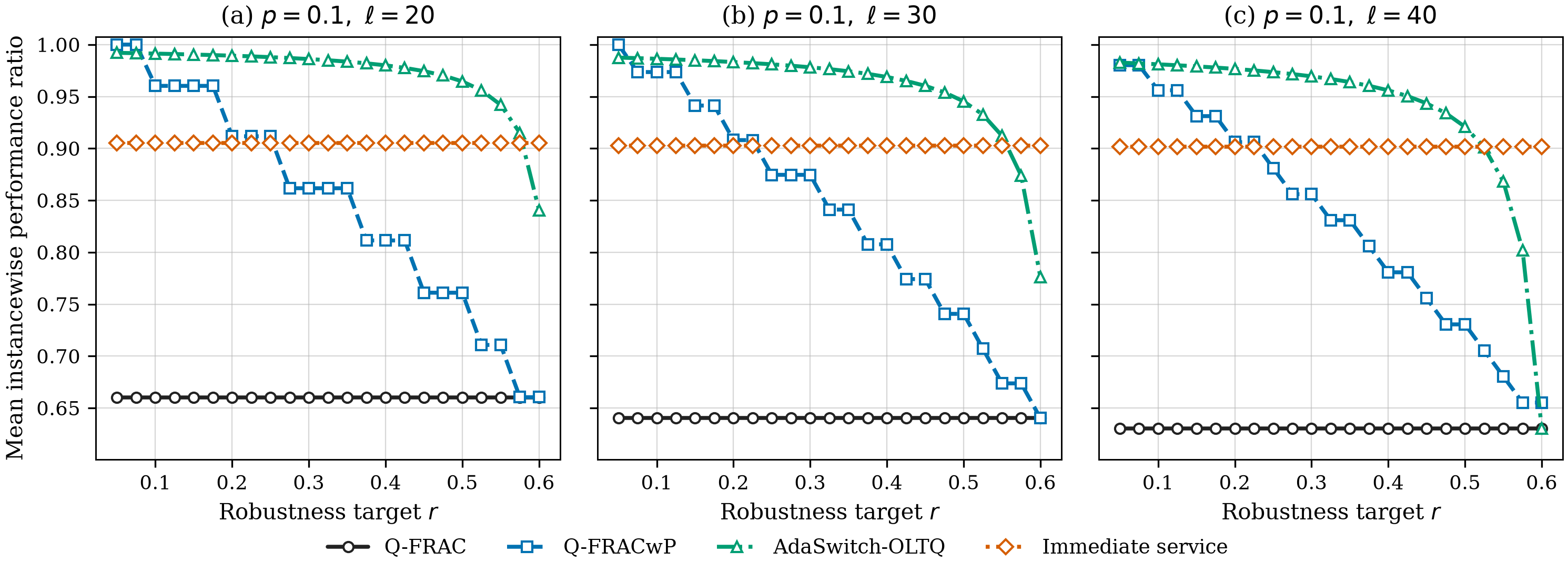}
    \caption{Perfect-prediction OLTQwP performance as the guaranteed robustness target $r$ for AdaSwitch-OLTQ and Q-FRACwP varies. Higher ratios are better.}
    \label{fig:empirical-consistency-different-robustness}
\end{figure}

Figure~\ref{fig:empirical-consistency-different-robustness} shows that AdaSwitch-OLTQ has higher mean reward ratios than Q-FRACwP over most of the grid. For example, at $(p,\ell,r)=(0.1,20,0.425)$, their means are $0.978$ and $0.811$, respectively, compared with $0.905$ for Immediate service. At some stricter robustness targets, Immediate service achieves a higher mean reward ratio than AdaSwitch-OLTQ, although it need not satisfy those robustness targets. At $r=0.600$ with $\ell=40$, the entry threshold exceeds the total reward bound $\ell(T+\ell-1)$. AdaSwitch therefore follows Q-FRAC throughout, explaining the endpoint drop despite perfect predictions.

\subsection{CAwP: Effects of Robustness Slack and Effective Horizon}\label{sec:experiment-caching}
We examine AdaSwitch-CA's sensitivity to robustness slack, effective horizon, and prediction quality. Marking~\citep{fiat1991competitive} serves as the prediction-free baseline, Predictive Marker~\citep{lykouris2021competitive} as a learning-augmented caching benchmark, and Predicted Belady~\citep{belady1966study,sleator1985amortized} as a prediction-following reference that is optimal under perfect predictions.

For the empirical comparison, we evaluate all policies on the same realized requests and derive all predictive inputs from a common predicted request sequence. Predictive Marker receives next-arrival forecasts obtained from this sequence. On a miss with a full cache, Predicted Belady evicts the page whose next predicted request is farthest in the future. The empirical comparison thus uses a common forecast source, while the error-dependent theoretical bounds use different prediction-error measures.

We use uniform and Zipf request distributions over \(100\) pages, with Zipf exponent \(0.8\). In both experiments, the effective horizon equals the request-sequence length, \(M=T\), and AdaSwitch-CA uses \(b=2\) and \(c=\bar{k}\). 
Across 200 paired trials, we report mean ratios of each policy's cost to the exact offline optimum and pointwise \(95\%\) confidence intervals. Section~\ref{sec:additional-caching-experiments} provides the complete protocol, including the within-trial averaging procedure, advice conversion, and additional stress-test comparisons.

\begin{figure}[h]
    \centering
    \includegraphics[width=0.45\textwidth]{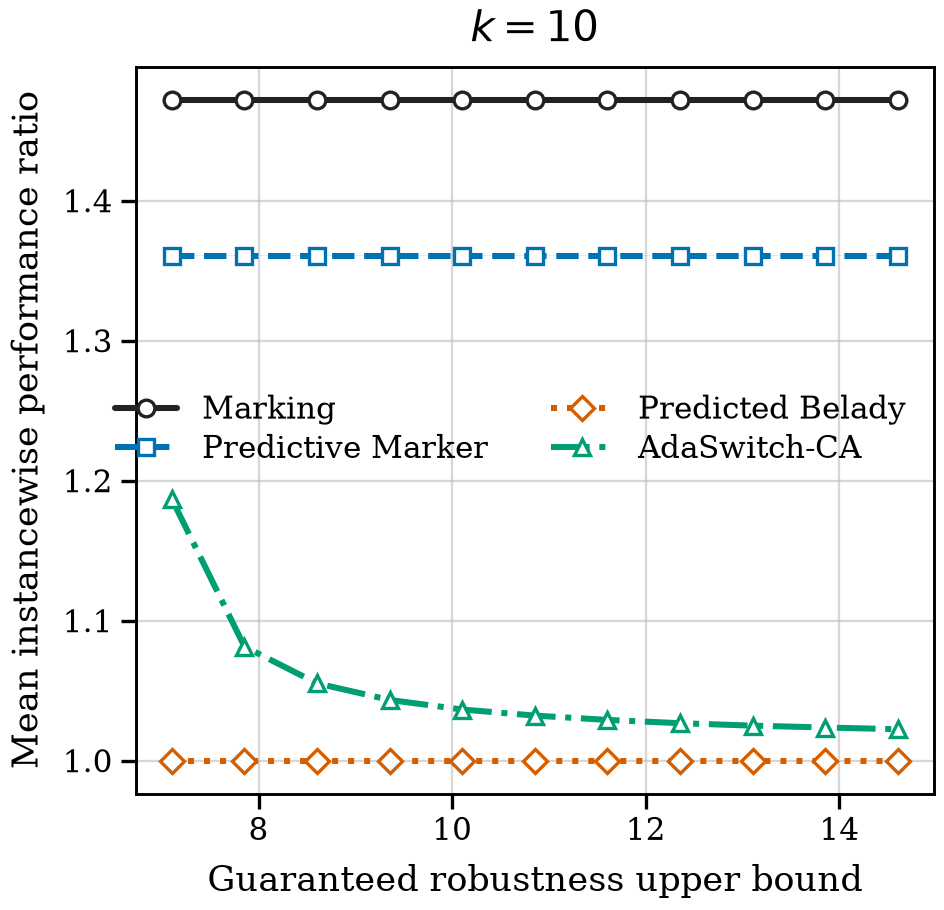}
    \caption{Perfect-prediction CAwP performance as AdaSwitch-CA's guaranteed robustness upper bound varies for $k=10$. The comparator policies are independent of $\epsilon$.  Lower ratios are better.}
    \label{fig:cache-consistency-robustness}
\end{figure}

The first experiment fixes \(k=10\) and varies AdaSwitch-CA's robustness slack \(\epsilon\) under perfect predictions. Its guaranteed robustness upper bound is \(2(\ln k+1)+\epsilon\), so increasing \(\epsilon\) relaxes the worst-case guarantee. The comparator policies are independent of \(\epsilon\). 
Figure~\ref{fig:cache-consistency-robustness} shows that AdaSwitch-CA's mean cost ratio decreases as its robustness slack increases, narrowing the gap from Predicted Belady. It remains below those of Marking and Predictive Marker across the tested slack values.

\begin{figure}[h]
    \centering
    \includegraphics[width=0.8\textwidth]{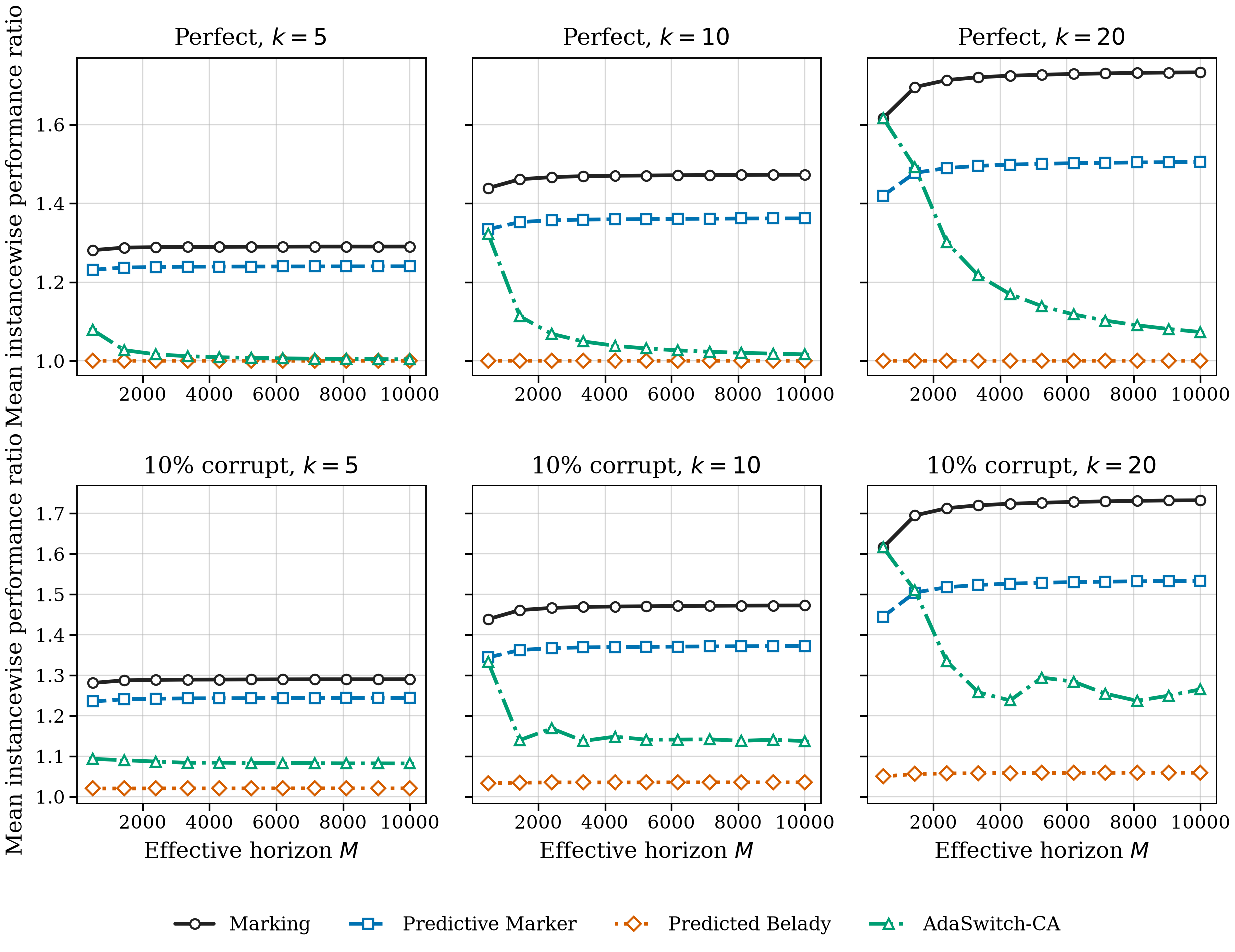}
    \caption{CAwP performance as the effective horizon increases.  Lower ratios are better.}
    \label{fig:cache-opt-scale}
\end{figure}

The second experiment varies the effective horizon for \(k\in\{5,10,20\}\) under perfect and \(10\%\)-corrupted predictions. AdaSwitch-CA uses \(\epsilon=2(\ln k+1)\) in this experiment. To generate corrupted predictions, we independently replace each entry with probability \(0.1\), choosing uniformly from the other \(99\) pages. 
Figure~\ref{fig:cache-opt-scale} shows that, over the tested horizons, AdaSwitch-CA tends to achieve lower mean cost ratios at longer horizons, especially for larger caches, and a smaller gap from Predicted Belady under perfect predictions. Under $10\%$ corruption, its mean cost ratio remains above $1$, with noticeable fluctuations for $k=20$. The mean cost ratios of Marking and Predictive Marker show less variation across the tested horizons.

\FloatBarrier
\section{Managerial Interpretation and Implementation Considerations}
\label{sec:managerial-interpretation}
\label{sec:parameter-calibration}

\noindent\textbf{Robustness and predictive entry.}
A stricter robustness target can delay predictive entry. For exact-oracle reward maximization, a target $r\in(0,\eta)$ gives $\epsilon=\eta-r$; for cost minimization, a target leading factor $y_{\mathrm c}>\eta$ gives $\epsilon=y_{\mathrm c}-\eta$ and expected cost at most $y_{\mathrm c}\mathrm{Opt}+\beta$. For example, with fixed patience limit $\ell$ in AdaSwitch-OLTQ and $\eta=\eta^{\mathrm{OLTQ}}$, raising the target from $\eta/2$ to $3\eta/4$ halves $\epsilon$ and doubles the entry reward threshold, while the clipped-error exit threshold remains $2\ell/\eta$.

\noindent\textbf{Approximation and parameter choices.}
Within the parameter ranges of Theorems~\ref{thm:ASA-error-dependent-gamma-oracle} and~\ref{thm:cost-version-ASA-error-dependent-gamma-oracle}, increasing $\alpha$ brings the limiting guarantee closer to the offline-oracle factor $\gamma$, but raises switching thresholds and additive penalties. Better limiting guarantees can therefore require larger instances before predictive operation becomes beneficial.

\noindent\textbf{Prediction error and instance scale.}
For fixed parameters, gains depend on prediction error and history losses relative to $\mathrm{Opt}$. The history scale $cL$ is $\ell^2$ for OLTQ, $\bar k$ for $k$-server and caching, and $\bar D$ for ORRA. Exact-oracle bounds approach $1$ when $cL/\mathrm{Opt}$ and $bL\varphi_M^*/\mathrm{Opt}$ vanish; approximate-oracle bounds approach their approximation-dependent limits. Cost variants also require $\beta/\mathrm{Opt}\to0$. When $\mathrm{Opt}=\Theta(M)$, sublinear accumulated error $\varphi_M^*=o(M)$ suffices with the other parameters fixed.

\noindent\textbf{Computation and deployment considerations.}
The guarantees exclude computation. Implementation requires a valid effective-horizon estimator, verified structural constants, and the specified fixed-history oracles; approximate variants also need independent resettable simulation. Operational studies should measure oracle latency and call counts, simulation cost, entry timing, switch count, and time spent in predictive mode. The online oracle alone may be preferable when normalized losses remain large or planning costs outweigh the anticipated gain.

\section{Conclusion and Future Directions}\label{sec:conclusion}
AdaSwitch combines prediction-informed offline plans with a restartable online oracle for history-dependent reward maximization and cost minimization. Under the stated structural and oracle-access assumptions, it preserves baseline guarantees in expectation up to tunable slack and the stated additive terms, while providing error-dependent bounds. The applications demonstrate bounded influence through expiring commitments in OLTQ and ORRA and bounded costs of reconciling configurations in $k$-server and caching. Synthetic experiments with exact offline oracles illustrate gains from relaxing robustness requirements and increasing the effective horizon in the tested settings.

Several directions remain open. Matching lower bounds under the same request-sequence prediction and oracle-access assumptions would clarify whether AdaSwitch's dependence on prediction error, instance scale, and robustness slack can be improved. Interval, distributional, or progressively refined predictions would require error measures and switching rules suited to the information available at each decision. Regret against specified policy benchmarks and guarantees under stochastic request models could complement competitive analysis. For approximate-oracle variants, reducing the number of simulated trajectories while controlling estimation error at adaptive switching times could lower computational cost. Finally, decaying effects of past decisions may permit sharper history-loss bounds and less conservative switching thresholds. Sensitivity results showing decay with temporal or network distance under suitable assumptions motivate this direction~\citep{na2020exponential,shin2021controllability,shin2022exponential}, but using them to bound differences in continuation values after different histories requires further analysis.

\bibliographystyle{informs2014}
\bibliography{references}

%% Here starts the e-companion (EC)
%%%%%%%%%%%%%%%%%%%%%%%%%%%%%%%%%%%%%%%%%%%%%%%%%%%%%%%%%%
\ECSwitch
%\ECDisclaimer
%%%%%%%%%%%%%%%%%%%%%%%%%%%%%%%%%%%%%%%%%%%%%%%%%%%%%%%%%%

\makeatletter
\providecommand{\theHsection}{}
\renewcommand{\theHsection}{EC.\arabic{section}}

\providecommand{\theHsubsection}{}
\renewcommand{\theHsubsection}{EC.\arabic{section}.\arabic{subsection}}

\providecommand{\theHsubsubsection}{}
\renewcommand{\theHsubsubsection}{EC.\arabic{section}.\arabic{subsection}.\arabic{subsubsection}}

\providecommand{\theHtheorem}{}
\renewcommand{\theHtheorem}{EC.\arabic{theorem}}

\providecommand{\theHlemma}{}
\renewcommand{\theHlemma}{EC.\arabic{lemma}}

\providecommand{\theHobservation}{}
\renewcommand{\theHobservation}{EC.\arabic{observation}}

\providecommand{\theHequation}{}
\renewcommand{\theHequation}{EC.\arabic{equation}}

\providecommand{\theHalgorithm}{}
\renewcommand{\theHalgorithm}{EC.\arabic{algorithm}}
\makeatother

%%% Main head for the e-companion
\ECHead{Electronic Companion}

\section{Supplementary Conventions}
\label{sec:ec-front-matter-supplement}
\label{sec:ec-framework-conventions}

Equip each $E_t$ and $A_t$ with sigma-algebras $\Sigma_t^E$ and $\Sigma_t^A$ and all history and trajectory spaces with the corresponding product sigma-algebras. A randomized policy $\bm\pi=(\pi_1,\pi_2,\ldots)$ consists of history-measurable stochastic kernels $\pi_t(\,\cdot\mid\bm e_{1:t},\bm a_{1:t-1})$ on $A_t$; after observing $e_t$, it samples $a_t$ from $\pi_t$. For each fixed admissible realized-sequence/prediction pair, these kernels and the stated simulation randomness determine the probability law for the expectations in Section~\ref{sec:framework-formulation}.

The regularity and measurable-selection assumptions of Section~\ref{sec:framework-formulation} apply throughout this companion. In particular, $d|_{E_t\times E_t}$ is $(\Sigma_t^E\otimes\Sigma_t^E)$-measurable for every $t$.

In Definition~\ref{def:eta-online-oracle}, $\bm\pi_h^{\mathcal I_h}$ is the restriction of $\bm\pi_h$ to the fixed history $\mathcal I_h=(\bm e_{1:h},\bm a_{1:h})$: for every $t>h$,
\[
(\bm\pi_h^{\mathcal I_h})_t
(\,\cdot\mid\bm e_{h+1:t},\bm a_{h+1:t-1})
\defeq
\pi_{h,t}(\,\cdot\mid
\bm e_{1:h}\circ\bm e_{h+1:t},
\bm a_{1:h}\circ\bm a_{h+1:t-1}).
\]

For any Monte Carlo switching test with fixed history $\mathcal I_h$ and finite continuation $\bm e'_{h+1:N}$, AdaSwitch has a measurable resettable simulator that, conditional on this query and all previously revealed randomness, generates any finite number of i.i.d.\ trajectories with the law of $\bm\pi_h^{\mathcal I_h}$. An online-oracle copy used to generate executed actions is a \emph{live copy}. Simulator seeds are mutually independent across trajectories and calls and independent of every live-copy seed. Simulation changes neither the state nor the random seed of any live copy. Only the approximate-oracle variants for reward and cost require this access assumption.

\section{Lipschitz Perturbation Bounds and Sequential Optimality}

\begin{observation}
\label{obs:lipschitz-telescoping}
    Consider any $m,n\in\mathbb{Z}_{\geq 0}$, $\mathcal{I}=(\bm{e}_{1:m}, \bm{a}_{1:m})$, $\bm{e}_{m+1:m+n},\bm{e}'_{m+1:m+n}\in\bm{E}_{m+1:m+n}$, and $\bm{a}_{m+1:m+n}\in\bm{A}_{m+1:m+n}$. We have:
    \begin{itemize}
        \item if $\mathcal{P}$ is $(u,v)$-Lipschitz, then
            \begin{align}
               \left|\mathrm{Opt}(\mathcal{P}^{\mathcal{I}},\bm{e}_{m+1:m+n})-\mathrm{Opt}(\mathcal{P}^{\mathcal{I}},\bm{e}'_{m+1:m+n})  \right|\leq L\sum_{j=m+1}^{m+n} \min\left(u\cdot d(e_j, e'_j),v\right);\notag
            \end{align}
        \item if $\mathcal{P}$ is $(u,v)$-strongly Lipschitz, then
             \begin{align}
                     \left|\mathrm{Val}(\mathcal{P}^{\mathcal{I}},\bm{e}_{m+1:m+n},\bm{a}_{m+1:m+n}) - \mathrm{Val}(\mathcal{P}^{\mathcal{I}},\bm{e}'_{m+1:m+n},\bm{a}_{m+1:m+n})\right| \leq  L\sum_{j=m+1}^{m+n}\min\left(u\cdot d(e_j, e'_j),v\right).\notag
             \end{align}
    \end{itemize}
\end{observation}

\begin{observation}
\label{obs:sequential-exact-optimality}
    Consider any $m\in\mathbb{Z}_{\geq 0}$, $\mathcal{I}=(\bm{e}_{1:m}, \bm{a}_{1:m})$, and $\bm{e}_{m+1:\infty}\in\bm{E}_{m+1:\infty}$. For each $t=m+1,m+2,\ldots$, after choosing the actions through period $t-1$, let $\mathcal{I}(t)=(\bm{e}_{1:t-1},\bm{a}_{1:t-1})$, choose any optimizer
    \[
    \bm a_{t:\infty}^{*,(t)}\in
    \arg\max_{\bm{a}''_{t:\infty}\in\bm A_{t:\infty}} \mathrm{Val}(\mathcal{P}^{\mathcal{I}(t)},\bm{e}_{t:\infty},\bm{a}''_{t:\infty}),
    \]
    and execute its first action $a_t=a_t^{*,(t)}$. Then
    \[
    \mathrm{Val}(\mathcal{P}^{\mathcal{I}},\bm{e}_{m+1:\infty},\bm{a}_{m+1:\infty}) =
       \mathrm{Opt}(\mathcal{P}^{\mathcal{I}},\bm{e}_{m+1:\infty}).
    \]
    Under the cost-minimization interpretation, the same conclusion holds when every $\arg\max$ above is replaced by $\arg\min$.
\end{observation}

\subsection{Proof of Observation~\ref{obs:lipschitz-telescoping}}
\label{sec:proof-obs-lipschitz-telescoping}
We prove the $(u,v)$-Lipschitz case; the strongly Lipschitz case uses the same telescoping argument, with the fixed-action one-coordinate inequality supplied directly by its definition. If $n=0$, both sides of each claimed inequality are zero. Henceforth assume $n\geq1$. We first reduce the claim to a one-coordinate bound. Define, for $q=0,\ldots,n$,
\[
\bm z^{(q)}_{m+1:m+n}
\defeq
\bm e'_{m+1:m+q}\circ\bm e_{m+q+1:m+n},
\]
using the empty-prefix and empty-suffix conventions. Then $\bm z^{(0)}=\bm e$, $\bm z^{(n)}=\bm e'$, and, for $q=1,\ldots,n$, the sequences $\bm z^{(q-1)}$ and $\bm z^{(q)}$ differ only in coordinate $m+q$. Applying the one-coordinate bound proved below in both directions and then the triangle inequality gives
\[
\left|\mathrm{Opt}(\mathcal P^{\mathcal I},\bm e_{m+1:m+n})
-\mathrm{Opt}(\mathcal P^{\mathcal I},\bm e'_{m+1:m+n})\right|
\leq L\sum_{q=1}^{n}\min\!\left(u\,d(e_{m+q},e'_{m+q}),v\right).
\]
It remains to prove this one-coordinate bound. Consider two sequences with a unique disagreement at coordinate $m+h$ for some $h\in[n]$. By symmetry, it suffices to show that
\[
\mathrm{Opt}(\mathcal{P}^{\mathcal{I}},\bm{e}_{m+1:m+n})\geq \mathrm{Opt}(\mathcal{P}^{\mathcal{I}},\bm{e}'_{m+1:m+n})-  L\cdot \min\left(u\cdot d(e_{m+h}, e'_{m+h}),v\right).
\]
Consider any sequence $\bm{a}^*_{m+1:m+n}$ such that
\[
\mathrm{Val}(\mathcal{P}^{\mathcal{I}},\bm{e}'_{m+1:m+n},\bm{a}^*_{m+1:m+n})=  \mathrm{Opt}(\mathcal{P}^{\mathcal{I}},\bm{e}'_{m+1:m+n}).
\]
Let $\widehat{\mathcal{I}}$ denote $(\bm{e}_{1:m+h-1},\bm{a}_{1:m}\circ \bm{a}^*_{m+1:m+h-1})$. We then have 
\begin{align}
    &\mathrm{Opt}(\mathcal{P}^{\mathcal{I}},\bm{e}_{m+1:m+n})  \geq \sum_{i=1}^{h-1} R_{i}^{\mathcal{I}}(\bm{e}_{m+1:m+i},\bm{a}^*_{m+1:m+i})+\mathrm{Opt}(\mathcal{P}^{\widehat{\mathcal{I}}},\bm{e}_{m+h:m+n}),\notag \\
    &\qquad\qquad\geq\sum_{i=1}^{h-1} R_{i}^{\mathcal{I}}(\bm{e}'_{m+1:m+i},\bm{a}^*_{m+1:m+i})+\mathrm{Opt}(\mathcal{P}^{\widehat{\mathcal{I}}},\bm{e}'_{m+h:m+n}) -L\cdot \min\left(u\cdot d(e_{m+h}, e'_{m+h}),v\right), \notag\\
    &\qquad\qquad=\mathrm{Opt}(\mathcal{P}^{\mathcal{I}},\bm{e}'_{m+1:m+n})- L\cdot \min\left(u\cdot d(e_{m+h}, e'_{m+h}),v\right),\notag
\end{align}
where the second inequality follows because $\mathcal{P}$ is $(u,v)$-Lipschitz and the equality follows from the optimality of $\bm{a}^*_{m+1:m+n}$: after fixing the prefix $\bm{a}^*_{m+1:m+h-1}$, the suffix $\bm{a}^*_{m+h:m+n}$ must be an optimal continuation under $\bm{e}'_{m+h:m+n}$.
\hfill\Halmos

\subsection{Proof of Observation~\ref{obs:sequential-exact-optimality}}
\label{sec:proof-obs-sequential-exact-optimality}
Let $M=M(\bm{e}_{1:\infty})$ be the effective horizon of $\bm{e}_{1:\infty}$. By admissibility, $M<\infty$. We prove the claim for every fixed history with request prefix $\bm e_{1:m}$ by backward induction on $m$.

\noindent\underline{Base case ($m\geq M$).} By the definition of the effective horizon, the continuation is payoff-irrelevant, so
\[
\mathrm{Val}(\mathcal{P}^{\mathcal{I}}, \bm{e}_{m+1:\infty},\bm{a}_{m+1:\infty}) =
       \mathrm{Opt}(\mathcal{P}^{\mathcal{I}},\bm{e}_{m+1:\infty})=0.
\]

\noindent\underline{Induction step ($m<M$).} Assume the observation holds for every such history of length $m+1$ and fix one of length $m$. By construction, $a_{m+1}$ is the first action of an optimal plan from $\mathcal I(m+1)$, so choose its optimal tail $\widehat{\bm{a}}_{m+2:\infty}$ such that
\begin{align}
    \mathrm{Val}(\mathcal{P}^{\mathcal{I}(m+1)}, \bm{e}_{m+1:\infty}, a_{m+1}\circ \widehat{\bm{a}}_{m+2:\infty}) = \mathrm{Opt}(\mathcal{P}^{\mathcal{I}(m+1)},\bm{e}_{m+1:\infty}) \label{eq:1-oracle-sequential-opt-choose-current-opt}
\end{align}
By the inductive hypothesis,
\begin{align}
\mathrm{Val}(\mathcal{P}^{\mathcal{I}(m+2)},\bm{e}_{m+2:\infty},\bm{a}_{m+2:\infty}) =
       \mathrm{Opt}(\mathcal{P}^{\mathcal{I}(m+2)},\bm{e}_{m+2:\infty}). \label{eq:1-oracle-sequential-opt-choose-future-opt}
\end{align}
Combining the two equalities gives
\begin{align}
    &\mathrm{Val}(\mathcal{P}^{\mathcal{I}(m+1)},\bm{e}_{m+1:\infty},\bm{a}_{m+1:\infty}) \notag\\
    &\qquad = R^{\mathcal{I}(m+1)}_1(e_{m+1}, a_{m+1})+ \mathrm{Val}(\mathcal{P}^{\mathcal{I}(m+2)},\bm{e}_{m+2:\infty},\bm{a}_{m+2:\infty})\notag \\
    &\qquad = R^{\mathcal{I}(m+1)}_1(e_{m+1}, a_{m+1})+\mathrm{Opt}(\mathcal{P}^{\mathcal{I}(m+2)},\bm{e}_{m+2:\infty})\notag  \\
    &\qquad \geq R^{\mathcal{I}(m+1)}_1(e_{m+1}, a_{m+1})+\mathrm{Val}(\mathcal{P}^{\mathcal{I}(m+2)},\bm{e}_{m+2:\infty}, \widehat{\bm{a}}_{m+2:\infty}) =\mathrm{Opt}(\mathcal{P}^{\mathcal{I}(m+1)},\bm{e}_{m+1:\infty}). \notag
\end{align}
Using Eqs.~\eqref{eq:1-oracle-sequential-opt-choose-future-opt} and~\eqref{eq:1-oracle-sequential-opt-choose-current-opt}, the displayed chain gives a value at least the optimum. Feasibility gives equality and completes the induction. For minimization, replace $\arg\max$ with $\arg\min$ and reverse the optimization inequalities.
\hfill\Halmos

\section{Supplementary Proofs for Section~\ref{sec:Adaptive-Switching-Algorithm-for-1-oracle}}
\label{sec:supp-exact-reward}

Throughout this section, $\mathcal P$ is an $(u,v)$-Lipschitz, $f$-bounded-influence reward-maximization problem, $b>0$, $b\geq u$, $c\geq\max\{v,f,1\}$, and $\hat d(e,e')\defeq\min\{d(e,e'),c/b\}$.

For the analysis, extend the finite plan in Eq.~\eqref{eq:adaswitch-1-predictive-follow} after $\widehat M_t$ using the standing measurable selection rule, and denote the extension by $\bm a^{*,(t)}_{t:\infty}$. The estimator guarantee makes it optimal for the full hybrid continuation.

\begin{lemma} \label{lem:predictive-following-infinite-reward}
Fix $m\in\mathbb Z_{\geq0}$, a history $\mathcal I=(\bm e_{1:m},\bm a_{1:m})$, a realized continuation $\bm e_{m+1:\infty}$, and an admissible request-sequence prediction $\bm e^*_{1:\infty}$ whose continuation differs from $\bm e_{m+1:\infty}$ in finitely many coordinates. For every $t\geq m+1$, execute the first action $a_t=a_t^{*,(t)}$ of the finite plan in Eq.~\eqref{eq:adaswitch-1-predictive-follow}. Then
    \begin{align}
       &\mathrm{Val}(\mathcal{P}^{\mathcal{I}},\bm{e}_{m+1:\infty},\bm{a}_{m+1:\infty})\geq
       \mathrm{Opt}(\mathcal{P}^{\mathcal{I}},\bm{e}_{m+1:\infty}) - 2 b L\sum_{i=m+1}^{\infty} \hat d(e_{i}, e^*_{i}).\notag 
    \end{align}
\end{lemma}

\begin{lemma}
\label{lem:predictive-following-finite-reward}
Fix $m,n\in\mathbb{Z}_{\geq 0}$, a history $\mathcal{I}=(\bm{e}_{1:m}, \bm{a}_{1:m})$, a realized continuation $\bm e_{m+1:\infty}$, and an admissible request-sequence prediction $\bm e^*_{1:\infty}$; their tails may disagree infinitely often. For every $t\in\{m+1,\ldots,m+n\}$, execute the first action $a_t=a_t^{*,(t)}$ of the finite plan in Eq.~\eqref{eq:adaswitch-1-predictive-follow}; this index set is empty when $n=0$. Then
    \begin{align}
       &\mathrm{Val}(\mathcal{P}^{\mathcal{I}},\bm{e}_{m+1:m+n},\bm{a}_{m+1:m+n}) \geq
       \mathrm{Opt}(\mathcal{P}^{\mathcal{I}},\bm{e}_{m+1:m+n}) - 2 bL\sum_{i=m+1}^{m+n} \hat d(e_{i}, e^*_{i})- c L. \notag
    \end{align}
\end{lemma}

\subsection{Proof of Lemma~\ref{lem:predictive-following-infinite-reward}}
\label{sec:proof-lem-predictive-following-infinite-reward}
Choose a finite endpoint
\[
N\geq\max\{m,M(\bm e_{1:m}\circ\bm e_{m+1:\infty}),M(\bm e_{1:m}\circ\bm e^*_{m+1:\infty})\}.
\]
Both continuation optima equal their truncations at $N$, so Observation~\ref{obs:lipschitz-telescoping}, $b\geq u$, and $c\geq v$ imply
\begin{equation}
\mathrm{Opt}(\mathcal P^{\mathcal I},\bm e_{m+1:\infty})
\leq \mathrm{Opt}(\mathcal P^{\mathcal I},\bm e^*_{m+1:\infty})
+bL\sum_{i=m+1}^{\infty}\hat d(e_i,e_i^*).
\label{eq:predictive-following-opt-comparison}
\end{equation}
It remains to prove
\begin{equation}
\mathrm{Val}(\mathcal P^{\mathcal I},\bm e_{m+1:\infty},\bm a_{m+1:\infty})
\geq \mathrm{Opt}(\mathcal P^{\mathcal I},\bm e^*_{m+1:\infty})
-bL\sum_{i=m+1}^{\infty}\hat d(e_i,e_i^*).
\label{eq:predictive-following-infinite-induction}
\end{equation}
By the effective-horizon estimator's guarantee, each finite plan in Eq.~\eqref{eq:adaswitch-1-predictive-follow} extends to an optimizer for its hybrid infinite continuation, as specified at the start of this section. We use induction on the finite number of disagreements. With no disagreement, each hybrid continuation equals the realized continuation, so Eq.~\eqref{eq:predictive-following-infinite-induction} follows from Observation~\ref{obs:sequential-exact-optimality}. Otherwise, let $h$ be the last disagreement and set $\widehat{\bm e}_{m+1:\infty}\defeq \bm e_{m+1:h-1}\circ\bm e^*_{h:\infty}$.
This continuation has one fewer disagreement. Let $\widehat{\bm a}$ be the trajectory generated on $\widehat{\bm e}$ by the first-action rule in Eq.~\eqref{eq:adaswitch-1-predictive-follow}, using the same standing measurable selection rule. The two trajectories agree before $h$. Define their common history explicitly by
\[
\mathcal I(h)\defeq
(\bm e_{1:m}\circ\bm e_{m+1:h-1},
\bm a_{1:m}\circ\bm a_{m+1:h-1}).
\]
Because $h$ is the last disagreement, each trajectory is generated by exact reoptimization on its realized continuation from period $h$ onward. Observation~\ref{obs:sequential-exact-optimality} therefore identifies the two generated suffix values with the optima for $\bm e_{h:\infty}$ and $\widehat{\bm e}_{h:\infty}$, respectively. Choose a finite endpoint
\[
N\geq\max\{h,M(\bm e_{1:m}\circ\bm e_{m+1:\infty}),M(\bm e_{1:m}\circ\widehat{\bm e}_{m+1:\infty})\}.
\]
The Lipschitz bound at this common truncation gives
\begin{align*}
&\mathrm{Val}(\mathcal P^{\mathcal I(h)},\bm e_{h:\infty},\bm a_{h:\infty})
-\mathrm{Val}(\mathcal P^{\mathcal I(h)},\widehat{\bm e}_{h:\infty},\widehat{\bm a}_{h:\infty})\notag\\
&\quad\geq-L\min\{u d(e_h,e_h^*),v\}
\geq-bL\hat d(e_h,e_h^*).
\end{align*}
Adding the common prefix and applying the induction hypothesis to $\widehat{\bm e}$ proves Eq.~\eqref{eq:predictive-following-infinite-induction}. Combining it with Eq.~\eqref{eq:predictive-following-opt-comparison} proves the lemma.
\hfill\Halmos

\subsection{Proof of Lemma~\ref{lem:predictive-following-finite-reward}}
\label{sec:proof-lem-predictive-following-finite-reward}
The case $n=0$ is immediate; assume $n\geq1$. Set
\[
\widetilde{\bm e}_{m+1:\infty}
\defeq\bm e_{m+1:m+n}\circ\bm e^*_{m+n+1:\infty},
\qquad
D_n\defeq\sum_{i=m+1}^{m+n}\hat d(e_i,e_i^*),
\]
and continue the generated action sequence after $m+n$ by applying the first-action rule in Eq.~\eqref{eq:adaswitch-1-predictive-follow} to the predicted tail. Denote the resulting trajectory by $\widetilde{\bm a}$ and its history through $m+n$ by $\mathcal J$. The full splice $\bm e_{1:m}\circ\widetilde{\bm e}_{m+1:\infty}$ has a finite effective horizon and differs from the prediction after $m$ in at most $n$ coordinates, so Lemma~\ref{lem:predictive-following-infinite-reward} applies. Let $\bm a^*_{m+1:m+n}$ optimize the finite interval, and let $\mathcal J^*$ be its induced history through $m+n$. The equality and successive inequalities below follow, respectively, from sequential exact optimality, Lemma~\ref{lem:predictive-following-infinite-reward}, feasible concatenation, and bounded influence with $f\leq c$:
\begin{align*}
&\mathrm{Val}(\mathcal P^{\mathcal I},\bm e_{m+1:m+n},\bm a_{m+1:m+n})
+\mathrm{Opt}(\mathcal P^{\mathcal J},\bm e^*_{m+n+1:\infty})\\
&\quad=\mathrm{Val}(\mathcal P^{\mathcal I},\widetilde{\bm e}_{m+1:\infty},\widetilde{\bm a}_{m+1:\infty})\\
&\quad\geq\mathrm{Opt}(\mathcal P^{\mathcal I},\widetilde{\bm e}_{m+1:\infty})-2bLD_n\\
&\quad\geq\mathrm{Opt}(\mathcal P^{\mathcal I},\bm e_{m+1:m+n})
+\mathrm{Opt}(\mathcal P^{\mathcal J^*},\bm e^*_{m+n+1:\infty})-2bLD_n\\
&\quad\geq\mathrm{Opt}(\mathcal P^{\mathcal I},\bm e_{m+1:m+n})
+\mathrm{Opt}(\mathcal P^{\mathcal J},\bm e^*_{m+n+1:\infty})-2bLD_n-cL.
\end{align*}
Cancelling the common tail optimum proves the lemma.
\hfill\Halmos

\subsection{Proof of Theorem~\ref{thm:ASA-error-dependent-1-oracle}}
\label{sec:proof-thm-ASA-error-dependent-1-oracle}

We first bound the reward earned in each epoch. For robustness, conservative-epoch surplus covers predictive and restart losses; for the error-dependent guarantee, each completed predictive epoch is charged to its accumulated clipped prediction error.

\noindent\underline{Epoch setup.}
Fix admissible realized and predicted sequences. Set
$M\defeq M(\bm e_{1:\infty})\geq1$ and
$\varphi_M^*\defeq\sum_{t=1}^{M}\hat d(e_t,e_t^*)$.
Let $\bm a$ be the action sequence generated by Algorithm~\ref{alg:Adaptive-Switching-Algorithm-for-1-oracle} and $\bm a^*_{1:M}$ an offline-optimal action sequence. Write
\[
\mathsf{ALG}\defeq
\mathrm{Val}(\mathcal P,\bm e_{1:\infty},\mathrm{AdaSwitch})
=\mathbb E\!\left[\mathrm{Val}(\mathcal P,\bm e_{1:\infty},\bm a_{1:\infty})\right],
\qquad
\mathsf{OPT}\defeq\mathrm{Opt}(\mathcal P,\bm e_{1:\infty}).
\]
For $1\leq t\leq M+1$, let
$\mathcal I(t)\defeq(\bm e_{1:t-1},\bm a_{1:t-1})$ and
$\mathcal I^*(t)\defeq(\bm e_{1:t-1},\bm a^*_{1:t-1})$.

Let $\{\mathcal G_t\}_{t\geq1}$ be the filtration generated by the algorithmic randomness revealed strictly before period $t$.
Set $s_1=1$. For each fixed index $i\geq1$, set $s_{i+1}=M+1$ when $s_i=M+1$; otherwise define
\begin{align}
Q_i(t)&\defeq
\begin{cases}
\mathrm{Opt}(\mathcal P^{\mathcal I(s_i)},\bm e_{s_i:t}),&2\nmid i,\\
\displaystyle\sum_{j=s_i}^{t}\hat d(e_j,e_j^*),&2\mid i,
\end{cases}
\qquad
\theta_i\defeq
\begin{cases}
\dfrac{10cL}{\epsilon},&2\nmid i,\\
\dfrac{2c}{\eta b},&2\mid i,
\end{cases}\notag\\
t_i&\defeq\inf\{t\in\{s_i,\ldots,M\}:Q_i(t)\geq\theta_i\},
\qquad s_{i+1}\defeq\min\{M+1,t_i+1\}, \label{eq:exact-oracle-capped-epoch-times}
\end{align}
where $\inf\varnothing\defeq+\infty$. Thus $s_{i+1}$ is the next epoch start, capped at $M+1$; if $s_{i+1}=M+1$, there is no subsequent payoff-relevant epoch.

Inductively, $s_i$ is a bounded stopping time; put $\mathcal F_i\defeq\mathcal G_{s_i}$. Conditional on $\mathcal F_i$, the starting history and fixed request sequences determine every threshold test, so Eq.~\eqref{eq:exact-oracle-capped-epoch-times} makes $s_{i+1}$ $\mathcal F_i$-measurable. Since $s_{i+1}\geq s_i$, it is also a stopping time: $\{s_{i+1}\leq t\}=\{s_{i+1}\leq t\}\cap\{s_i\leq t\}\in\mathcal G_t$. This completes the induction. Thus conditioning on $\mathcal F_i$ fixes both the starting history and the endpoint, allowing the fixed-endpoint online-oracle guarantee to apply.

Let $N\defeq\max\{i\geq1:s_i\leq M\}\leq M$. Epoch $i$ comprises periods $s_i,\ldots,s_{i+1}-1$ and is \emph{completed} if $i<N$, meaning that another epoch starts by $M$. For each fixed index $i\geq1$,
\begin{align*}
\{i\leq N\}=\{s_i\leq M\}\in\mathcal F_i,
\qquad
\{i<N\}=\{s_{i+1}\leq M\}\in\mathcal F_i.
\end{align*}
Write $V_i$ for AdaSwitch's epoch reward, $V_i^*$ for the global offline solution's reward on the same interval, $O_i$ for the interval optimum from AdaSwitch's history, and $D_i$ for the interval's clipped prediction error. Set these quantities to zero after epoch $N$:
\begin{align*}
V_i&\defeq
\mathbb I(i\leq N)\,
\mathrm{Val}(\mathcal P^{\mathcal I(s_i)},\bm e_{s_i:s_{i+1}-1},
\bm a_{s_i:s_{i+1}-1}),\\
V_i^*&\defeq
\mathbb I(i\leq N)\,
\mathrm{Val}(\mathcal P^{\mathcal I^*(s_i)},\bm e_{s_i:s_{i+1}-1},
\bm a^*_{s_i:s_{i+1}-1}),\\
O_i&\defeq
\mathbb I(i\leq N)\,
\mathrm{Opt}(\mathcal P^{\mathcal I(s_i)},\bm e_{s_i:s_{i+1}-1}),\\
D_i&\defeq
\mathbb I(i\leq N)\sum_{j=s_i}^{s_{i+1}-1}\hat d(e_j,e_j^*).
\end{align*}
Here $V_i^*$, $O_i$, and $D_i$ are $\mathcal F_i$-measurable. Pathwise, the epochs partition the effective horizon, so
\[
\sum_{i=1}^{\infty}V_i
=\mathrm{Val}(\mathcal P,\bm e_{1:\infty},\bm a_{1:\infty}),
\qquad
\sum_{i=1}^{\infty}V_i^*=\mathsf{OPT}.
\]

\noindent\underline{Epoch reward bounds.}
For an even (predictive) epoch, Lemma~\ref{lem:predictive-following-finite-reward} gives $\mathbb E[V_i\mid\mathcal F_i]\geq O_i-2bLD_i-cL\,\mathbb I(i\leq N)$, while bounded influence with $f\leq c$ gives $O_i\geq V_i^*-cL\,\mathbb I(i\leq N)$. The bound used for prediction-error dependence is therefore
\begin{align}
\mathbb E[V_i\mid\mathcal F_i]
\geq V_i^*-2bLD_i-2cL\,\mathbb I(i\leq N).
\label{eq:1-oracle-bigger-bl-error-cl}
\end{align}
For robustness, each clipped-error increment is at most $c/b$, so the exit threshold and its one-period overshoot give $D_i\leq3c\mathbb I(i\leq N)/(\eta b)$. Combining this with Eq.~\eqref{eq:1-oracle-bigger-bl-error-cl} and nonnegative rewards yields
\begin{align}
\mathbb E[V_i\mid\mathcal F_i]
&\geq\max\left\{0,V_i^*-\left(\frac{6c}{\eta}+2c\right)L\mathbb I(i\leq N)\right\}\notag\\
&\geq\eta V_i^*-(6+2\eta)cL\,\mathbb I(i\leq N)\notag\\
&\geq\eta V_i^*-8cL\,\mathbb I(i\leq N).
\label{eq:1-oracle-predictive-eta-epsilon-0}
\end{align}
The last two inequalities use $0<\eta\leq1$ and the zero extension $V_i^*=0$ on $\{i>N\}$.

For an odd (conservative) epoch, the policy-kernel construction implies that, conditional on $\mathcal F_i$, the executed actions have the law of $\bm\pi_{s_i-1}^{\mathcal I(s_i)}$ through the $\mathcal F_i$-measurable endpoint. For $i=1$ and $i>1$, respectively, the initial and restarted clauses of Definition~\ref{def:eta-online-oracle} therefore give
\begin{align}
\mathbb E[V_i\mid\mathcal F_i]
\geq\eta O_i-cL\,\mathbb I(2\leq i\leq N).
\label{eq:1-oracle-conservative-eta-epsilon-0}
\end{align}
The reference histories coincide for $i=1$; otherwise bounded influence gives $O_i\geq V_i^*-cL\mathbb I(2\leq i\leq N)$. For the error-dependent argument, this gives
\begin{align}
\mathbb E[V_i\mid\mathcal F_i]
\geq\eta V_i^*-(1+\eta)cL\,\mathbb I(2\leq i\leq N).
\label{eq:1-oracle-conservative-full-eta}
\end{align}
For robustness, a completed conservative epoch has $O_i\geq10cL/\epsilon$. Splitting $\eta=(\eta-\epsilon)+\epsilon$ in Eq.~\eqref{eq:1-oracle-conservative-eta-epsilon-0} yields
\begin{align}
\mathbb E[V_i\mid\mathcal F_i]
\geq(\eta-\epsilon)V_i^*
+10cL\,\mathbb I(i<N)
-2cL\,\mathbb I(2\leq i\leq N).
\label{eq:1-oracle-conservative-eta-epsilon-1}
\end{align}

\noindent\underline{Robustness.}
For each realization, set
\[
G\defeq\left\lfloor\frac{N-1}{2}\right\rfloor,\qquad
q\defeq\mathbb I(2\mid N),\qquad
F\defeq G+q=\left\lfloor\frac N2\right\rfloor.
\]
Here $G$ counts both completed predictive epochs and noninitial conservative epochs; $q$ indicates a final predictive epoch; and $F$ counts both predictive epochs and completed conservative epochs. Each completed conservative epoch supplies a $10cL$ surplus, covering an $8cL$ predictive loss and up to $2cL$ from a subsequent conservative epoch. Taking expectations and summing Eqs.~\eqref{eq:1-oracle-predictive-eta-epsilon-0} and~\eqref{eq:1-oracle-conservative-eta-epsilon-1}, using the tower property and $N\leq M$, gives
\begin{align}
\mathsf{ALG}
&\geq(\eta-\epsilon)\mathsf{OPT}
+cL\,\mathbb E[-8F+10F-2G]\notag\\
&=(\eta-\epsilon)\mathsf{OPT}
+2cL\,\mathbb E[q]
\geq(\eta-\epsilon)\mathsf{OPT}.
\label{eq:1-oracle-robustness}
\end{align}

\noindent\underline{Error-dependent bound.}
Let $S_{\mathrm c}^*$ denote the offline-optimal reward accrued during conservative epochs:
\[
S_{\mathrm c}^*\defeq\sum_{\substack{i\geq1\\2\nmid i}}V_i^*.
\]
Taking expectations and summing Eqs.~\eqref{eq:1-oracle-bigger-bl-error-cl} and~\eqref{eq:1-oracle-conservative-full-eta} over the epoch partition, using $F=G+q$ and disjoint predictive epochs, gives
\begin{align}
\mathsf{ALG}
\geq\mathsf{OPT}
-2bL\varphi_M^*
-\mathbb E\!\left[(1-\eta)S_{\mathrm c}^*
+cL\bigl((3+\eta)G+2q\bigr)\right].
\label{eq:1-oracle-error-aggregate-sharp}
\end{align}
To bound $S_{\mathrm c}^*$, remove the last period of each conservative epoch. The offline solution's prefix reward is at most the prefix optimum from its own history, which is at most the prefix optimum from AdaSwitch's epoch-start history plus $cL$ when $i>1$. The latter optimum is below $10cL/\epsilon$, and restoring the last period adds at most $L$. This also holds for a one-period epoch, whose prefix is empty. Thus
\begin{align*}
V_i^*
\leq L\left[
\left(\frac{10c}{\epsilon}+1\right)\mathbb I(i\leq N)
+c\,\mathbb I(2\leq i\leq N)
\right].
\end{align*}
There are $G+1$ conservative epochs. Each of the $G$ completed predictive epochs consumes at least $2c/(\eta b)$ clipped error. Summing the last bound and charging these disjoint error budgets gives, pathwise,
\[
S_{\mathrm c}^*\leq
L\left[
\frac{10c}{\epsilon}+1
+G\left(\frac{10c}{\epsilon}+c+1\right)
\right],
\qquad
G\leq\frac{\eta b\varphi_M^*}{2c},
\qquad q\leq1.
\]
Substitution in Eq.~\eqref{eq:1-oracle-error-aggregate-sharp} gives
\begin{align}
\mathsf{ALG}
\geq\mathsf{OPT}
-L\left[(1-\eta)\left(\frac{10c}{\epsilon}+1\right)+2c\right]
-2bL\varphi_M^*
-\frac{\eta bL\varphi_M^*}{2c}
\left[(1-\eta)\left(\frac{10c}{\epsilon}+c+1\right)+(3+\eta)c\right],
\label{eq:1-oracle-error-exact-sharp}
\end{align}

Because $c\geq1$ and $0<\epsilon<\eta\leq1$,
\begin{align*}
(1-\eta)\left(\frac{10c}{\epsilon}+1\right)+2c
&\leq\frac{10c}{\epsilon},\\
2+\frac{\eta}{2c}
\left[(1-\eta)\left(\frac{10c}{\epsilon}+c+1\right)+(3+\eta)c\right]
&\leq\frac{7\eta}{\epsilon}.
\end{align*}
Thus Eq.~\eqref{eq:1-oracle-error-exact-sharp} implies
\[
\mathsf{ALG}\geq\mathsf{OPT}
-\frac{L}{\epsilon}\left(10c+7b\eta\varphi_M^*\right).
\]
Dividing by $\mathsf{OPT}>0$ and combining this with Eq.~\eqref{eq:1-oracle-robustness} proves Eq.~\eqref{eq:ASA-error-dependent-1-oracle}.\hfill\Halmos

\subsection{\texorpdfstring{Proof of Corollary~\ref{cor:ASA-error-dependent-1-oracle-Rpre}}{Proof of the predicted-optimum corollary}}
\label{sec:proof-cor-ASA-error-dependent-1-oracle-Rpre}
Let $R\defeq\mathrm{Opt}(\mathcal{P},\bm{e}_{1:\infty})$, $R^*\defeq\mathrm{Opt}(\mathcal{P},\bm{e}^*_{1:\infty})>0$, and define $x\defeq bL\varphi_\vee^*/R^*$ and $y_{\mathrm{sw}}\defeq cL/R^*$. Both infinite-horizon optima equal their optima truncated at $M^\vee$. Observation~\ref{obs:lipschitz-telescoping} on this common finite horizon therefore gives
\[
R\geq R^*-bL\varphi_\vee^*=R^*(1-x),
\]
and $\varphi_M^*\leq\varphi_\vee^*$. If the proposed error-dependent branch in Eq.~\eqref{eq:ASA-error-dependent-1-oracle-Rpre} is at most $\eta-\epsilon$, the robustness bound in Theorem~\ref{thm:ASA-error-dependent-1-oracle} proves the corollary. Otherwise,
\[
10y_{\mathrm{sw}}+8\eta x<\epsilon(1-\eta+\epsilon)<\epsilon<\eta,
\]
where the second inequality uses $\epsilon<\eta$. In particular, $x<1$, and
\[
(10y_{\mathrm{sw}}+8\eta x)(1-x)-(10y_{\mathrm{sw}}+7\eta x)
=x\bigl(\eta-(10y_{\mathrm{sw}}+8\eta x)\bigr)\geq0.
\]
Consequently,
\[
\frac{L(10c+7b\eta\varphi_M^*)}{\epsilon R}
\leq \frac{10y_{\mathrm{sw}}+7\eta x}{\epsilon(1-x)}
\leq \frac{10y_{\mathrm{sw}}+8\eta x}{\epsilon},
\]
which converts Eq.~\eqref{eq:ASA-error-dependent-1-oracle} into Eq.~\eqref{eq:ASA-error-dependent-1-oracle-Rpre}.\hfill\Halmos

\subsection{Adjustable thresholds for exact-oracle reward AdaSwitch}
\label{sec:adjustable-thresholds-1-oracle}

Algorithm~\ref{alg:Adaptive-Switching-Algorithm-for-1-oracle} corresponds to the dimensionless entry and exit multipliers $\tau_{\mathrm{ent}}=10/\epsilon$ and $D_{\mathrm{exit}}=2$. Let $\mathrm{AdaSwitch}(\tau_{\mathrm{ent}},D_{\mathrm{exit}})$ denote the same algorithm with
\[
\Theta_{\mathrm{ent}}=\tau_{\mathrm{ent}}cL,
\qquad
\Theta_{\mathrm{exit}}=\frac{D_{\mathrm{exit}}c}{\eta b},
\]
where $\tau_{\mathrm{ent}},D_{\mathrm{exit}}>0$; $\epsilon$ is omitted from the inputs, and all other operations are unchanged.

\begin{proposition}[Adjustable thresholds for exact-oracle reward AdaSwitch]
\label{prop:ASA-adjustable-thresholds-1-oracle}
Under the assumptions of Theorem~\ref{thm:ASA-error-dependent-1-oracle} other than the slack parameter $\epsilon$, and with the thresholds replaced as above, run $\mathrm{AdaSwitch}(\tau_{\mathrm{ent}},D_{\mathrm{exit}})$ and define
\[
C_{\mathrm{exit}}\defeq\frac{2D_{\mathrm{exit}}}{\eta}+5,
\qquad
r(\tau_{\mathrm{ent}},D_{\mathrm{exit}})
\defeq
\left[
\frac{\tau_{\mathrm{ent}}\eta-1}{\tau_{\mathrm{ent}}+C_{\mathrm{exit}}}
\right]_+,
\]
\[
A_{\mathrm{adj}}(\tau_{\mathrm{ent}})
\defeq c(1-\eta)\tau_{\mathrm{ent}}+(1-\eta+2c),
\]
and
\[
B_{\mathrm{adj}}(\tau_{\mathrm{ent}},D_{\mathrm{exit}})
\defeq
2+\frac{\eta(1-\eta)\tau_{\mathrm{ent}}}{D_{\mathrm{exit}}}
+\frac{\eta}{D_{\mathrm{exit}}}
\left(4+\frac{1-\eta}{c}\right).
\]
Then $\mathrm{AdaSwitch}(\tau_{\mathrm{ent}},D_{\mathrm{exit}})$ satisfies
\begin{align}
\mathrm{Comp}
&\left(\mathcal P,\bm e_{1:\infty},\bm e^*_{1:\infty},
\mathrm{AdaSwitch}(\tau_{\mathrm{ent}},D_{\mathrm{exit}})\right)\notag\\
&\geq
\max\left\{
r(\tau_{\mathrm{ent}},D_{\mathrm{exit}}),
1-
\frac{L\left[
A_{\mathrm{adj}}(\tau_{\mathrm{ent}})
+bB_{\mathrm{adj}}(\tau_{\mathrm{ent}},D_{\mathrm{exit}})\varphi_M^*
\right]}
{\mathrm{Opt}(\mathcal P,\bm e_{1:\infty})}
\right\}.
\label{eq:ASA-adjustable-thresholds-1-oracle}
\end{align}
Here $r$ is the robustness bound; within the bracketed loss, $A_{\mathrm{adj}}$ is independent of prediction error and $bB_{\mathrm{adj}}$ is the coefficient of $\varphi_M^*$.
\end{proposition}

\noindent\emph{Proof.}
For the adjusted algorithm, use the capped-epoch construction and notation of Section~\ref{sec:proof-thm-ASA-error-dependent-1-oracle}, with thresholds $\Theta_{\mathrm{ent}}$ and $\Theta_{\mathrm{exit}}$ above. The same conditioning argument and zero-extension convention apply. The exit threshold and one-period overshoot bound the error in each predictive epoch. Each of the $G$ completed predictive epochs accumulates at least $\Theta_{\mathrm{exit}}$ clipped error on a disjoint interval. Hence
\[
D_i\leq\left(\frac{D_{\mathrm{exit}}}{\eta}+1\right)\frac cb\mathbb I(i\leq N)
\quad(2\mid i),
\qquad
G\leq\frac{\eta b\varphi_M^*}{D_{\mathrm{exit}}c}.
\]
There are $F$ completed conservative epochs, $F$ predictive epochs, and $G$ noninitial conservative epochs, with $F=G+q$. For any $r\in[0,\eta]$, split $\eta=r+(\eta-r)$ in the conservative bound. Taking expectations and summing the adjusted epoch bounds gives
\[
\mathsf{ALG}\geq r\mathsf{OPT}+cL\,\mathbb E\!\left[
F\left\{\tau_{\mathrm{ent}}(\eta-r)-r\left(\frac{2D_{\mathrm{exit}}}{\eta}+4\right)\right\}-(1+r)G
\right].
\]
Since $G\leq F$, the right-hand side is at least $r\mathsf{OPT}$ whenever
\begin{equation}
\label{eq:ASA-adjustable-robustness-condition}
\tau_{\mathrm{ent}}(\eta-r)
\geq1+r\left(\frac{2D_{\mathrm{exit}}}{\eta}+5\right).
\end{equation}
When $\tau_{\mathrm{ent}}\eta>1$, solving this condition at equality gives the positive expression defining $r(\tau_{\mathrm{ent}},D_{\mathrm{exit}})$; otherwise nonnegative rewards give the bound $r=0$.

For the error-dependent bound, Eq.~\eqref{eq:1-oracle-error-aggregate-sharp} remains valid, while the adjusted entry threshold, the one-period reward bound, and bounded influence give
\[
S_{\mathrm c}^*\leq L\left[\tau_{\mathrm{ent}}c+1
+G(\tau_{\mathrm{ent}}c+c+1)\right].
\]
Substituting the bounds on $S_{\mathrm c}^*$ and $G$, together with $q\leq1$, into Eq.~\eqref{eq:1-oracle-error-aggregate-sharp} yields the coefficients $A_{\mathrm{adj}}$ and $bB_{\mathrm{adj}}$, hence the second bound in Eq.~\eqref{eq:ASA-adjustable-thresholds-1-oracle}. Dividing by the positive realized optimum completes the proof.\hfill\Halmos

For a target robustness $r\in(0,\eta)$ and a fixed $D_{\mathrm{exit}}>0$, Eq.~\eqref{eq:ASA-adjustable-robustness-condition} is equivalent to
\begin{equation*}
\tau_{\mathrm{ent}}
\geq
\tau_{\mathrm{ent}}^{\min}(r,D_{\mathrm{exit}})
\defeq
\frac{1+r(2D_{\mathrm{exit}}/\eta+5)}{\eta-r}.
\end{equation*}
Equality gives the smallest entry multiplier satisfying the displayed sufficient condition and makes the robustness bound equal to $r$; larger entry multipliers give a bound of at least $r$.

For the canonical thresholds and $\epsilon=\eta-r$, the error-dependent branch in Theorem~\ref{thm:ASA-error-dependent-1-oracle} exceeds $r$ exactly when
\[
\frac{L(10c+7b\eta\varphi_M^*)}
{\mathrm{Opt}(\mathcal P,\bm e_{1:\infty})}
<(1-r)(\eta-r).
\]
Under perfect predictions, this reduces to $\mathrm{Opt}(\mathcal P,\bm e_{1:\infty})>10cL/[(1-r)(\eta-r)]$.

\section{Supplementary Proofs for Section~\ref{sec:ASA-for-gamma-oracle}}
\label{sec:proof-of-theorem-ASA-error-dependent-gamma}
Throughout, let $\mathcal P$ be $(u,v)$-strongly Lipschitz and $f$-bounded-influence, let $b>0$, $b\geq u$, and $c\geq\max\{v,f,1\}$, and set $\lambda\defeq\min\{1,L\}$.

\subsection{Proof of Theorem~\ref{thm:ASA-error-dependent-gamma-oracle}}
Each bound follows from one of the next two lemmas.

\begin{lemma}
\label{lem:gamma-oracle-eta-epsilon}
Under the setting of Theorem~\ref{thm:ASA-error-dependent-gamma-oracle}, AdaSwitch with a $\gamma$-approximate offline oracle and an $\eta$-online oracle is $(\eta-\frac{19}{20}\epsilon)$-robust.
\end{lemma}
\begin{lemma}
\label{lem:gamma-oracle-error-dependent}
Under the setting of Theorem~\ref{thm:ASA-error-dependent-gamma-oracle}, let $M=M(\bm e_{1:\infty})$ and $\varphi_M^*=\sum_{t=1}^{M}\hat d(e_t,e_t^*)$. AdaSwitch with a $\gamma$-approximate offline oracle and an $\eta$-online oracle has an instancewise performance ratio of at least
\[
\frac{\gamma\alpha}{\alpha+\gamma}
-\frac{L}{\epsilon\cdot\mathrm{Opt}(\mathcal{P},\bm{e}_{1:\infty})}
\left(17\alpha c+\frac{17}{5}b\eta\varphi_M^*\right).
\]
\end{lemma}

\subsection{Proof of Lemma~\ref{lem:gamma-oracle-eta-epsilon}}
\label{sec:proof-of-lemma-gamma-oracle-eta-epsilon}
We first bound each predictive epoch's loss, then cover it with the expected surplus from the preceding conservative epoch. The remaining Monte Carlo errors are summable and account for the final robustness slack.

\medskip
\noindent\underline{Epoch setup.}
Fix the realized and predicted sequences, let $M=M(\bm e_{1:\infty})\geq1$, and put $\varphi_M^*\defeq\sum_{t=1}^{M}\hat d(e_t,e_t^*)$. Let $\bm a$ be the random AdaSwitch action sequence and $\bm a^*$ an offline-optimal action sequence. As in the exact-oracle proof, write $\mathcal I(t)$ and $\mathcal I^*(t)$ for their histories and set
\[
\mathsf{ALG}\defeq
\mathbb E\!\left[\mathrm{Val}(\mathcal P,\bm e_{1:\infty},\bm a_{1:\infty})\right],
\qquad
\mathsf{OPT}\defeq\mathrm{Opt}(\mathcal P,\bm e_{1:\infty}),
\]
and
\[
\rho\defeq\frac{\gamma\alpha}{\alpha+\gamma},
\qquad
\delta\defeq\eta-\frac{15}{16}\epsilon,
\qquad
\kappa\defeq\frac{\rho}{\delta}\geq1.
\]

Set
\[
\Theta_{\mathrm c}\defeq\frac{16\eta\alpha cL}{\epsilon},
\qquad
\Theta_{\mathrm p}\defeq\frac{5\kappa\alpha c}{b}.
\]
Set $s_1=1$. For a conservative epoch beginning at $s_i$, write $\widehat R_{i,r}$ for the empirical estimate computed in period $r$ by Line~\ref{line:adaswitch-gamma-monte-carlo}, using the fixed history $\mathcal I(s_i)$.
The conservative endpoint remains random at epoch entry; we need it to be conditionally independent of the live oracle trajectory. On a product probability space, couple the executed conservative actions to the prefix of an uninterrupted online-oracle continuation $\bm a^{\mathrm{on},i}_{s_i:\infty}$ from $\mathcal I(s_i)$. Assign independent Monte Carlo seed arrays to all potential calendar periods, independently of the live oracle seeds. Reveal array $r$ only after the executed action in that period. The unrevealed array and fixed-history query $(\mathcal I(s_i),\bm e_{s_i:r})$ define $\widehat R_{i,r}$ even if the call is not reached; the algorithm observes the array only if the call occurs. Let $\{\mathcal G_t\}$ be the filtration generated by randomness revealed before period $t$. With $\inf\varnothing\defeq+\infty$, recursively define
\begin{align}
s_{i+1}\defeq
\begin{cases}
M+1,&s_i=M+1,\\[2pt]
\displaystyle\min\!\left\{M+1,1+\inf\left\{r\in\{s_i,\ldots,M\}:\widehat R_{i,r}\geq\Theta_{\mathrm c}\right\}\right\},
&s_i\leq M,~2\nmid i,\\[5pt]
\displaystyle\min\!\left\{M+1,1+\inf\left\{r\in\{s_i,\ldots,M\}:\sum_{k=s_i}^{r}\hat d(e_k,e_k^*)\geq\Theta_{\mathrm p}\right\}\right\},
&s_i\leq M,~2\mid i.
\end{cases} \label{eq:gamma-capped-epoch-times}
\end{align}
The $+1$ records that a test after period $r$ changes the mode at $r+1$; a switch after $M$ creates no payoff-relevant epoch. Inductively, $\{s_{i+1}\leq t\}$ is determined by tests completed by $t-1$, so every $s_i$ is a bounded $\{\mathcal G_t\}$-stopping time. Define $\mathcal F_i\defeq\mathcal G_{s_i}$ and
\[
N\defeq\max\{i\geq1:s_i\leq M\}.
\]
Then $N<\infty$, $\{i\leq N\}=\{s_i\leq M\}\in\mathcal F_i$, and $\{i<N\}=\{s_{i+1}\leq M\}$. Predictive endpoints are $\mathcal F_i$-measurable because the clipped-error sequence is fixed. Conditional on $\mathcal F_i$, conservative endpoints depend only on future Monte Carlo arrays and are independent of the uninterrupted online-oracle continuation. For $i>N$, use $s_i=s_{i+1}=M+1$. Define the zero-extended quantities
\[
\begin{aligned}
V_i&\defeq\mathbb I(i\leq N)\,
\mathrm{Val}\left(\mathcal P^{\mathcal I(s_i)},\bm e_{s_i:s_{i+1}-1},
\bm a_{s_i:s_{i+1}-1}\right),\\
V_i^*&\defeq\mathbb I(i\leq N)\,
\mathrm{Val}\left(\mathcal P^{\mathcal I^*(s_i)},\bm e_{s_i:s_{i+1}-1},
\bm a^*_{s_i:s_{i+1}-1}\right),\\
O_i&\defeq\mathbb I(i\leq N)\,
\mathrm{Opt}\left(\mathcal P^{\mathcal I(s_i)},\bm e_{s_i:s_{i+1}-1}\right),\\
D_i&\defeq\mathbb I(i\leq N)
\sum_{k=s_i}^{s_{i+1}-1}\hat d(e_k,e_k^*).
\end{aligned}
\]
The epochs partition $[M]$, so pathwise $\mathsf{OPT}=\sum_iV_i^*$ and, by Tonelli,
\[
\mathsf{ALG}=\sum_{i=1}^{\infty}\mathbb E[V_i].
\]

\medskip
\noindent\underline{Predictive epochs.}
Fix an even $i\leq N$, let $N_i$ be the number of executed batches in the epoch, and enumerate their start periods as $s_i^1=s_i<s_i^2<\cdots<s_i^{N_i}$. Set $s_i^{N_i+1}=s_{i+1}$. For batch $j\in[N_i]$, write $[q,r]=[s_i^j,s_i^{j+1}-1]$ and put $D_{q:r}\defeq\sum_{k=q}^r\hat d(e_k,e_k^*)$. Since $\widetilde e_q^{(q)}=e_q$, the Lipschitz loss from replacing hybrid requests by realized requests is at most $bLD_{q:r}$.

If $j<N_i$, the batch reaches its planned endpoint, so its hybrid value is at least both $\gamma$ times the hybrid optimum and $\alpha cL$. The inequality $\max\{\gamma x,\alpha cL\}\geq\rho(x+cL)$ for $x\geq0$ supplies the $cL$ allowance needed to compare histories. Applying the strongly Lipschitz bound to the fixed plan, the Lipschitz bound to the optimum, and bounded influence to the history change gives
\begin{align*}
\mathrm{Val}(\mathcal P^{\mathcal I(q)},\bm e_{q:r},\bm a_{q:r})
&\geq \mathrm{Val}(\mathcal P^{\mathcal I(q)},\widetilde{\bm e}^{(q)}_{q:r},\bm a_{q:r})-bLD_{q:r}\\
&\geq \rho\!\left(\mathrm{Opt}(\mathcal P^{\mathcal I(q)},\widetilde{\bm e}^{(q)}_{q:r})+cL\right)-bLD_{q:r}\\
&\geq \rho\,\mathrm{Val}(\mathcal P^{\mathcal I^*(q)},\bm e_{q:r},\bm a^*_{q:r})-2bLD_{q:r}.
\end{align*}

For the last batch, which may be truncated by the exit rule or by $M$, again write $[q,r]$. If the construction loop first returns a candidate of value at least $\alpha cL$ at endpoint $h$, the preceding candidate has value below $\alpha cL$. Its approximation guarantee bounds the hybrid optimum through $h-1$ by $\alpha cL/\gamma$ (the prefix is empty when $h=q$). Since $r\leq h$, monotonicity of the optimum and the one-period reward bound give the bound below.

If the loop instead exhausts $\widehat M_q$ while the candidate's value remains below $\alpha cL$, the candidate's approximation guarantee bounds the hybrid optimum through $\widehat M_q$ by $\alpha cL/\gamma$. Monotonicity covers earlier endpoints, and the estimator makes the later hybrid tail payoff-irrelevant. Thus, in both cases,
\[
\mathrm{Opt}(\mathcal P^{\mathcal I(q)},\widetilde{\bm e}^{(q)}_{q:r})
\leq\frac{\alpha cL}{\gamma}+L.
\]
Bounded influence and the Lipschitz bound therefore give
\begin{align*}
\gamma\,\mathrm{Val}(\mathcal P^{\mathcal I^*(q)},\bm e_{q:r},\bm a^*_{q:r})
&\leq\alpha cL+\gamma(1+c)L+\gamma bLD_{q:r}\\
&\leq2\alpha cL+bLD_{q:r},
\end{align*}
where $\gamma\leq1$ and $\gamma(1+c)\leq\alpha c$ because $\alpha\geq3$ and $c\geq1$. Since its realized reward is nonnegative and $\rho\leq\gamma$, the last batch satisfies the same $\rho$-benchmark bound as earlier batches, with an additional loss of $2\alpha cL$. Summing over all batches and zero-extending gives
\begin{align}
V_i
&\geq \rho V_i^*-2bLD_i
-2\alpha cL\,\mathbb I(i\leq N).
\label{eq:gamma-oracle-predictive-total-gamma-alpha-alpha1}
\end{align}
The one-period overshoot is at most $c/b$, so
\[
D_i\leq\frac{(5\kappa\alpha+1)c}{b}\,
\mathbb I(i\leq N).
\]
Since $2(5\kappa\alpha+1)+2\alpha<13\kappa\alpha$ for $\kappa\geq1$ and $\alpha\geq3$, Eq.~\eqref{eq:gamma-oracle-predictive-total-gamma-alpha-alpha1} and $V_i\geq0$ give
\begin{align}
V_i
\geq\max\left\{0,\rho V_i^*
-13\kappa\alpha cL\,\mathbb I(i\leq N)\right\}
\geq\delta V_i^*
-13\alpha cL\,\mathbb I(i\leq N).
\label{eq:gamma-oracle-predictive-total-eta}
\end{align}

The last step uses $\rho=\kappa\delta$ and $\max\{0,\kappa x\}\geq x$ for $\kappa\geq1$.

\medskip
\noindent\underline{Conservative epochs.}
Fix an odd $i$ and condition on $\mathcal F_i$ and $\{i\leq N\}$. For $n\in\{s_i,\ldots,M\}$, let
\[
w_{i,n}\defeq\Pr(s_{i+1}=n+1\mid\mathcal F_i).
\]
These weights sum to one. By the construction above, conditioning further on $s_{i+1}=n+1$ leaves the law of $\bm a^{\mathrm{on},i}_{s_i:n}$ unchanged. Averaging the initial ($i=1$) or restarted ($i>1$) oracle bound over $w_{i,n}$ and using $f\leq c$ gives
\begin{align}
\mathbb E[V_i\mid\mathcal F_i]
&\geq\mathbb E\left[
\eta O_i-cL\,\mathbb I(2\leq i\leq N)
\middle|\mathcal F_i\right].
\label{eq:gamma-si-si1-conservative-i-neq-N-0}
    \end{align}
Because $\delta<\eta$, the fraction $1-\delta/\eta=15\epsilon/(16\eta)$ of the conservative reward provides a surplus over the $\delta$ benchmark. Bounded influence and Eq.~\eqref{eq:gamma-si-si1-conservative-i-neq-N-0} give
\begin{align}
\mathbb E[V_i-\delta V_i^*\mid\mathcal F_i]
&\geq\mathbb E\left[
V_i-\delta O_i-cL\,\mathbb I(2\leq i\leq N)
\middle|\mathcal F_i\right]\notag\\
&\geq \mathbb E\left[
 \frac{15\epsilon}{16\eta}V_i
-2cL\,\mathbb I(2\leq i\leq N)
\middle|\mathcal F_i\right]. \label{eq:gamma-oracle-si-si1-conservative-i-eq-N}
\end{align}
The restart and history-comparison losses each contribute at most $cL$. The following lemma, proved in Section~\ref{sec:proof-lem-gamma-oracle-many-sample}, controls the reward at the random endpoint. We use its lower bound here and its upper bound in Section~\ref{sec:proof-lem-gamma-oracle-error-dependent}.
\begin{lemma}
\label{lem:gamma-oracle-many-sample}
Using the same notation, for every odd $i\geq1$,
\begin{align}
&\mathbb E\left[V_i- \mathbb I(i<N)\left(\frac{16\eta\alpha c L}{\epsilon}-\lambda\right)\middle|\mathcal F_i\right]\geq -\frac{\epsilon\lambda}{160 i^2}, \notag \\
&\mathbb E\left[V_i-\mathbb I(i\leq N)\frac{17\eta\alpha c L}{\epsilon}\middle|\mathcal F_i \right]\leq 0.\notag
\end{align}
\end{lemma}
Substituting the first estimate into Eq.~\eqref{eq:gamma-oracle-si-si1-conservative-i-eq-N}, taking expectations, and using $15\epsilon/(16\eta)\leq1$ gives
\begin{equation}
\mathbb E[V_i-\delta V_i^*]
\geq \mathbb E\left[(15\alpha cL-\lambda)\cdot\mathbb I(i<N)-2cL\cdot\mathbb I(2\leq i\leq N)\right]-\frac{\epsilon\lambda}{160\cdot i^2} \label{eq:gamma-oracle-si-si1-conservative-i-eq-N-1}.
\end{equation}

\medskip
\noindent\underline{Combining epoch bounds.}
In expectation, the $15\alpha cL$ surplus from each completed conservative epoch covers the $13\alpha cL$ loss of the following predictive epoch, leaving room for the restart and history-comparison terms and the $\lambda$ allowance from sampling.
For a realization with $N$ epochs, set
\[
F\defeq\left\lfloor\frac N2\right\rfloor,
\qquad
G\defeq\left\lfloor\frac{N-1}{2}\right\rfloor.
\]
There are $F$ predictive epochs, $F$ completed conservative epochs, and $G$ noninitial conservative epochs. Combining Eqs.~\eqref{eq:gamma-oracle-predictive-total-eta} and~\eqref{eq:gamma-oracle-si-si1-conservative-i-eq-N-1} gives
\begin{align}
\mathsf{ALG}
\geq\delta\,\mathsf{OPT}
-\sum_{\substack{i\geq1\\2\nmid i}}\frac{\epsilon\lambda}{160i^2}
+\mathbb E\!\left[(2\alpha cL-\lambda)F-2cLG\right].\notag
\end{align}
Since $G\leq F$, $\alpha\geq3$, $c\geq1$, and $\lambda\leq L$, the last expectation is nonnegative. Also $\sum_{\substack{i\geq1\\2\nmid i}}i^{-2}<2$, so
\[
\mathsf{ALG}
\geq
\max\left\{0,
\delta\,\mathsf{OPT}
-\frac{\epsilon\lambda}{80}\right\}.
\]
If $\mathsf{OPT}\geq\lambda$, use $\delta-\epsilon/80=\eta-19\epsilon/20$ to obtain $\mathsf{ALG}\geq(\eta-19\epsilon/20)\mathsf{OPT}$. If $\mathsf{OPT}<\lambda$, no conservative-to-predictive switch can occur: before the first such switch, every Monte Carlo sample is the value of a feasible trajectory, so each such sample $Z$ satisfies
\[
Z\leq\mathrm{Opt}(\mathcal P,\bm e_{1:t})
\leq\mathsf{OPT}
<\lambda.
\]
Thus every empirical average is below $\lambda<\Theta_{\mathrm c}$, so the entry test is never satisfied. AdaSwitch follows $\bm\pi_0$ throughout and obtains $\mathsf{ALG}\geq\eta\mathsf{OPT}$. This proves the lemma.\hfill\Halmos

\subsection{Proof of Lemma~\ref{lem:gamma-oracle-error-dependent}}
\label{sec:proof-lem-gamma-oracle-error-dependent}
Retain the preceding proof's notation. We combine epoch bounds, absorb sampling slack into a fixed loss, and control the number of completed predictive epochs by cumulative prediction error.

\medskip
\noindent\underline{Conservative-epoch bounds.}
For odd $i$, the first inequality of Lemma~\ref{lem:gamma-oracle-many-sample} gives, after taking expectations, the reward bound
\begin{align}
\mathbb E[V_i]
\geq
\mathbb E\left[(\Theta_{\mathrm c}-\lambda)\mathbb I(i<N)\right]
-\frac{\epsilon\lambda}{160i^2}.
\label{eq:gamma-oracle-bigger-gamma-minus-sth-0}
\end{align}
For the offline benchmark, Eq.~\eqref{eq:gamma-si-si1-conservative-i-neq-N-0}, the second inequality of Lemma~\ref{lem:gamma-oracle-many-sample}, and bounded influence give
\begin{align}
\mathbb E[V_i^*\mid\mathcal F_i]
\leq
\mathbb E\left[
\frac{17\alpha cL}{\epsilon}\mathbb I(i\leq N)
+\frac{2cL}{\eta}\mathbb I(2\leq i\leq N)
\middle|\mathcal F_i\right].
\label{eq:gamma-oracle-bigger-gamma-minus-sth-2}
\end{align}
Indeed, Eq.~\eqref{eq:gamma-si-si1-conservative-i-neq-N-0} gives $\mathbb E[O_i\mid\mathcal F_i]\leq\mathbb E[V_i\mid\mathcal F_i]/\eta+cL\mathbb I(2\leq i\leq N)/\eta$. Apply the reward upper bound in Lemma~\ref{lem:gamma-oracle-many-sample} and bounded influence. The histories coincide for $i=1$; otherwise the restart and history-comparison terms sum to at most $(1+1/\eta)cL\leq2cL/\eta$.

\medskip
\noindent\underline{Combining epoch bounds.}
Let $S$ denote the accumulated concentration slack:
\[
S\defeq\sum_{\substack{i\geq1\\2\nmid i}}\frac{\epsilon\lambda}{160i^2}
<\frac{\epsilon\lambda}{80}\leq\lambda.
\]
There are $F$ predictive epochs in total, $G$ of them completed; there are $G+1$ conservative epochs in total, $F$ of them completed and $G$ noninitial. Take expectations in Eq.~\eqref{eq:gamma-oracle-bigger-gamma-minus-sth-2} and subtract $\rho$ times this bound from Eq.~\eqref{eq:gamma-oracle-bigger-gamma-minus-sth-0} to bound each conservative contribution $\mathbb E[V_i-\rho V_i^*]$. Summing with the predictive bound~\eqref{eq:gamma-oracle-predictive-total-gamma-alpha-alpha1} and using $\sum_{2\mid i}D_i\leq\varphi_M^*$ gives
\begin{align*}
\mathsf{ALG}
\geq \rho\,\mathsf{OPT}-2bL\varphi_M^*
+\mathbb E\left[(\Theta_{\mathrm c}-\lambda-2\alpha cL)F\right]
-\frac{17\rho\alpha cL}{\epsilon}\mathbb E[G+1]
-\frac{2\rho cL}{\eta}\mathbb E[G]-S.
\end{align*}
The $F$-term is nonnegative: $\Theta_{\mathrm c}-\lambda-2\alpha cL\geq14\alpha cL-\lambda\geq0$, since $\epsilon<\eta$, $\alpha\geq3$, $c\geq1$, and $\lambda\leq L$.
The fixed additive allowance absorbs $S$. Indeed, $\gamma\leq1$ gives $\rho\leq\alpha/(\alpha+1)$. Together with $\alpha\geq3$, $\epsilon<\eta\leq1$, $c\geq1$, and $\lambda\leq L$, this yields $17(1-\rho)\alpha cL/\epsilon\geq\lambda\geq S$, hence
\[
\frac{17\rho\alpha cL}{\epsilon}+S
\leq\frac{17\alpha cL}{\epsilon}.
\]
Dropping the nonnegative $F$-term and using $\mathbb E[G+1]=\mathbb E[G]+1$ therefore gives
\[
\mathsf{ALG}
\geq \rho\,\mathsf{OPT}-2bL\varphi_M^*
-\frac{17\alpha cL}{\epsilon}
-\left(\frac{17\rho\alpha cL}{\epsilon}+\frac{2\rho cL}{\eta}\right)\mathbb E[G].
\]

\medskip
\noindent\underline{Prediction-error bound.}
Each completed predictive epoch accumulates at least $\Theta_{\mathrm p}$ clipped error by Eq.~\eqref{eq:gamma-capped-epoch-times}. These epochs are disjoint subsets of $[M]$, so, using $\rho=\kappa\delta$, we have pathwise
\begin{align*}
G\Theta_{\mathrm p}\leq\varphi_M^*,
\qquad
G\leq\frac{b\delta\varphi_M^*}{5\rho\alpha c}.
\end{align*}
Substituting this bound gives
\[
\mathsf{ALG}
\geq \rho\,\mathsf{OPT}
-\frac{L}{\epsilon}\left[
17\alpha c+
\left(2\epsilon+\frac{\delta}{5}\left(17+\frac{2\epsilon}{\alpha\eta}\right)\right)b\varphi_M^*
\right].
\]
It remains to bound the coefficient of $b\varphi_M^*$ by $17\eta/5$. Set $\chi\defeq\epsilon/\eta\in(0,1)$. Since $\delta/\eta=1-15\chi/16$ and $\alpha\geq3$,
\[
\frac{1}{\eta}\left[2\epsilon+\frac{\delta}{5}\left(17+\frac{2\epsilon}{\alpha\eta}\right)\right]
\leq
2\chi+\left(1-\frac{15\chi}{16}\right)\left(\frac{17}{5}+\frac{2\chi}{15}\right)
=\frac{17}{5}-\frac{253}{240}\chi-\frac18\chi^2
\leq\frac{17}{5}.
\]
Consequently,
\[
\mathsf{ALG}
\geq
\rho\,\mathsf{OPT}
-\frac{L}{\epsilon}
\left(17\alpha c+\frac{17}{5}b\eta\varphi_M^*\right).
\]
Since $\mathsf{OPT}>0$, division by $\mathsf{OPT}$ proves Lemma~\ref{lem:gamma-oracle-error-dependent}.\hfill\Halmos

\subsection{Proof of Lemma~\ref{lem:gamma-oracle-many-sample}}
\label{sec:proof-lem-gamma-oracle-many-sample}
\begingroup
Fix an odd $i$ and condition on $\mathcal F_i$. Since $\{i\leq N\}\in\mathcal F_i$, both claims are immediate when $i>N$; hence assume $i\leq N$. Couple the executed conservative actions to an uninterrupted online-oracle continuation $\bm a^{\mathrm{on},i}_{s_i:\infty}$ from $\mathcal I(s_i)$ that is conditionally independent of the Monte Carlo arrays. For $j\geq1$, let
\[
X_j\defeq R_j^{\mathcal I(s_i)}
\left(\bm e_{s_i:s_i+j-1},\bm a^{\mathrm{on},i}_{s_i:s_i+j-1}\right),
\qquad
r_j\defeq\mathbb E[X_j\mid\mathcal F_i],
\qquad \mu_0\defeq0,\qquad
\mu_j\defeq\sum_{k=1}^{j}r_k.
\]
Reindex the estimates by elapsed period: $\widehat R_j\defeq\widehat R_{i,s_i+j-1}$ has conditional mean $\mu_j$. With $J_i\defeq s_{i+1}-s_i$, we have $J_i\perp(X_j)_{j\geq1}\mid\mathcal F_i$; hence Tonelli gives
\begin{align}
\mathbb E[V_i\mid\mathcal F_i]
&=\sum_{j\geq1}r_j\Pr(J_i\geq j\mid\mathcal F_i).
\label{eq:reward-conservative-stopped-value}
\end{align}

Each sample contributing to $\widehat R_j$ lies in $[0,jL]$, and the estimate averages $n_{s_i,s_i+j-1}=\lceil Hj^2\log(\mathrm e\,(s_i+j-1))\rceil$ such samples. Put $\psi(x)\defeq2Hx^2/L^2$. Because the simulation array for period $s_i+j-1$ is independent of $\mathcal F_i$, conditional Hoeffding's inequality and $s_i\geq i$ give, for $\psi(x)\geq3$,
\begin{align}
p_i(x)&\defeq\sum_{j\geq1}
\Pr\!\left(\left|\widehat R_j-\mu_j\right|\geq x\mid\mathcal F_i\right)
\leq2\mathrm e^{-\psi(x)}\sum_{n\geq s_i}n^{-\psi(x)}
\leq\frac{3\mathrm e^{-\psi(x)}}{i^2}. \label{eq:reward-conservative-hoeffding-tail}
\end{align}
The last step uses the integral bound for the displayed series. With $\lambda=\min\{1,L\}$, the sampling scale in Algorithm~\ref{alg:Adaptive-Switching-Algorithm-for-gamma-oracle} gives
\[
\psi_{\mathrm r}\defeq \psi(\lambda/2)=\frac{H\lambda^2}{2L^2}
\geq\max\!\left\{3,
\log\!\left(\frac{7680\eta\alpha cL}{\epsilon^2\lambda}\right)\right\}.
\]
Thus $3\mathrm e^{-\psi_{\mathrm r}}
\leq\epsilon^2\lambda/(2560\eta\alpha cL)$. Using $\epsilon<\eta\leq1$,
Eq.~\eqref{eq:reward-conservative-hoeffding-tail} gives
\begin{align}
p_i(\lambda/2)
&\leq\frac{\epsilon^2\lambda}{2560\eta\alpha cL i^2}
\leq\frac{\lambda}{2\Theta_{\mathrm c}},
\qquad
\Theta_{\mathrm c}p_i(\lambda/2)\leq\frac{\epsilon\lambda}{160i^2}.
\label{eq:reward-conservative-tail-specialized}
\end{align}

For the lower estimate, since $\Theta_{\mathrm c}>\lambda$, let
\[
A\defeq\left\{j\in\{1,\ldots,M-s_i\}:
\mu_j\geq\Theta_{\mathrm c}-\lambda/2\right\}.
\]
If $A\neq\varnothing$, let $j_0\defeq\min A$. The event $J_i<j_0$ requires $\widehat R_j\geq\Theta_{\mathrm c}$ for some $j<j_0$, and hence $|\widehat R_j-\mu_j|\geq\lambda/2$. Therefore $\Pr(J_i\geq j_0\mid\mathcal F_i)\geq1-p_i(\lambda/2)$, so Eqs.~\eqref{eq:reward-conservative-stopped-value} and~\eqref{eq:reward-conservative-tail-specialized} give
\[
\mathbb E[V_i\mid\mathcal F_i]
\geq\mu_{j_0}\bigl(1-p_i(\lambda/2)\bigr)
\geq(\Theta_{\mathrm c}-\lambda/2)
\left(1-\frac{\lambda}{2\Theta_{\mathrm c}}\right)
\geq\Theta_{\mathrm c}-\lambda.
\]
If $A=\varnothing$, $\{i<N\}=\{J_i\leq M-s_i\}$ requires such a deviation for some $j\leq M-s_i$. Since rewards are nonnegative,
\begin{align*}
\mathbb E\!\left[
V_i-(\Theta_{\mathrm c}-\lambda)\mathbb I(i<N)
\mid\mathcal F_i\right]
&\geq-(\Theta_{\mathrm c}-\lambda)\Pr(i<N\mid\mathcal F_i)
\geq-\Theta_{\mathrm c}\,p_i(\lambda/2)
\geq-\frac{\epsilon\lambda}{160i^2}.
\end{align*}
When $A\neq\varnothing$, the bound $\mathbb E[V_i\mid\mathcal F_i]\geq\Theta_{\mathrm c}-\lambda$ and $\mathbb I(i<N)\leq1$ prove the first estimate with the stronger lower bound zero.

For the upper estimate, only $1\leq j\leq M-s_i+1$ can contribute to $V_i$. If $\mu_j<\Theta_{\mathrm c}+\lambda/2$ throughout this range, Eq.~\eqref{eq:reward-conservative-stopped-value} gives $\mathbb E[V_i\mid\mathcal F_i]\leq\Theta_{\mathrm c}+\lambda/2$. Otherwise, let $j_1$ be the first index with $\mu_{j_1}\geq\Theta_{\mathrm c}+\lambda/2$. Since $r_j\leq L$,
\[
\mu_{j_1}\leq\Theta_{\mathrm c}+\lambda/2+L.
\]
For $j_1\leq j\leq M-s_i$, $J_i\geq j+1$ implies $\widehat R_j<\Theta_{\mathrm c}$ and therefore $|\widehat R_j-\mu_j|\geq\lambda/2$. Consequently,
\begin{align*}
\mathbb E[V_i\mid\mathcal F_i]
\leq\mu_{j_1}
+L\sum_{j=j_1}^{M-s_i}
\Pr(J_i\geq j+1\mid\mathcal F_i)
\leq\Theta_{\mathrm c}+\lambda/2+L+L\,p_i(\lambda/2)
\leq\frac{17\eta\alpha cL}{\epsilon}.
\end{align*}
The last inequality follows from $\lambda/2+L+Lp_i(\lambda/2)\leq5L/2\leq\eta\alpha cL/\epsilon$. This proves the second estimate.
\hfill\Halmos
\endgroup

\subsection{\texorpdfstring{Proof of Eq.~\eqref{eq:ASA-error-dependent-gamma-oracle-Rpre}}{Proof of the predicted-optimum comparison}}
\label{sec:proof-gamma-oracle-Rpre}
\begingroup
Let $R\defeq\mathrm{Opt}(\mathcal P,\bm e_{1:\infty})$,
$R^*\defeq\mathrm{Opt}(\mathcal P,\bm e^*_{1:\infty})>0$,
and define $x\defeq bL\varphi_\vee^*/R^*$, $y_{\mathrm{sw}}\defeq\alpha cL/R^*$, and
$\rho\defeq\gamma\alpha/(\alpha+\gamma)$. Both infinite-horizon optima equal their optima truncated at $M^\vee$. Observation~\ref{obs:lipschitz-telescoping}, applied on this common finite horizon, gives
$
R\geq R^*(1-x),  \varphi_M^*\leq\varphi_\vee^*.
$
Set $z\defeq17y_{\mathrm{sw}}+\frac{22}{5}\eta x$ and $B_{\mathrm{pred}}\defeq\rho-z/\epsilon$. If
$B_{\mathrm{pred}}\leq\eta-\frac{19}{20}\epsilon$, the robustness bound proves the claim. Otherwise, since $\eta-19\epsilon/20>0$, we have
$0<z<\epsilon(\rho-\eta+19\epsilon/20)<\epsilon\rho\leq\epsilon<\eta$.
Together with $(22/5)\eta x\leq z$, this gives $x<5/22<1$, so $1-x>0$. Moreover,
\[
z(1-x)-\left(17y_{\mathrm{sw}}+\frac{17}{5}\eta x\right)=x(\eta-z)\geq0.
\]
Consequently,
\[
\frac{L\left(17\alpha c+\frac{17}{5}b\eta\varphi_M^*\right)}{\epsilon R}
\leq\frac{17y_{\mathrm{sw}}+\frac{17}{5}\eta x}{\epsilon(1-x)}
\leq\frac z\epsilon,
\]
so the realized-optimum bound is at least $B_{\mathrm{pred}}$, proving Eq.~\eqref{eq:ASA-error-dependent-gamma-oracle-Rpre}.
\hfill\Halmos
\endgroup

\section{Supplementary Proofs for Section~\ref{sec:adaswitch-cost-minimization}}

\subsection{\texorpdfstring{Proof of Theorem~\ref{thm:cost-version-ASA-error-dependent-1-oracle}}{Proof of the cost-minimization theorem}}
\label{sec:proof-thm-cost-version-ASA-error-dependent-1-oracle}

\begingroup
Algorithm~\ref{alg:Adaptive-Switching-Algorithm-for-1-oracle-c} records only the changes to Algorithm~\ref{alg:Adaptive-Switching-Algorithm-for-1-oracle}; its initialization, timing, histories, online-oracle resets, and clipped-error update are otherwise unchanged.

\begin{algorithm}[H]
\caption{Changes to Algorithm~\ref{alg:Adaptive-Switching-Algorithm-for-1-oracle} for cost minimization}
\label{alg:Adaptive-Switching-Algorithm-for-1-oracle-c}
\begin{algorithmic}[1]
\State Use an exact offline oracle and an $(\eta,\beta)$-online oracle with known factor $\eta$ for cost minimization.
\State In a conservative epoch, set $s\gets\mathrm{Opt}(\mathcal P^{\mathcal I(\tau)},\bm e_{\tau:t})$ and enter predictive mode when $s\geq10(\eta+\epsilon)cL/\epsilon$.
\State In a predictive epoch, set $\widehat M_t\gets\mathrm{EstimateM}(\bm e_{1:t},\bm e^*_{1:\infty})$ and replace the maximization call by
\begin{align}
\bm a^{*,(t)}_{t:\widehat M_t}
\in\arg\min_{\bm a''_{t:\widehat M_t}\in\bm A_{t:\widehat M_t}}
\mathrm{Val}\!\left(\mathcal P^{\mathcal I(t)},
e_t\circ\bm e^*_{t+1:\widehat M_t},\bm a''_{t:\widehat M_t}\right).
\label{eq:adaswitch-1-predictive-follow-c}
\end{align}
\State Execute $a_t=a_t^{*,(t)}$ and return to conservative mode when $\varphi\geq2(\eta+\epsilon)c/b$.
\end{algorithmic}
\end{algorithm}

For the analysis, extend each finite predictive plan returned by Algorithm~\ref{alg:Adaptive-Switching-Algorithm-for-1-oracle-c} beyond $\widehat M_t$ using the fixed measurable selection rule; the estimator makes the extension optimal on the full hybrid continuation.

\begin{lemma}[Predictive-following bound for cost minimization]
\label{lem:cost-predictive-follow-finite}
Let $\mathcal P$ be $(u,v)$-Lipschitz and $f$-bounded-influence, with $b>0$, $b\geq u$, and $c\geq\max\{v,f,1\}$. Fix $m,n\in\mathbb Z_{\geq0}$, a history $\mathcal I=(\bm e_{1:m},\bm a_{1:m})$, a realized continuation $\bm e_{m+1:\infty}$, and an admissible request-sequence prediction $\bm e^*_{1:\infty}$. For every $t\in\{m+1,\ldots,m+n\}$, let $\mathcal I(t)=(\bm e_{1:t-1},\bm a_{1:t-1})$ and execute the first action $a_t=a_t^{*,(t)}$ of the finite plan in Eq.~\eqref{eq:adaswitch-1-predictive-follow-c}. Then
\begin{align*}
\mathrm{Val}(\mathcal P^{\mathcal I},\bm e_{m+1:m+n},\bm a_{m+1:m+n})
&\leq \mathrm{Opt}(\mathcal P^{\mathcal I},\bm e_{m+1:m+n})
+2bL\sum_{j=m+1}^{m+n}\hat d(e_j,e_j^*)+cL.
\end{align*}
\end{lemma}

\noindent\emph{Proof.}
The case $n=0$ is immediate. Otherwise set
\[
\widetilde{\bm e}_{m+1:\infty}
\defeq\bm e_{m+1:m+n}\circ\bm e^*_{m+n+1:\infty},
\qquad
D_n\defeq\sum_{j=m+1}^{m+n}\hat d(e_j,e_j^*),
\]
and extend the generated actions by sequential minimization on the predicted tail, obtaining $\widetilde{\bm a}_{m+1:\infty}$. This splice differs from the prediction in at most the first $n$ continuation coordinates. The last-disagreement induction from Lemma~\ref{lem:predictive-following-infinite-reward} applies with minimization and reversed inequalities, using Observation~\ref{obs:sequential-exact-optimality}. The induction and the Lipschitz comparison of the two optima, respectively, give
\[
\begin{aligned}
\mathrm{Val}(\mathcal P^{\mathcal I},\widetilde{\bm e}_{m+1:\infty},
\widetilde{\bm a}_{m+1:\infty})
&\leq \mathrm{Opt}(\mathcal P^{\mathcal I},\bm e^*_{m+1:\infty})+bLD_n\\
&\leq \mathrm{Opt}(\mathcal P^{\mathcal I},\widetilde{\bm e}_{m+1:\infty})+2bLD_n.
\end{aligned}
\]
Let $C^{\mathrm{gen}}_{m+1:m+n}$ be the generated prefix cost, $\mathcal J$ its history through $m+n$, and $\mathcal J^*$ that of a finite-interval optimizer. The preceding bound, sequential optimality, and feasible concatenation give
\[
\begin{aligned}
C^{\mathrm{gen}}_{m+1:m+n}+\mathrm{Opt}(\mathcal P^{\mathcal J},\bm e^*_{m+n+1:\infty})
&\leq \mathrm{Opt}(\mathcal P^{\mathcal I},\bm e_{m+1:m+n})
+\mathrm{Opt}(\mathcal P^{\mathcal J^*},\bm e^*_{m+n+1:\infty})+2bLD_n\\
&\leq \mathrm{Opt}(\mathcal P^{\mathcal I},\bm e_{m+1:m+n})
+\mathrm{Opt}(\mathcal P^{\mathcal J},\bm e^*_{m+n+1:\infty})+2bLD_n+cL,
\end{aligned}
\]
where the second inequality uses bounded influence. Cancelling the common tail term proves the claim.\hfill\Halmos

We now prove Theorem~\ref{thm:cost-version-ASA-error-dependent-1-oracle}. For robustness, slack from completed conservative epochs covers predictive and restart losses. For the error-dependent bound, we cap conservative-epoch optima and charge completed predictive epochs to accumulated clipped prediction error.

\noindent\underline{Epoch setup.}
Fix the request sequences and put $M\defeq M(\bm e_{1:\infty})\geq1$ and $\varphi_M^*\defeq\sum_{t=1}^{M}\hat d(e_t,e_t^*)$. Let $\bm a$ be the algorithmic action sequence, $\bm a^*_{1:M}$ an offline-optimal action sequence, and $\{\mathcal G_t\}_{t\geq1}$ the filtration generated by the algorithmic randomness revealed before period $t$. Define $\mathcal I(t)$ and $\mathcal I^*(t)$ from these action sequences and write
\[
\mathsf{ALG}\defeq
\mathrm{Val}(\mathcal P,\bm e_{1:\infty},\mathrm{AdaSwitch})
=\mathbb E\!\left[\mathrm{Val}(\mathcal P,\bm e_{1:\infty},\bm a_{1:\infty})\right],
\qquad
\mathsf{OPT}\defeq\mathrm{Opt}(\mathcal P,\bm e_{1:\infty}).
\]
Set $s_1=1$ and, when $s_i\leq M$, define
\begin{align*}
Q_i^{\mathrm c}(t)&\defeq
\begin{cases}
\mathrm{Opt}(\mathcal P^{\mathcal I(s_i)},\bm e_{s_i:t}),&2\nmid i,\\
\displaystyle\sum_{j=s_i}^{t}\hat d(e_j,e_j^*),&2\mid i,
\end{cases}
\qquad
\theta_i^{\mathrm c}\defeq
\begin{cases}
\dfrac{10(\eta+\epsilon)cL}{\epsilon},&2\nmid i,\\
\dfrac{2(\eta+\epsilon)c}{b},&2\mid i,
\end{cases}\notag\\
t_i^{\mathrm c}&\defeq\inf\{t\in\{s_i,\ldots,M\}:Q_i^{\mathrm c}(t)\geq\theta_i^{\mathrm c}\},
\qquad
s_{i+1}\defeq\min\{M+1,t_i^{\mathrm c}+1\}.
\end{align*}
Here $\inf\varnothing=+\infty$ and $s_{i+1}=M+1$ if $s_i=M+1$. 
With $\mathcal F_i\defeq\mathcal G_{s_i}$, the induction after Eq.~\eqref{eq:exact-oracle-capped-epoch-times} makes each $s_i$ a bounded stopping time. For the fixed request/prediction pair, the epoch-start history determines every threshold test; actions sampled later in that epoch do not enter these tests. Thus $s_{i+1}$ is $\mathcal F_i$-measurable. With $N\defeq\max\{i:s_i\leq M\}<\infty$, we have
\[
\{i\leq N\}=\{s_i\leq M\}\in\mathcal F_i,
\qquad
\{i<N\}=\{s_{i+1}\leq M\}\in\mathcal F_i.
\]
Epoch $i\leq N$ spans periods $s_i,\ldots,s_{i+1}-1$ and is \emph{completed} if $i<N$. Define its costs, continuation optimum, and clipped error below, extending them by zero after $N$:
\begin{align*}
V_i&\defeq
\mathbb I(i\leq N)\,
\mathrm{Val}(\mathcal P^{\mathcal I(s_i)},\bm e_{s_i:s_{i+1}-1},
\bm a_{s_i:s_{i+1}-1}),\\
V_i^*&\defeq
\mathbb I(i\leq N)\,
\mathrm{Val}(\mathcal P^{\mathcal I^*(s_i)},\bm e_{s_i:s_{i+1}-1},
\bm a^*_{s_i:s_{i+1}-1}),\\
O_i&\defeq
\mathbb I(i\leq N)\,
\mathrm{Opt}(\mathcal P^{\mathcal I(s_i)},\bm e_{s_i:s_{i+1}-1}),\\
D_i&\defeq
\mathbb I(i\leq N)\sum_{j=s_i}^{s_{i+1}-1}\hat d(e_j,e_j^*).
\end{align*}
The variables $V_i^*$, $O_i$, and $D_i$ are $\mathcal F_i$-measurable. Moreover, the epochs partition the realized effective horizon, so, pathwise,
\[
\sum_{i=1}^{\infty}V_i
=\mathrm{Val}(\mathcal P,\bm e_{1:\infty},\bm a_{1:\infty}),
\qquad
\sum_{i=1}^{\infty}V_i^*
=\mathsf{OPT}.
\]

\noindent\underline{Epoch cost bounds.}
For a predictive epoch ($2\mid i$), Lemma~\ref{lem:cost-predictive-follow-finite} and bounded influence give
\begin{align}
V_i
&\leq V_i^*+2bLD_i+2cL\,\mathbb I(i\leq N),
\label{eq:cost-1-oracle-predictive-1}\\
V_i
&\leq \eta V_i^*+8(\eta+\epsilon)cL\,\mathbb I(i\leq N).
\label{eq:cost-1-oracle-predictive-eta}
\end{align}
The two $cL$ terms come from the lemma and the bounded-influence comparison between $\mathcal I(s_i)$ and $\mathcal I^*(s_i)$. The exit rule and one-period overshoot give $D_i\leq3(\eta+\epsilon)c/b$, and the second inequality follows from $\eta\geq1$.

For a conservative epoch ($2\nmid i$), the fixed-endpoint online-oracle guarantee applies conditional on $\mathcal F_i$. The initial clause of Definition~\ref{def:eta-online-oracle-cost} contributes $\beta$ only when $i=1$; subsequent restarts contribute at most $\eta fL\leq\eta cL$. Since the histories agree initially, bounded influence and feasibility give $O_i\leq V_i^*+cL\,\mathbb I(2\leq i\leq N)$. Thus, writing $\eta O_i=O_i+(\eta-1)O_i$,
\begin{equation}
\mathbb E[V_i\mid\mathcal F_i]
\leq \eta\left[O_i+cL\,\mathbb I(2\leq i\leq N)\right]
+\beta\,\mathbb I(i=1)
\leq V_i^*+(\eta-1)O_i
+(\eta+1)cL\,\mathbb I(2\leq i\leq N)
+\beta\,\mathbb I(i=1).
\label{eq:cost-1-oracle-conservative-full-eta}
\end{equation}

On $\{i<N\}$, the conservative threshold gives $O_i\geq10(\eta+\epsilon)cL/\epsilon$. Rewriting $\eta=(\eta+\epsilon)-\epsilon$ in the first inequality of Eq.~\eqref{eq:cost-1-oracle-conservative-full-eta} and then using the same bounded-influence comparison gives
\begin{align}
\mathbb E[V_i]
&\leq\beta\,\mathbb I(i=1)+(\eta+\epsilon)\,
\mathbb E\left[
V_i^*+2cL\,\mathbb I(2\leq i\leq N)
-10cL\,\mathbb I(i<N)\right].
\label{eq:cost-1-oracle-conservative-eta}
\end{align}

\noindent\underline{Robustness.}
Let $F\defeq\lfloor N/2\rfloor$ and $G\defeq\lfloor(N-1)/2\rfloor$. There are $F$ predictive epochs, $F$ completed conservative epochs, and $G\leq F$ noninitial conservative epochs. Each completed conservative epoch supplies $10(\eta+\epsilon)cL$ of slack, covering an $8(\eta+\epsilon)cL$ predictive allowance and up to $2(\eta+\epsilon)cL$ for a subsequent restart and history comparison. Summing Eqs.~\eqref{eq:cost-1-oracle-predictive-eta} and~\eqref{eq:cost-1-oracle-conservative-eta}, with $\eta\leq\eta+\epsilon$, gives
\begin{align}
\mathsf{ALG}
&=\sum_{i=1}^{\infty}\mathbb E[V_i]\notag\\
&\leq\beta+(\eta+\epsilon)\mathsf{OPT}+(\eta+\epsilon)cL\,\mathbb E\left[-2F+2G\right]\notag\\
&\leq\beta+(\eta+\epsilon)\mathsf{OPT}.
\label{eq:cost-1-oracle-worst-case}
\end{align}

\noindent\underline{Error-dependent bound.}
Let $q\defeq\mathbb I(2\mid N)$ and $S_{\mathrm c}^{\mathrm{opt}}\defeq\sum_{2\nmid i}O_i$, the sum of conservative-epoch optima. There are $G+1$ conservative epochs and $G+q$ predictive epochs, of which $G$ are completed. Taking expectations, summing Eqs.~\eqref{eq:cost-1-oracle-predictive-1} and~\eqref{eq:cost-1-oracle-conservative-full-eta}, and using $\sum_iV_i^*=\mathsf{OPT}$ and disjointness of predictive epochs gives
\begin{align}
\mathsf{ALG}
\leq\beta+\mathsf{OPT}
+2bL\varphi_M^*
+\mathbb E\left[
(\eta-1)S_{\mathrm c}^{\mathrm{opt}}+cL\bigl((\eta+3)G+2q\bigr)\right].
\label{eq:cost-1-oracle-error-aggregate-sharp}
\end{align}
Put $K_{\mathrm c}\defeq L\left(\frac{10(\eta+\epsilon)c}{\epsilon}+1\right)$. Before the final period of a conservative epoch, the entry threshold has not been reached; extending an optimal prefix by one period costs at most $L$. The empty-interval convention covers a one-period epoch. Thus, for odd $i$,
\begin{align*}
O_i\leq K_{\mathrm c}\,\mathbb I(i\leq N).
\end{align*}
Consequently, $S_{\mathrm c}^{\mathrm{opt}}\leq K_{\mathrm c}(G+1)$ pathwise. Each of the $G$ completed predictive epochs consumes at least $2(\eta+\epsilon)c/b$ clipped error on a disjoint interval. Hence
\[
G\leq\frac{b\varphi_M^*}{2(\eta+\epsilon)c},
\qquad q\leq1.
\]
Substituting these bounds into Eq.~\eqref{eq:cost-1-oracle-error-aggregate-sharp} yields
\begin{align}
\mathsf{ALG}
&\leq\beta+\mathsf{OPT}
+2bL\varphi_M^*+(\eta-1)K_{\mathrm c}+2cL
+\frac{b\varphi_M^*}{2(\eta+\epsilon)c}
\left[(\eta-1)K_{\mathrm c}+(\eta+3)cL\right]\notag\\
&\leq\beta+\mathsf{OPT}
+\frac{L}{\epsilon}\Big(
\left[2\epsilon+(\eta-1)(10\eta+11\epsilon)\right]c
+\left[5(\eta-1)+4\epsilon\right]b\varphi_M^*
\Big).
\label{eq:cost-1-oracle-error-exact-sharp}
\end{align}
The last inequality uses
\begin{align*}
(\eta-1)K_{\mathrm c}+2cL
&\leq\frac{cL}{\epsilon}
\left[2\epsilon+(\eta-1)(10\eta+11\epsilon)\right],\\
\frac{\epsilon}{2(\eta+\epsilon)}
\left(\eta+3+\frac{\eta-1}{c}\right)
&\leq\frac{\epsilon(\eta+1)}{\eta+\epsilon}
\leq2\epsilon.
\end{align*}
Dividing by $\mathsf{OPT}>0$ and combining Eq.~\eqref{eq:cost-1-oracle-error-exact-sharp} with Eq.~\eqref{eq:cost-1-oracle-worst-case} proves Theorem~\ref{thm:cost-version-ASA-error-dependent-1-oracle}.
\hfill\Halmos
\endgroup

\subsection{Proof of Theorem~\ref{thm:cost-version-ASA-error-dependent-gamma-oracle}}
\label{sec:proof-thm-cost-version-ASA-error-dependent-gamma-oracle}
Algorithm~\ref{alg:Adaptive-Switching-Algorithm-for-gamma-oracle-c} adapts Algorithm~\ref{alg:Adaptive-Switching-Algorithm-for-gamma-oracle} using $\epsilon>0$ and the cost thresholds, cost estimates, and sampling scale specified below. All other definitions remain in force, including the hybrid prefixes, effective-horizon estimator, mode and epoch variables, and within-period timing. Set
$
\lambda\defeq\min\{1,L\},
T_{\mathrm c}\defeq\frac{18\eta(\eta+\epsilon)\gamma\alpha cL}{\epsilon},
\Delta_{\mathrm c}\defeq\frac{16\eta\lambda}{15\epsilon},
m_{\mathrm c}\defeq\min\{\lambda,\Delta_{\mathrm c}\},
$
and fix
$
H\geq\frac{2L^2}{\Delta_{\mathrm c}^2}
\max\!\left\{3,\log\!\left(\frac{960T_{\mathrm c}}{m_{\mathrm c}}\right)\right\}.
$

\begin{algorithm}[htbp]
\caption{Changes to Algorithm~\ref{alg:Adaptive-Switching-Algorithm-for-gamma-oracle} for cost minimization}
\label{alg:Adaptive-Switching-Algorithm-for-gamma-oracle-c}
\begin{algorithmic}[1]
\State Use a $\gamma$-approximate offline oracle and an $(\eta,\beta)$-online oracle ($\beta\geq0$) for cost minimization, with known factors $1\leq\gamma\leq\eta$ and resettable-simulation access; fix $\epsilon>0$ and $\alpha,b,c>0$.
\State During a conservative epoch starting at $\tau$, after executing period $t$, run $n_{\tau,t}=\lceil H(t-\tau+1)^2\log(\mathrm e\,t)\rceil$ fresh resettable simulations of an uninterrupted online-oracle continuation on $\bm e_{\tau:t}$ and let $\widehat C^{\mathrm{on}}_{\tau,t}$ be their mean total cost. Let $\bm a^{\mathrm{off}}_{\tau:t}$ be the plan returned by the $\gamma$-approximate offline oracle for $(\mathcal P^{\mathcal I(\tau)},\bm e_{\tau:t})$, and set
\[
C^{\mathrm{off}}_{\tau,t}
\defeq \mathrm{Val}(\mathcal P^{\mathcal I(\tau)},\bm e_{\tau:t},\bm a^{\mathrm{off}}_{\tau:t}).
\]
Enter predictive mode iff $\widehat C^{\mathrm{on}}_{\tau,t}\geq T_{\mathrm c}$ and $C^{\mathrm{off}}_{\tau,t}\geq\gamma\lambda$ both hold.
\State At a predictive batch start $q$, use the loop in Algorithm~\ref{alg:Adaptive-Switching-Algorithm-for-gamma-oracle} with costs in place of rewards: extend while $r<\widehat M_q$ and the current candidate costs less than $\alpha cL$, and require each returned plan to satisfy
\begin{align*}
\mathrm{Val}(\mathcal P^{\mathcal I(q)},
\widetilde{\bm e}^{(q)}_{q:r},\bm a_{q:r})
\leq\gamma\,\mathrm{Opt}(\mathcal P^{\mathcal I(q)},
\widetilde{\bm e}^{(q)}_{q:r}).
\end{align*}
\State Replace Line~\ref{line:adaswitch-gamma-predictive-if-end} by this rule: if $r=\widehat M_q$ and the candidate through $r$ costs less than $\alpha cL$, retain this plan, set $\tau_p\gets\infty$, and extend it beyond $r$ using the standing measurable selection rule.
\State \label{line:adaswitch-gamma-predictive-stop-c} Keep the same clipped-error update and return to conservative mode after period $t$ iff
\[
\varphi\geq\Theta_{\mathrm p}^{\mathrm c}\defeq
\frac{5(\eta+\epsilon)\alpha c}{b}.
\]
\end{algorithmic}
\end{algorithm}

The next two lemmas establish the theorem's worst-case and error-dependent bounds, respectively; taking their minimum proves the theorem.

\begin{lemma}
\label{lem:cost-gamma-oracle-robustness}
Under the setting of Theorem~\ref{thm:cost-version-ASA-error-dependent-gamma-oracle}, AdaSwitch incurs expected cost at most
\[
\left(\eta+\frac{19\epsilon}{20}\right)\mathrm{Opt}(\mathcal P,\bm e_{1:\infty})
+\left(1+\frac{15\epsilon}{16\eta}\right)\beta.
\]
\end{lemma}

\begin{lemma}
\label{lem:cost-gamma-oracle-error-dependent}
Under the setting of Theorem~\ref{thm:cost-version-ASA-error-dependent-gamma-oracle}, let $M=M(\bm e_{1:\infty})$ and $\varphi_M^*=\sum_{t=1}^{M}\hat d(e_t,e_t^*)$. Then AdaSwitch has an instancewise performance ratio of at most
\[
\frac{\beta}{\mathrm{Opt}(\mathcal P,\bm e_{1:\infty})}
+\frac{\gamma\alpha}{\alpha-\gamma}
+\frac{L}{\epsilon\cdot\mathrm{Opt}(\mathcal P,\bm e_{1:\infty})}
\left[
\gamma\alpha\eta(20\eta+21\epsilon)c
+\gamma\left(4\eta+\frac{8}{3}\epsilon\right)b\varphi_M^*
\right].
\]
\end{lemma}

\subsection{Proof of Lemma~\ref{lem:cost-gamma-oracle-robustness}}
\label{sec:proof-lem-cost-gamma-oracle-robustness}

We bound predictive-epoch overhead and offset it using expected credit from completed conservative epochs. The offline-cost test handles small offline optima, for which the remaining sampling error cannot be absorbed.

\medskip
\noindent\underline{Epoch setup.}
Fix the request sequences and $M=M(\bm e_{1:\infty})\geq1$. Let $\bm a$ be the executed action sequence and $\bm a^*$ an offline-optimal action sequence, with histories $\mathcal I(t)$ and $\mathcal I^*(t)$. Write
\[
\mathsf{ALG}\defeq
\mathbb E\!\left[\mathrm{Val}(\mathcal P,\bm e_{1:\infty},\bm a)\right],
\qquad
\mathsf{OPT}\defeq\mathrm{Opt}(\mathcal P,\bm e_{1:\infty}),
\qquad
\varphi_M^*\defeq\sum_{t=1}^{M}\hat d(e_t,e_t^*),
\]
and set
\[
\rho_{\mathrm c}\defeq\frac{\gamma\alpha}{\alpha-\gamma},
\qquad
\delta_{\mathrm c}\defeq\eta+\frac{15\epsilon}{16}.
\]

Use the product space and filtration $\{\mathcal G_t\}$ constructed before Eq.~\eqref{eq:gamma-capped-epoch-times}: live online-oracle seeds are independent of the Monte Carlo arrays, and the array indexed by period $r$ is revealed only after the action in period $r$ has been executed. For a conservative epoch starting at $s_i$, let $\widehat C^{\mathrm{on}}_{i,r}$ be the estimated cost of an uninterrupted online-oracle continuation and $C^{\mathrm{off}}_{i,r}$ the offline-oracle cost at $r$. Set $s_1=1$, $\Theta_{\mathrm p}^{\mathrm c}=5(\eta+\epsilon)\alpha c/b$, and, with $\inf\varnothing=+\infty$, define
\begin{align*}
s_{i+1}\defeq
\begin{cases}
M+1,&s_i=M+1,\\[2pt]
\displaystyle\min\!\left\{M+1,1+\inf\!\left\{r\in\{s_i,\ldots,M\}:\widehat C^{\mathrm{on}}_{i,r}\geq T_{\mathrm c}~\text{and}~C^{\mathrm{off}}_{i,r}\geq\gamma\lambda\right\}\right\},
&s_i\leq M,~2\nmid i,\\[6pt]
\displaystyle\min\!\left\{M+1,1+\inf\!\left\{r\in\{s_i,\ldots,M\}:\sum_{k=s_i}^{r}\hat d(e_k,e_k^*)\geq\Theta_{\mathrm p}^{\mathrm c}\right\}\right\},
&s_i\leq M,~2\mid i.
\end{cases}
\end{align*}
The $+1$ reflects post-period switching. The fixed-history offline-cost threshold test is measurable from the history and requests through the current period and reveals no live-oracle future randomness, so the stopping-time and conditional-independence argument following Eq.~\eqref{eq:gamma-capped-epoch-times} remains valid. Set $\mathcal F_i=\mathcal G_{s_i}$ and $N=\max\{i:s_i\leq M\}$.
On epoch $[s_i,s_{i+1}-1]$, let $V_i$ be the executed cost, $V_i^*$ the corresponding segment cost of the offline-optimal trajectory, $O_i$ the fixed-history continuation optimum, and $D_i\defeq\sum_{t=s_i}^{s_{i+1}-1}\hat d(e_t,e_t^*)$. Set all four quantities to zero for $i>N$. Then
$\mathsf{ALG}=\sum_i\mathbb E[V_i]$ and $\mathsf{OPT}=\sum_iV_i^*$.

\medskip
\noindent\underline{Predictive epochs.}
For an even $i\leq N$, let the $N_i$ batches start at $s_i^1=s_i<\cdots<s_i^{N_i}$, and set $s_i^{N_i+1}=s_{i+1}$. For batch $j\in[N_i]$, write $[q,r]=[s_i^j,s_i^{j+1}-1]$ and $D_{i,j}=\sum_{k=q}^{r}\hat d(e_k,e_k^*)$. The hybrid and realized sequences agree at $q$, so their cumulative clipped distance on $[q,r]$ is at most $D_{i,j}$. Every earlier batch $j<N_i$ is executed in full. Let $C_{\mathrm h},C_{\mathrm r},C^*$, and $O_{\mathrm h}$ denote, on this interval, the selected plan's hybrid and realized costs, the offline-optimal trajectory's realized cost, and the hybrid optimum. Here $C_{\mathrm h},C_{\mathrm r},O_{\mathrm h}$ use history $\mathcal I(q)$, whereas $C^*$ uses $\mathcal I^*(q)$. The threshold $C_{\mathrm h}\geq\alpha cL$, approximation guarantee, strong Lipschitzness, and bounded influence give
\[
\frac{\alpha-\gamma}{\alpha}C_{\mathrm h}+\gamma cL
\leq C_{\mathrm h}\leq\gamma O_{\mathrm h},\qquad
C_{\mathrm h}\geq C_{\mathrm r}-bLD_{i,j},\qquad
O_{\mathrm h}\leq C^*+cL+bLD_{i,j}.
\]
The first inequality absorbs the history-comparison term $\gamma cL$. Substituting the other bounds and canceling this term gives $C_{\mathrm r}\leq\rho_{\mathrm c}C^*+(1+\rho_{\mathrm c})bLD_{i,j}$; since $\rho_{\mathrm c}\geq1$,
\begin{align}
C_{\mathrm r}
&\leq\rho_{\mathrm c}C^*+2\rho_{\mathrm c}bLD_{i,j}.
\label{eq:cost-version-predictive-0}
\end{align}
For the last batch, denote the returned plan by $\bar{\bm a}_{q:p_{\mathrm{end}}}$, with endpoint $p_{\mathrm{end}}$. If its hybrid cost reaches $\alpha cL$, the preceding candidate through $p_{\mathrm{end}}-1$ costs less than $\alpha cL$; hence the hybrid optimum through $p_{\mathrm{end}}-1$ is also below $\alpha cL$, and the one-period bound gives
\[
\mathrm{Val}(\mathcal P^{\mathcal I(q)},\widetilde{\bm e}^{(q)}_{q:p_{\mathrm{end}}},\bar{\bm a}_{q:p_{\mathrm{end}}})
\leq\gamma(\alpha cL+L)\leq2\gamma\alpha cL.
\]
In this case $r\leq p_{\mathrm{end}}$ and $\bm a_{q:r}=\bar{\bm a}_{q:r}$, so nonnegative costs preserve the bound upon truncation to the executed prefix. If instead construction reaches $\widehat M_q$ with cost below $\alpha cL$, Algorithm~\ref{alg:Adaptive-Switching-Algorithm-for-gamma-oracle-c} retains that prefix. The estimator guarantee makes any appended hybrid tail cost zero, so the bound holds even if $r>\widehat M_q$. In either case, the strongly Lipschitz bound for the executed plan gives
\begin{align}
\mathrm{Val}(\mathcal{P}^{\mathcal{I}(q)},\bm{e}_{q:r},\bm{a}_{q:r})
&\leq 2\gamma\alpha cL+bLD_{i,j}.
\label{eq:cost-version-predictive-1}
\end{align}
Summing the bounds for earlier batches in Eq.~\eqref{eq:cost-version-predictive-0} and the bound for the last batch in Eq.~\eqref{eq:cost-version-predictive-1} gives
\begin{align}
V_i
&\leq \rho_{\mathrm c}V_i^*
+2\rho_{\mathrm c}bLD_i
+2\gamma\alpha cL\mathbb I(i\leq N).
\label{eq:cost-version-predictive-3}
\end{align}
Moreover, each clipped error is at most $c/b$, so the post-period test in Line~\ref{line:adaswitch-gamma-predictive-stop-c} can overshoot its threshold by at most $c/b$. Thus
$D_i\leq(\Theta_{\mathrm p}^{\mathrm c}+c/b)\mathbb I(i\leq N)\leq6(\eta+\epsilon)\alpha c\,\mathbb I(i\leq N)/b$. Using $\gamma\leq(\eta+\epsilon)\rho_{\mathrm c}$, Eq.~\eqref{eq:cost-version-predictive-3} implies
\begin{align}
V_i
&\leq \rho_{\mathrm c}V_i^*
+14(\eta+\epsilon)\rho_{\mathrm c}\alpha cL
\mathbb I(i\leq N).
\label{eq:cost-version-predictive-4}
\end{align}

\medskip
\noindent\underline{Conservative epochs.}
For odd $i$, conditional on $\mathcal F_i$, the offline-cost threshold tests and future Monte Carlo arrays determine the endpoint independently of the live-oracle continuation. For $i=1$, the initial online-oracle clause adds $\beta$ to the upper bound. For $1<i\leq N$, the history has positive length, so the restarted clause applies and incurs at most $\eta fL\leq\eta cL$. Applying the argument for Eq.~\eqref{eq:gamma-si-si1-conservative-i-neq-N-0} at the random epoch endpoint gives
\begin{align}
\mathbb E[V_i\mid\mathcal F_i]
&\leq\eta\,\mathbb E\!\left[
O_i+cL\mathbb I(2\leq i\leq N)
\mid\mathcal F_i\right]+\beta\mathbb I(i=1).
\label{eq:gamma-si-si1-conservative-i-neq-N-0-cost}
\end{align}

Bounded influence gives
$V_i^*\geq O_i-cL\mathbb I(2\leq i\leq N)$.
Solving Eq.~\eqref{eq:gamma-si-si1-conservative-i-neq-N-0-cost} for $\mathbb E[O_i\mid\mathcal F_i]$ and using $\delta_{\mathrm c}\leq\eta+\epsilon$ yields
\begin{align}
\mathbb E\left[V_i-\delta_{\mathrm c}V_i^*
\mid\mathcal F_i\right]
&\leq\mathbb E\left[
\left(1-\frac{\delta_{\mathrm c}}{\eta}\right)V_i
+2\delta_{\mathrm c}cL\mathbb I(2\leq i\leq N)
\mid\mathcal F_i\right]+\frac{\delta_{\mathrm c}}{\eta}\beta\mathbb I(i=1)\notag\\
&\leq\mathbb E\left[
-\frac{15\epsilon}{16\eta}V_i
+2(\eta+\epsilon)cL\mathbb I(2\leq i\leq N)
\mid\mathcal F_i\right]+\frac{\delta_{\mathrm c}}{\eta}\beta\mathbb I(i=1).
\label{eq:gamma-oracle-cost-conservative-0}
\end{align}

The negative coefficient of $V_i$ turns a lower cost bound into credit for predictive epochs. The first estimate below supplies this credit; the second is used in Section~\ref{sec:proof-lem-cost-gamma-oracle-error-dependent}. Both are proved in Section~\ref{sec:proof-lem-gamma-oracle-many-sample-c}.
\begin{lemma}
\label{lem:gamma-oracle-many-sample-c}
    Using the same notation as above, for every odd $i\in\mathbb{Z}_+$, we have
    \begin{align}
            &\mathbb E\left[V_i-\mathbb I(i<N)\left(T_{\mathrm c}-\Delta_{\mathrm c} \right)\middle|\mathcal F_i\right]\geq -\frac{\lambda}{160 i^2}, \notag \\
            &\mathbb E\left[V_i-\mathbb I(i\leq N)\frac{19\eta(\eta+\epsilon)\gamma\alpha c L}{\epsilon}\middle|\mathcal F_i\right]\leq \beta\mathbb I(i=1).\notag
    \end{align}
\end{lemma}
Because
\[
\frac{15\epsilon}{16\eta}(T_{\mathrm c}-\Delta_{\mathrm c})
\geq\frac{33}{2}(\eta+\epsilon)\gamma\alpha cL-\lambda
\]
and $15/(16\eta)\leq1$, substituting the first inequality of Lemma~\ref{lem:gamma-oracle-many-sample-c} into Eq.~\eqref{eq:gamma-oracle-cost-conservative-0} and taking total expectations gives
\begin{align}
\mathbb E\left[
V_i-\delta_{\mathrm c}V_i^*
\right]
&\leq
\mathbb E\left[
-\left(\frac{33}{2}(\eta+\epsilon)\gamma\alpha cL-\lambda\right)\mathbb I(i<N)
+2(\eta+\epsilon)cL\mathbb I(2\leq i\leq N)
\right]\notag\\
&\quad+\frac{\epsilon\lambda}{160i^2}+\frac{\delta_{\mathrm c}}{\eta}\beta\mathbb I(i=1).
\label{eq:gamma-oracle-cost-conservative-2}
\end{align}

\medskip
\noindent\underline{Combining epoch bounds.}
For each realization, set $F\defeq\lfloor N/2\rfloor$ and $G\defeq\lfloor(N-1)/2\rfloor$. Each of the $F$ predictive epochs is preceded by a completed conservative epoch ($i<N$), whose expected credit offsets its overhead. There are $G\leq F$ noninitial conservative epochs, so their restart and history-comparison terms can also be charged to these pairs. Moreover,
\[
\rho_{\mathrm c}
=\gamma+\frac{\gamma^2}{\alpha-\gamma}
\leq\gamma+\frac{\epsilon}{2}
\leq\delta_{\mathrm c}.
\]
Thus Eqs.~\eqref{eq:cost-version-predictive-4} and~\eqref{eq:gamma-oracle-cost-conservative-2}, together with $\sum_{i\geq1}i^{-2}<2$, give
\begin{align*}
\mathsf{ALG}
&\leq\delta_{\mathrm c}\mathsf{OPT}
+cL\,\mathbb E[F\Xi_{\mathrm c}]
+\frac{\epsilon\lambda}{80}+\frac{\delta_{\mathrm c}}{\eta}\beta,
\end{align*}
where the net allowance per pair, divided by $cL$, is
\[
\Xi_{\mathrm c}\defeq
14(\eta+\epsilon)\gamma\alpha\frac{\alpha}{\alpha-\gamma}
-\frac{33}{2}(\eta+\epsilon)\gamma\alpha
+\frac{\lambda}{cL}+2(\eta+\epsilon).
\]
Since $\alpha/(\alpha-\gamma)\leq16/15$, $\lambda/(cL)\leq1$, and $\gamma\alpha\geq16$,
\[
\Xi_{\mathrm c}
\leq
\left(\frac{224}{15}-\frac{33}{2}\right)
(\eta+\epsilon)\gamma\alpha+1+2(\eta+\epsilon)
\leq1-\frac{346}{15}(\eta+\epsilon)
\leq0.
\]
Here $\gamma\alpha\geq16$ follows from $\alpha\geq16\gamma$ and $\gamma\geq1$, while $\eta+\epsilon\geq1$. Hence $\mathsf{ALG}\leq\delta_{\mathrm c}\mathsf{OPT}+\epsilon\lambda/80+(\delta_{\mathrm c}/\eta)\beta$. If $\mathsf{OPT}\geq\lambda$, this is at most $(\eta+19\epsilon/20)\mathsf{OPT}+(\delta_{\mathrm c}/\eta)\beta$. If $\mathsf{OPT}<\lambda$, then before the first switch,
\[
C^{\mathrm{off}}_{1,t}
\leq\gamma\,\mathrm{Opt}(\mathcal P,\bm e_{1:t})
\leq\gamma\mathsf{OPT}<\gamma\lambda.
\]
The middle inequality is horizon monotonicity for nonnegative costs. The offline-cost threshold condition therefore prevents the first switch, and the online-oracle guarantee gives $\mathsf{ALG}\leq\eta\mathsf{OPT}+\beta$. Since $\delta_{\mathrm c}/\eta=1+15\epsilon/(16\eta)\geq1$, both cases give the claimed bound. This proves Lemma~\ref{lem:cost-gamma-oracle-robustness}.
\hfill\Halmos

\subsection{Proof of Lemma~\ref{lem:cost-gamma-oracle-error-dependent}}
\label{sec:proof-lem-cost-gamma-oracle-error-dependent}
Retain the notation and Eq.~\eqref{eq:cost-version-predictive-3} from the preceding proof. There are $F$ predictive epochs, $G$ of them completed, and $G+1$ conservative epochs. The executed batch intervals are disjoint subintervals of $[M]$, so their total clipped error is at most $\varphi_M^*$. Nonnegative costs imply $\sum_{2\mid i}V_i^*\leq\mathsf{OPT}$. Taking total expectations in the second inequality of Lemma~\ref{lem:gamma-oracle-many-sample-c} and summing the predictive and conservative bounds gives
\begin{align}
\mathsf{ALG}
\leq \beta+\rho_{\mathrm c}\mathsf{OPT}
+2\rho_{\mathrm c}bL\varphi_M^*
+\mathbb{E}\left[2\gamma\alpha cL F
+19\frac{\eta(\eta+\epsilon)\gamma\alpha cL}{\epsilon}(G+1)\right].
\label{eq:cost-gamma-gamma-last-step}
\end{align}
Each completed predictive epoch has clipped error at least $\Theta_{\mathrm p}^{\mathrm c}$. Hence, for every realization,
\begin{align*}
G+1
&\leq1+\frac{b\varphi_M^*}{5(\eta+\epsilon)\alpha c}.
\end{align*}
Since $F\leq G+1$, the expected epoch-count term in Eq.~\eqref{eq:cost-gamma-gamma-last-step} is at most
\[
\gamma\alpha cL\left(2+\frac{19\eta(\eta+\epsilon)}{\epsilon}\right)
\left(1+\frac{b\varphi_M^*}{5(\eta+\epsilon)\alpha c}\right).
\]
For the prediction-independent term and the prediction-error coefficient, respectively, $\alpha/(\alpha-\gamma)\leq16/15$, $\eta+\epsilon\geq1$, and $\eta\geq1$ give
\[
19\eta(\eta+\epsilon)+2\epsilon
\leq\eta(20\eta+21\epsilon),
\]
\[
\frac{2\alpha}{\alpha-\gamma}
+\frac{2}{5(\eta+\epsilon)}
+\frac{19\eta}{5\epsilon}
\leq\frac{8}{3}+\frac{4\eta}{\epsilon}.
\]
Substitution in Eq.~\eqref{eq:cost-gamma-gamma-last-step} and division by $\mathsf{OPT}>0$ prove Lemma~\ref{lem:cost-gamma-oracle-error-dependent}.
\hfill\Halmos

\subsection{Proof of Lemma~\ref{lem:gamma-oracle-many-sample-c}}
\label{sec:proof-lem-gamma-oracle-many-sample-c}
\begingroup
We control sampling errors, then use them to bound premature switching and prolonged continuation.

\medskip
\noindent\underline{Setup and sampling bound.}
Fix an odd $i$ and condition on $\mathcal F_i$. The claims are immediate on $\{i>N\}$, so assume $i\leq N$. Conditional on $\mathcal F_i$, the offline-cost tests are fixed, so the coupling used in the proof of Lemma~\ref{lem:gamma-oracle-many-sample} remains valid. Let $\bm a^{\mathrm{on},i}_{s_i:\infty}$ be the coupled uninterrupted online-oracle continuation, and define
\[
X_j\defeq C_j^{\mathcal I(s_i)}
\left(\bm e_{s_i:s_i+j-1},\bm a^{\mathrm{on},i}_{s_i:s_i+j-1}\right),
\qquad
r_j\defeq\mathbb E[X_j\mid\mathcal F_i],
\qquad \mu_0\defeq0,\qquad
\mu_j\defeq\sum_{k=1}^{j}r_k.
\]
Let $\widehat C^{\mathrm{on}}_j=\widehat C^{\mathrm{on}}_{i,s_i+j-1}$,
$C^{\mathrm{off}}_j=C^{\mathrm{off}}_{i,s_i+j-1}$, and $J_i=s_{i+1}-s_i$.
The estimate $\widehat C^{\mathrm{on}}_j$ has conditional mean $\mu_j$, the expected uninterrupted cost through $j$ periods.
Then $J_i\perp(X_j)_{j\geq1}\mid\mathcal F_i$, and Tonelli gives
\begin{align}
\mathbb E[V_i\mid\mathcal F_i]
&=\sum_{j\geq1}r_j\Pr(J_i\geq j\mid\mathcal F_i).
\label{eq:cost-conservative-stopped-value}
\end{align}

For $x\geq0$, define the sum of conditional deviation probabilities
\[
p_i(x)\defeq\sum_{j\geq1}
\Pr\!\left(\left|\widehat C^{\mathrm{on}}_j-\mu_j\right|\geq x\mid\mathcal F_i\right).
\]
Set $a_{\mathrm c}\defeq H\Delta_{\mathrm c}^2/(2L^2)$ and recall $m_{\mathrm c}=\min\{\lambda,\Delta_{\mathrm c}\}$. The prescribed sampling scale gives $a_{\mathrm c}\geq3$ and $3\mathrm e^{-a_{\mathrm c}}\leq m_{\mathrm c}/(320T_{\mathrm c})$. Hence the conditional Hoeffding bound in Eq.~\eqref{eq:reward-conservative-hoeffding-tail} yields
\begin{align}
p_i(\Delta_{\mathrm c}/2)
&\leq\frac{3\mathrm e^{-a_{\mathrm c}}}{i^2}
\leq\frac{m_{\mathrm c}}{320T_{\mathrm c}i^2}
\leq \min\left\{\frac{\Delta_{\mathrm c}}{4T_{\mathrm c}},
\frac{\lambda}{320T_{\mathrm c}i^2}\right\}.
\label{eq:cost-conservative-tail-specialized}
\end{align}
Moreover, $T_{\mathrm c}/\Delta_{\mathrm c}\geq270$.

\medskip
\noindent\underline{Lower estimate.}
A completed epoch ends before $M$, so only $j\leq M-s_i$ can trigger its completion. If $\mu_j\geq T_{\mathrm c}-\Delta_{\mathrm c}/2$ for some such $j$, let $j_0$ be the first. For $j<j_0$, $\mu_j<T_{\mathrm c}-\Delta_{\mathrm c}/2$, so $J_i<j_0$ requires an upward estimation error of at least $\Delta_{\mathrm c}/2$; the offline-cost condition can only delay switching. Thus
\[
\Pr(J_i\geq j_0\mid\mathcal F_i)\geq1-p_i(\Delta_{\mathrm c}/2),\qquad
\mathbb E[V_i\mid\mathcal F_i]
\geq \mu_{j_0}(1-p_i(\Delta_{\mathrm c}/2))
\geq (T_{\mathrm c}-\Delta_{\mathrm c}/2)\left(1-\frac{\Delta_{\mathrm c}}{4T_{\mathrm c}}\right)
\geq T_{\mathrm c}-\Delta_{\mathrm c}.
\]
If no such $j_0$ exists, completion $\{i<N\}=\{J_i\leq M-s_i\}$ requires such an upward error, so
\[
\Pr(i<N\mid\mathcal F_i)(T_{\mathrm c}-\Delta_{\mathrm c})
\leq p_i(\Delta_{\mathrm c}/2)T_{\mathrm c}
\leq\frac{\lambda}{160i^2}.
\]
The first case gives the desired lower bound with zero slack, since $\mathbb I(i<N)\leq1$; the second uses $V_i\geq0$. Thus, in both cases,
\[
\mathbb E\!\left[
V_i-(T_{\mathrm c}-\Delta_{\mathrm c})\mathbb I(i<N)\mid\mathcal F_i
\right]\geq-\frac{\lambda}{160i^2}.
\]

\medskip
\noindent\underline{Upper estimate.}
Set
$B_{\mathrm c}=T_{\mathrm c}/18=\eta(\eta+\epsilon)\gamma\alpha cL/\epsilon$ and $\beta_i=\beta\mathbb I(i=1)$.
Here the range $1\leq j\leq M-s_i+1$ includes the final executed period $M$. If $\mu_j<T_{\mathrm c}+B_{\mathrm c}/2+\beta_i$ throughout this range, Eq.~\eqref{eq:cost-conservative-stopped-value} gives
$\mathbb E[V_i\mid\mathcal F_i]\leq T_{\mathrm c}+B_{\mathrm c}/2+\beta_i$.
Otherwise, let $j_1$ be the first index reaching this level. Since $r_j\leq L$, $\mu_{j_1}\leq T_{\mathrm c}+B_{\mathrm c}/2+\beta_i+L$.
For $j\geq j_1$, the expected cost is large enough to force the offline-cost condition to hold: the online-oracle bound and $c\geq f$ give
\begin{align*}
\mathrm{Opt}\left(\mathcal P^{\mathcal I(s_i)},\bm e_{s_i:s_i+j-1}\right)
\geq (\mu_j-\beta_i)/\eta-cL
\geq T_{\mathrm c}/\eta-cL
\geq\gamma\lambda.
\end{align*}
The last inequality uses $(\eta+\epsilon)/\epsilon\geq1$, $\alpha\geq16\gamma$, $\gamma,c\geq1$, and $\lambda\leq L$. Since $C^{\mathrm{off}}_j\geq\mathrm{Opt}$, for $j_1\leq j\leq M-s_i$, continuing to $j+1$ requires $\widehat C^{\mathrm{on}}_j<T_{\mathrm c}$. Thus $J_i\geq j+1$ forces a downward estimation error:
\[
|\widehat C^{\mathrm{on}}_j-\mu_j|\geq B_{\mathrm c}/2.
\]
Since $B_{\mathrm c}=T_{\mathrm c}/18\geq15\Delta_{\mathrm c}$,
monotonicity and Eq.~\eqref{eq:cost-conservative-tail-specialized} give
$p_i(B_{\mathrm c}/2)\leq p_i(\Delta_{\mathrm c}/2)\leq\Delta_{\mathrm c}/(4T_{\mathrm c})\leq1$.
Together with $B_{\mathrm c}\geq16L$, this bounds the one-period overshoot and expected remaining cost by $L\{1+p_i(B_{\mathrm c}/2)\}\leq2L\leq B_{\mathrm c}/2$.
Hence
\begin{align*}
\mathbb E[V_i\mid\mathcal F_i]
&\leq\mu_{j_1}+L\sum_{j=j_1}^{M-s_i}
\Pr(J_i\geq j+1\mid\mathcal F_i)\\
&\leq T_{\mathrm c}+B_{\mathrm c}/2+\beta_i+L+L\,p_i(B_{\mathrm c}/2)
\leq T_{\mathrm c}+B_{\mathrm c}+\beta_i.
\end{align*}
Since $T_{\mathrm c}+B_{\mathrm c}=19\eta(\eta+\epsilon)\gamma\alpha cL/\epsilon$, this proves the upper estimate.\hfill\Halmos
\endgroup

\section{Supplementary Proofs for Section~\ref{sec:lead-time-quotation}}
\label{sec:supp-OLTQ}
\subsection{Proof of Lemma~\ref{lem:p-lead-bounded-influence}}\label{sec:proof-p-lead-bounded-influence}

\begingroup
Fix two histories $\mathcal I,\mathcal I'$ through period \(m-1\), a common continuation \(\bm e_{m:n}\), and common actions \(\bm a_{m:n}\). The histories can affect continuation reward only in slots \(m,\ldots,m+\ell-2\). There are at most \(\ell-1\) such slots, and each contributes a reward in \([0,\ell]\). Hence, recalling that \(L=\ell\),
\[
\left|\mathrm{Val}(\mathrm{OLTQ}^{\mathcal I},\bm e_{m:n},\bm a_{m:n})
-\mathrm{Val}(\mathrm{OLTQ}^{\mathcal I'},\bm e_{m:n},\bm a_{m:n})\right|
\leq \ell(\ell-1)=(\ell-1)L.
\]
Applying this inequality to an optimizer under each history gives the same bound for the optimal values, proving \((\ell-1)\)-bounded influence.

For the strongly Lipschitz condition, fix the history, continuation, and actions, and put \(e^-\defeq\min\{e_m,e_m'\}\), \(e^+\defeq\max\{e_m,e_m'\}\), and \(\Delta_e\defeq e^+-e^-\). Only slots assigned to one of the additional copies \(e^-+1,\ldots,e^+\) can change reward. At such a slot \(t=m+j\), where \(0\leq j\leq\ell-1\), the added copy either uses an unavailable slot and earns zero, earns \(\ell-j\) without displacement, or displaces a request arriving at some \(s\in\{m+1,\ldots,t\}\), producing net change \(m-s\). Thus the slot-level change lies in \([-j,\ell-j]\). If \(\Delta\) is the total change, then
\[
\Delta\leq\sum_{j=0}^{\ell-1}(\ell-j)=\frac{\ell(\ell+1)}2,
\qquad
-\Delta\leq\sum_{j=0}^{\ell-1}j=\frac{\ell(\ell-1)}2.
\]
At most \(\Delta_e\) distinct slots change, each by at most \(\ell\), so
\[
|\Delta|
\leq \ell\min\!\left\{\Delta_e,\frac{\ell+1}{2}\right\}
=L\min\!\left\{d(e_m,e_m'),\frac{\ell+1}{2}\right\}.
\]
This proves that OLTQ is \(\left(1,\frac{\ell+1}{2}\right)\)-strongly Lipschitz.\hfill\Halmos
\endgroup

\subsection{Proof of Lemma~\ref{lem:O-HRR-star-offline-oracle}}
\label{sec:proof-O-HRR-star-offline-oracle}
\begingroup
Fix \(m,n,\mathcal I\), and an admissible continuation as in the offline-oracle construction of Section~\ref{sec:lead-time-quotation}, and let \(T=T_{m,n}^{\mathrm{off}}\). History-reserved slots are unavailable and contribute a fixed reward. At any other multiply assigned slot, rejecting all but the priority winner preserves value. Thus the residual problem is a finite maximum-weight bipartite matching: its left vertices are continuation copies, its right vertices are available slots in \(\{m+1,\ldots,T\}\), and a copy arriving at \(s\) may use \(t\in[s,s+\ell-1]\) with weight \(w(s,t)=s+\ell-t\). Finiteness is immediate when \(n<\infty\) and follows from admissibility when \(n=\infty\).

We process the available slots through \(T\) in increasing order while maintaining an optimal matching that agrees with O-HRR$^*$ on earlier slots. If O-HRR$^*$ leaves \(t\) idle, every copy feasible at \(t\) has already been assigned to an earlier agreed slot, so the maintained matching also leaves \(t\) idle. Otherwise let \(g\), arriving at \(s_g\), be the copy selected by O-HRR$^*$. If the maintained matching already assigns \(g\) to \(t\), leave it unchanged and proceed to the next slot. Otherwise, let \(h\) be the copy assigned by that matching to \(t\), if any, and let \(t_g>t\) be the slot assigned to \(g\), if any. The greedy choice gives \(s_h\leq s_g\) whenever \(h\) exists.

We move \(g\) to \(t\). If \(h\) is absent, reward weakly increases; if \(h\) exists but \(g\) was unscheduled, the gain is \(s_g-s_h\geq0\). If both \(h\) and \(t_g\) exist and \(h\) is feasible at \(t_g\), swapping them preserves their combined reward. Otherwise \(s_h\leq s_g\leq t<t_g\), and infeasibility implies \(t_g\geq s_h+\ell\); dropping \(h\) then changes reward by
\[
w(s_g,t)-w(s_h,t)-w(s_g,t_g)=t_g-(s_h+\ell)\geq0.
\]
Thus an optimal matching can be made to agree with O-HRR$^*$ at \(t\). Iteration proves that O-HRR$^*$ solves Eq.~\eqref{eq:lead-time-1-oracle-problem} and is an exact offline oracle.\hfill\Halmos
\endgroup

\subsection{Proof of Lemma~\ref{lem:Q-FRAC-star-online-oracle}}
\label{sec:proof-Q-FRAC-eta}
\begingroup
We first prove the no-history guarantee for complete phases and finite endpoints, then account for inherited reservations after a fixed history.
Write \(k_{\mathrm Q}\defeq k_\ell^*\), \(h_{\mathrm Q}\defeq\ell-k_{\mathrm Q}+1\),
\[
A\defeq\sum_{j=k_{\mathrm Q}}^{\ell}j,
\qquad
B\defeq\sum_{j=1}^{\ell}j.
\]
The update maintains \(t\leq U_t\) and \(U_t+N_t-1\leq t+\ell-k_{\mathrm Q}\), so each scheduled copy earns at least \(k_{\mathrm Q}\).

\medskip
\noindent\underline{Complete phases.}
First consider no fixed history. Partition the slots filled by the internal schedule into maximal consecutive runs, called phases. A phase's cohort comprises the copies arriving during that phase. Because the internal schedule fills consecutive slots, each scheduled copy is processed in the phase in which it arrives, and any positive arrival in an otherwise idle period starts a phase. Thus the cohorts partition all copies. Let \(Z_i^{\mathrm Q}\) be the phase reward under Q-FRAC$^*$ and \(Z_i^*\) the maximum reward obtainable from its cohort alone.

If no cohort copy is rejected, Q-FRAC$^*$ schedules every copy, so \(Z_i^{\mathrm Q}\geq(k_{\mathrm Q}/\ell)Z_i^*\). Otherwise, let \(r_i\) be the phase start, \(s_i\) its last rejection period, and \(C\) the number of cohort copies arriving after \(s_i\); all \(C\) are scheduled. At the last rejection, the update gives \(U_{s_i}+N_{s_i}=s_i+h_{\mathrm Q}\), so the occupied prefix ends at \(s_i+h_{\mathrm Q}-1\), and the \(C\) later copies use slots beyond it.

Every copy processed in this phase arrived at or after \(r_i\), so the copy in slot \(r_i+j\) earns at least \(\ell-j\) for \(0\leq j<h_{\mathrm Q}\). These first \(h_{\mathrm Q}\) slots contribute at least \(A\); the remaining \(s_i-r_i+C\) copies each earn at least \(k_{\mathrm Q}\). With \(z=s_i-r_i+C\),
\[
Z_i^{\mathrm Q}\geq A+k_{\mathrm Q}z.
\]
For copies arriving by \(s_i\), an offline schedule earns at most \(\ell\) in each of the \(s_i-r_i\) slots before \(s_i\), followed by at most \(\ell,\ell-1,\ldots,1\). Each of the \(C\) later copies contributes at most \(\ell\). Hence
\[
Z_i^*\leq B+\ell z.
\]
When rejection occurs, \((A+k_{\mathrm Q}z)/(B+\ell z)\) is a convex combination of \(A/B\) and \(k_{\mathrm Q}/\ell\). Thus, in either case,
\[
\frac{Z_i^{\mathrm Q}}{Z_i^*}
\geq\min\left\{\frac{k_{\mathrm Q}}{\ell},\frac{A}{B}\right\}
=\min\left\{\frac{k_{\mathrm Q}}{\ell},g_\ell(k_{\mathrm Q})\right\}
=\eta^{\mathrm{OLTQ}}.
\]

The equivalent formula for $\eta^{\mathrm{OLTQ}}$ displayed in Section~\ref{sec:lead-time-quotation} follows by extending \(g_\ell\) to \([0,\ell]\):
\[
g_\ell(x)-\frac{x}{\ell}
=\frac{\ell(\ell+1)-x(x+\ell)}{\ell(\ell+1)}
\]
has its unique nonnegative zero at \(x=\theta_\ell^*\ell\). The term \(j/\ell\) increases in \(j\), whereas \(g_\ell(j)\) decreases. Hence the objective defining \(k_\ell^*\) is maximized by the better of \(\lfloor\theta_\ell^*\ell\rfloor\) and \(\lceil\theta_\ell^*\ell\rceil\), giving the displayed formula.

\medskip
\noindent\underline{Finite endpoints.}
Fix a finite continuation endpoint \(N\). An offline schedule may interleave cohorts, but its reward from each completed phase's cohort is at most \(Z_i^*\). If \(N\) cuts phase \(i\), Q-FRAC$^*$ fills all \(x=N-r_i+1\) slots from its start \(r_i\) through \(N\), earning at least \(k_{\mathrm Q}x\). Since that cohort has no arrivals before \(r_i\), any schedule earns at most \(\ell x\) from it by \(N\). Combining \(k_{\mathrm Q}/\ell\geq\eta^{\mathrm{OLTQ}}\) with the bounds for complete phases and summing by cohort proves the guarantee through \(N\). Admissibility reduces an infinite continuation to a finite endpoint. This proves the first online-oracle condition and, for \(m=0\), the restarted condition because the policy equals \(\bm\pi_0\).

\medskip
\noindent\underline{Fixed histories.}
Fix \(m\geq1\), \(n\in\mathbb Z_{\geq0}\cup\{\infty\}\), a history \(\mathcal I=(\bm e_{1:m},\bm a_{1:m})\), and an admissible continuation \(\bm e_{m+1:m+n}\). Let \(\widehat V(\bm p)\) be the reward of a continuation schedule \(\bm p\) in the virtual no-history instance beginning at \(m+1\), and let \(\widehat{\mathrm{Opt}}=\max_{\bm p}\widehat V(\bm p)\). Time shifting and the no-history result give, for the action sequence \(\bm a\) generated by restarted Q-FRAC$^*$,
\[
\widehat V(\bm a)\geq\eta^{\mathrm{OLTQ}}\widehat{\mathrm{Opt}}.
\]

Let \(S\) be the at most \(\ell-1\) slots in
\(\{m+1,\ldots,\min\{T_{m,n}^{\mathrm{off}},m+\ell-1\}\}\) reserved by history copies, and let \(J\) be the total reward credited in those slots to the fixed-history copies. If \(\widehat V_S(\bm p)\) is the virtual continuation reward in \(S\), each reserved slot replaces that reward by its fixed history reward, while rewards outside \(S\) agree. Hence
\[
\mathrm{Val}(\mathrm{OLTQ}^{\mathcal I},\bm e_{m+1:m+n},\bm p)
=J+\widehat V(\bm p)-\widehat V_S(\bm p).
\]
Moreover, \(\widehat V_S(\bm a)\leq\ell|S|\) and
\(\mathrm{Opt}(\mathrm{OLTQ}^{\mathcal I},\bm e_{m+1:m+n})
\leq J+\widehat{\mathrm{Opt}}\). Hence
\begin{align*}
\mathrm{Val}(\mathrm{OLTQ}^{\mathcal I},\bm e_{m+1:m+n},\bm\pi_m^{\mathcal I})
&\geq J+\eta^{\mathrm{OLTQ}}\widehat{\mathrm{Opt}}-\ell|S|\\
&\geq\eta^{\mathrm{OLTQ}}
\mathrm{Opt}(\mathrm{OLTQ}^{\mathcal I},\bm e_{m+1:m+n})
 +(1-\eta^{\mathrm{OLTQ}})J-\ell|S|\\
&\geq\eta^{\mathrm{OLTQ}}
\mathrm{Opt}(\mathrm{OLTQ}^{\mathcal I},\bm e_{m+1:m+n})
-\ell(\ell-1).
\end{align*}
Since \(f=\ell-1\) and \(L=\ell\), the additive term is \(fL\), proving both online-oracle conditions.\hfill\Halmos
\endgroup

\subsection{A Hybrid of AdaSwitch-OLTQ and Q-FRACwP}
\label{sec:strengthened-adaswitch-oltq}
\begingroup
\begin{proposition}
\label{prop:strengthened-adaswitch-oltq}
Fix the class of sequences with a known arrival horizon and \(r\in(0,\eta^{\mathrm{OLTQ}})\) from Section~\ref{sec:lead-time-quotation}. If \(\beta(r)=1\), define Strengthened AdaSwitch-OLTQ to be Q-FRACwP. If \(\beta(r)<1\), define
$
\Lambda_r\defeq
\frac{10\ell^2}{(\eta^{\mathrm{OLTQ}}-r)(1-\beta(r))}.
$
Compute \(\mathrm{Opt}(\mathrm{OLTQ},\bm e^*_{1:\infty})\) using O-HRR$^*$. Invoke AdaSwitch-OLTQ if this value is at least \(\Lambda_r\) and Q-FRACwP otherwise. In either case, Strengthened AdaSwitch-OLTQ is \(r\)-robust and \(\beta(r)\)-consistent. When \(\beta(r)<1\), its perfect-prediction bound is
\[
\max\left\{\beta(r),
1-\frac{10\ell^2}{(\eta^{\mathrm{OLTQ}}-r)\mathrm{Opt}(\mathrm{OLTQ},\bm e^*_{1:\infty})}
\right\},
\]
which is strictly larger than \(\beta(r)\) whenever the predicted optimum exceeds \(\Lambda_r\). Still assuming \(\beta(r)<1\), let \(\mathrm{Comp}_{\mathrm S}\) denote the instancewise performance ratio of Strengthened AdaSwitch-OLTQ for the given realized sequence and prediction. Then
\[
\mathrm{Comp}_{\mathrm S}\geq
\begin{cases}
\displaystyle \max\left\{r,1-\frac{\ell(10\ell+7\eta^{\mathrm{OLTQ}}\varphi_M^*)}{(\eta^{\mathrm{OLTQ}}-r)\mathrm{Opt}(\mathrm{OLTQ},\bm e_{1:\infty})}\right\},
& \text{if }\mathrm{Opt}(\mathrm{OLTQ},\bm e^*_{1:\infty})\geq\Lambda_r,\\[2ex]
r, & \text{otherwise.}
\end{cases}
\]
\end{proposition}

\noindent\textit{Proof.}
If \(\beta(r)=1\), Q-FRACwP gives all claims. Suppose \(\beta(r)<1\). With \(\epsilon=\eta^{\mathrm{OLTQ}}-r\), Theorem~\ref{thm:competitive-ratio-adaswitch-OLTQ} gives \(r\)-robustness and the displayed error-dependent bound on the AdaSwitch branch; Q-FRACwP gives \(r\)-robustness, \(\beta(r)\)-consistency, and the second piecewise branch. Under perfect prediction, \(\varphi_M^*=0\) and the optima coincide; the AdaSwitch bound exceeds \(\beta(r)\) whenever the predicted optimum exceeds \(\Lambda_r\).\hfill\Halmos
\endgroup

\section{Supplementary Proofs for Section~\ref{sec:k-server}}
\begingroup

\subsection{Proof of Lemma~\ref{lem:k-server-finite-influence}}
\label{sec:proof-k-server-finite-influence}

\noindent\textbf{Fixed-history offline oracle.}
Let $\bm S^{\mathcal I}$ be the configuration after history $\mathcal I$ from initial configuration $\bm S$.
\begin{observation}
\label{obs:k-server-start-middle-equal}
After reindexing, the residual problem $k\mathrm{SE}_{\bm S}^{\mathcal I}$ coincides with $k\mathrm{SE}_{\bm S^{\mathcal I}}$.
\end{observation}
After deleting $\bot$ periods, apply the acyclic minimum-cost-flow reduction~\citep{chrobak1991new,tarjan1983data} with initial configuration $\bm S^{\mathcal I}$ to obtain an exact fixed-history offline oracle under the finite-input convention of Section~\ref{sec:k-server}.

\begin{lemma}
\label{lem:k-server-different-initial-state}
For any $N\in\mathbb Z_{\geq0}\cup\{\infty\}$, configurations $\bm S,\bm S'\in X^k$, and request sequence $\bm e_{1:N}\in\bm E_{1:N}$ under the standing admissibility convention, we have 
$
\left|\mathrm{Opt}(k\mathrm{SE}_{\bm S},\bm e_{1:N})
-\mathrm{Opt}(k\mathrm{SE}_{\bm S'},\bm e_{1:N})\right|
\leq \sum_{i=1}^k d(S_i,S_i').
$
Here $\bm e_{1:0}$ denotes the empty sequence.
\end{lemma}

\noindent\textbf{Proof of Lemma~\ref{lem:k-server-different-initial-state}.}
Fix a common labeled action sequence and a finite prefix endpoint $q\leq N$. Let $t_i$ be the first period through $q$ in which server $i$ serves a real request, with $t_i=\infty$ if there is none. Server $i$'s trajectories in the two coupled runs coincide after it serves at $t_i$, so
\[
\mathrm{Val}(k\mathrm{SE}_{\bm S},\bm e_{1:q},\bm a_{1:q})
-\mathrm{Val}(k\mathrm{SE}_{\bm S'},\bm e_{1:q},\bm a_{1:q})
=\sum_{i:t_i<\infty}\left[d(S_i,e_{t_i})-d(S_i',e_{t_i})\right]
\leq\sum_{i=1}^k d(S_i,S_i').
\]
Interchanging $\bm S$ and $\bm S'$ gives the reverse actionwise inequality. If $N<\infty$, take $q=N$ and optimize in both directions. If $N=\infty$, apply the two inequalities to every finite prefix of a common infinite action sequence. Because the costs are nonnegative, their prefix values converge monotonically to their infinite-sequence values; passing to the limit and then taking minima over the common action space gives the two optimal-value inequalities.
\hfill\Halmos

The one-period cost bound is $L=1$. Consider histories $\mathcal I,\mathcal I'$ ending at the same time, let $\bm S^{\mathcal I},\bm S^{\mathcal I'}$ be their resulting configurations, and reindex a common continuation as $\widetilde{\bm e}_{1:N}$. Observation~\ref{obs:k-server-start-middle-equal} and Lemma~\ref{lem:k-server-different-initial-state} give
\[
\left|\mathrm{Opt}(k\mathrm{SE}_{\bm S^{\mathcal I}},\widetilde{\bm e}_{1:N})
-\mathrm{Opt}(k\mathrm{SE}_{\bm S^{\mathcal I'}},\widetilde{\bm e}_{1:N})\right|
\leq\sum_{i=1}^k d(S_i^{\mathcal I},S_i^{\mathcal I'})
\leq k=kL.
\]
Thus $k\mathrm{SE}_{\bm S}$ has $k$-bounded influence.

For the Lipschitz condition, reindex the subproblem after a fixed history so that the perturbed request is first. Let $\bm S$ be the resulting configuration, let $x,y\in X$ be the two first requests, and let $\bm q$ be their common, possibly empty suffix. Choose an optimum for $y\circ\bm q$, and let $a$ be its first server label. Define
\[
c_a(z)\defeq\mathbb I(z\neq\bot)d(S_a,z),
\qquad
u_a(z)\defeq
\begin{cases}
z,&z\neq\bot,\\
S_a,&z=\bot,
\end{cases}
\]
and let $\bm S(z)$ replace coordinate $a$ of $\bm S$ by $u_a(z)$. Because $d\leq1$ and $d(z,\bot)=1$ for $z\neq\bot$, both $c_a$ and $u_a$ are $1$-Lipschitz. Optimality gives the exact first-period decomposition for the $y$-instance, while taking server $a$ first and then optimizing the suffix is feasible for the $x$-instance. Lemma~\ref{lem:k-server-different-initial-state} therefore yields
\[
\begin{aligned}
&\mathrm{Opt}(k\mathrm{SE}_{\bm S},x\circ\bm q)
-\mathrm{Opt}(k\mathrm{SE}_{\bm S},y\circ\bm q)\\
&\quad\leq c_a(x)-c_a(y)
+\mathrm{Opt}(k\mathrm{SE}_{\bm S(x)},\bm q)
-\mathrm{Opt}(k\mathrm{SE}_{\bm S(y)},\bm q)\\
&\quad\leq |c_a(x)-c_a(y)|+d(u_a(x),u_a(y))
\leq2d(x,y).
\end{aligned}
\]
Interchanging $x$ and $y$ proves the absolute bound, and $L=1$ and $d(x,y)\leq1$ give $2d(x,y)=L\min\{2d(x,y),2\}$. Hence the problem is $(2,2)$-Lipschitz.
\hfill\Halmos

\subsection{A WFA Policy Allowing Co-Located Servers}
\label{sec:k-server-WFA}

We construct one policy satisfying both the additive and strict WFA bounds when initial server locations may repeat. We first specify lazy WFA on set configurations, then separate co-located servers, pass to a limiting policy, and verify the fixed-history online-oracle conditions.

\medskip
\noindent\underline{Lazy WFA on set configurations.}
We use the standard set-configuration work function and lazy WFA of \citet[Section~2]{emek2010additive}. Let $\bm S_t=(S_{t,1},\ldots,S_{t,k})$ list the distinct server locations after period $t$, with physical labels $1,\ldots,k$, and let $\bm C^{j\to e}$ replace coordinate $j$ by $e$. For a finite real-request sequence $\bm r$, the work function $W_{\bm r}(\bm C)$ is the minimum cost of serving $\bm r$ from $\bm S_0$ and ending at the terminal set $\bm C$, allowing terminal repositioning. Work-function values ignore physical labels.

Let $\bm r^{<t}$ be the real-request subsequence of $\bm e_{1:t-1}$. On a $\bot$ period, this auxiliary WFA does nothing; on a real hit, it chooses the unique resident server without moving. On a real miss, the source-point selector specified below chooses a move satisfying
\[
a_t\in\arg\min_{j\in[k]}
\left\{W_{\bm r^{<t}}(\bm S_{t-1}^{j\to e_t})+d(S_{t-1,j},e_t)\right\},
\qquad \bm S_t=\bm S_{t-1}^{a_t\to e_t}.
\]

\begin{lemma}[Additive and strict bounds]
\label{lem:wfa-multiset-strict}
Under the standing well-posedness and measurability convention of Section~\ref{sec:ec-framework-conventions}, assume in particular a fixed jointly measurable selector for the nonempty finite sets of minimizing source points used below, common to all $\delta>0$. Let $(X,d)$ be a metric space and let $\bm C=(C_1,\ldots,C_k)\in X^k$, with repetitions allowed. Let $k\mathrm{SE}_{\bm C}$ denote the standard $k$-server problem on $(X,d)$ with initial configuration $\bm C$. The construction below defines a deterministic, measurable, nonanticipating $k$-server policy $\mathrm{WFA}^{\mathrm{lim}}_{\bm C}$ that moves at most one server per request and satisfies, for every finite real-request sequence $\bm q$,
\[
\mathrm{Val}(k\mathrm{SE}_{\bm C},\bm q,\mathrm{WFA}^{\mathrm{lim}}_{\bm C})
\leq\min\left\{
(2k-1)\,\mathrm{Opt}(k\mathrm{SE}_{\bm C},\bm q)+2\mathrm{Disp}(\bm C),
(4k-2)\,\mathrm{Opt}(k\mathrm{SE}_{\bm C},\bm q)
\right\}.
\]
\end{lemma}

\noindent\textbf{Proof.}
The empty-sequence bound is immediate. For $k=1$, define $\mathrm{WFA}^{\mathrm{lim}}_{\bm C}$ to be the sole-server policy, which is optimal. Hence assume $k\geq2$.

\medskip
\noindent\underline{Bounds on the lifted space.}
For $\delta>0$, define
\[
X_\delta\defeq X\times\{0,1,\ldots,k\},
\qquad d_\delta((x,u),(y,v))\defeq d(x,y)+\delta\mathbb I(u\neq v),
\]
and lift the initial state and requests by
\[
\bm C^\delta\defeq((C_1,1),\ldots,(C_k,k)),
\qquad q_t^\delta\defeq(q_t,0).
\]
Then $d_\delta$ is a metric and $\bm C^\delta$ is a $k$-element set. Run the above WFA on the full space $X_\delta$ from $\bm C^\delta$, with its work vector indexed by all $k$-element subsets, and evaluate it on $\bm q^\delta$.

Keep each server's physical label $i\in[k]$ separate from its auxiliary coordinate: the label identifies the action, whereas the coordinate separates lifted points. A candidate move's \emph{source point} is its origin in the current lifted set. The selector assumed in the lemma depends only on the set of minimizing source points and is common to all $\delta>0$. Move the server at the selected point and use its physical label as the action. Thus physical labels do not affect WFA's decisions on set configurations, as needed for the strictification argument below.

Let $A_\delta(\bm q)$ be the lifted WFA cost and $O_\delta(\bm q)$ the lifted set-configuration optimum. The proof of Theorem~4.3 in \citet{koutsoupias1995k}, using Lemma~4.2 and Fact~5 there, gives
\begin{equation}
A_\delta(\bm q)
\leq(2k-1)O_\delta(\bm q)+2\mathrm{Disp}(\bm C)+k(k-1)\delta,
\label{eq:wfa-separated-additive}
\end{equation}
because the lifted initial dispersion is $\mathrm{Disp}(\bm C)+\binom{k}{2}\delta$. This WFA is lazy, request-sequence-oblivious, and invariant under common additive shifts of its work vector, so Theorem~1 and its extension in Section~3 of \citet{emek2010additive} give
\begin{equation}
A_\delta(\bm q)\leq(4k-2)O_\delta(\bm q).
\label{eq:wfa-separated-strict}
\end{equation}

\medskip
\noindent\underline{The limiting policy.}
To define a single nonanticipating policy, we show that the lifted trajectory and serving labels stabilize as $\delta\downarrow0$ on every fixed finite request history. Induct on the history length. Once the preceding labeled configuration has stabilized, a hit has a fixed unique serving server. On a miss, each candidate work-function value is attained by a lazy trajectory followed by terminal repositioning~\citep[Section~2]{emek2010additive}. For the fixed history and candidate terminal configuration, there are only finitely many serving-label sequences and terminal matchings, even if $X$ is infinite. Each resulting cost, including the candidate move, is affine in $\delta$. Thus each candidate score is a minimum of finitely many affine functions. The weak ordering of these scores, and hence the set of minimizing source points, is constant on some interval $(0,\delta_0)$. The fixed selector then chooses the same source point and physical label, completing the induction.

Set $\delta_\nu\defeq1/\nu$. For a nonempty real-request history $\bm q_{1:t}$, let $a^{\mathrm{lim}}(\bm C,\bm q_{1:t})$ be its eventual serving label along $\{\delta_\nu\}_{\nu\geq1}$, and define $\mathrm{WFA}^{\mathrm{lim}}_{\bm C}$ to use this label. On a $\bot$ period, choose label $1$ and leave the real-request history and configuration unchanged.

The induction makes these choices consistent across prefixes. Projection therefore gives a feasible nonanticipating policy on $X$ that moves at most one server per request: a lifted miss moves the selected server to the requested base point, even when another lifted server has the same base location. Each $\delta_\nu$-action map is jointly measurable by this rule, and its finite-valued pointwise eventual limit is jointly measurable.

\medskip
\noindent\underline{Passing the bounds to the limit.}
For the algorithmic cost, fix $\bm q=(q_1,\ldots,q_s)$ and let $A_0(\bm q)$ be the projected policy's cost. After simultaneous stabilization along its prefixes, each lifted move exceeds its projected cost by at most $\delta$. Thus, for all sufficiently small $\delta$, we have
$
0\leq A_\delta(\bm q)-A_0(\bm q)\leq s\delta,
$
and hence $A_\delta(\bm q)\to A_0(\bm q)$.

For the offline optimum, let $O_0(\bm q)$ be the original multiset optimum. Projecting a lazy lifted optimum cannot increase cost, so $O_0(\bm q)\leq O_\delta(\bm q)$. Conversely, take a labeled lazy original optimum, assign any resident serving label on a hit, and couple server $i$ with the lifted server initialized at $(C_i,i)$. Whenever $(q_t,0)$ is unoccupied, move the lifted partner of the original serving label there. This preserves equal base locations and set feasibility, with the same base-movement cost as the original optimum. Each lifted server changes its auxiliary coordinate only on its first move, from $i$ to $0$; every later move remains in coordinate $0$. Thus the total extra cost is at most $k\delta$, giving
\begin{equation*}
O_0(\bm q)\leq O_\delta(\bm q)\leq O_0(\bm q)+k\delta.
\end{equation*}
The last display gives $O_\delta(\bm q)\to O_0(\bm q)$. Together with $A_\delta(\bm q)\to A_0(\bm q)$, letting $\delta\downarrow0$ in Eqs.~\eqref{eq:wfa-separated-additive} and~\eqref{eq:wfa-separated-strict} proves the lemma.
\hfill\Halmos

\medskip
\noindent\underline{Fixed-history online oracles.}
Fix $m\geq0$, a history $\mathcal I=(\bm e_{1:m},\bm a_{1:m})$, and either a finite continuation or an admissible infinite continuation $\bm e_{m+1:m+n}$. Restart $\mathrm{WFA}^{\mathrm{lim}}$ from the history-induced configuration $\bm S^{\mathcal I}$, which may have co-located servers. Observation~\ref{obs:k-server-start-middle-equal} identifies the residual problem with $k\mathrm{SE}_{\bm S^{\mathcal I}}$. For finite $n$, delete all $\bot$ periods. For $n=\infty$, first truncate at
$T\defeq\max\{m,M(\bm e_{1:m}\circ\bm e_{m+1:\infty})\}<\infty$
and then delete them. These operations preserve both continuation values, so Lemma~\ref{lem:wfa-multiset-strict} gives both bounds from $\bm S^{\mathcal I}$ for every continuation under consideration. Since $2\mathrm{Disp}(\bm S^{\mathcal I})\leq k(k-1)\leq\eta_{\mathrm{add}}fL$, its additive bound satisfies the restarted condition for $m\geq1$ and the initial condition with $\beta_{\mathrm{add}}=2\mathrm{Disp}(\bm S)$. Its strict bound satisfies both conditions with $\beta_{\mathrm{str}}=0$. Thus this fixed-history policy family is an $(\eta_{\mathrm{add}},\beta_{\mathrm{add}})$-online oracle and an $\eta_{\mathrm{str}}$-online oracle.

\subsection{Proof of Lemma~\ref{lem:strict-marking-bound}}
\label{sec:proof-marking-strict-online-oracle}

We bound Marking's expected phase cost, compare the total compulsory misses with the offline optimum, and then apply the finite-sequence bound after each fixed history.

\medskip
\noindent\underline{Caching and phase conventions.}
In the uniform-metric specialization, we may restrict attention to lazy policies, which leave the cache unchanged on a hit and replace one slot on a miss, because moving on a hit is dominated. A fixed history may leave duplicates; $\bot$ entries and all but one copy of each duplicated real page are treated as empty slots. Fresh-start Marking keeps all marks clear through the initial run of hits and starts its first phase at the first miss. Within a phase, the algorithm marks each requested page. On a miss, it fills an empty slot if one is available; otherwise, it evicts an unmarked page chosen uniformly at random. If a miss occurs when the cache is full and all $k$ pages are marked, it first clears the marks and starts a new phase.

Fix a finite request sequence $\bm e_{1:n}$ and delete all $\bot$ requests. An optimal lazy schedule can also be chosen empty-first, since filling an empty slot weakly dominates evicting a uniquely resident real page. Replace each empty slot by a distinct auxiliary dummy page that is never requested. Under the empty-first rule, identifying each dummy with its original slot until overwritten gives a cost-preserving correspondence between augmented and original schedules. Hence their offline optima coincide, and Marking and an optimal lazy schedule $A$ start from the same $k$-element augmented cache.

Delete the maximal initial hit prefix: both caches remain unchanged, all Marking marks remain clear, and both costs are zero. If no request remains, the result is immediate. Partition the remainder greedily into maximal blocks containing at most $k$ distinct pages, and let $P$ be the number of resulting phases. For phase $i\in[P]$, write $B_i$ for its requested-page set, $K_i^{\mathrm M},K_i^A$ for the augmented caches at its start, and $C_i^{\mathrm M},C_i^A$ for the corresponding phase costs. Every phase begins with a Marking miss. Every phase $i<P$ contains exactly $k$ distinct pages and ends with these pages marked and cached, so $K_{i+1}^{\mathrm M}=B_i$.

\medskip
\noindent\underline{Expected phase cost.}
With $\mathcal D$ denoting the dummy-page set, define
\[
q_i\defeq|B_i\setminus K_i^{\mathrm M}|,
\qquad z_i\defeq|K_i^{\mathrm M}\cap\mathcal D|.
\]
Since $K_1^{\mathrm M}$ is the fixed initial cache and $K_i^{\mathrm M}=B_{i-1}$ for $i\geq2$, the quantities $q_i$ and $z_i$ are deterministic despite the within-phase random evictions.
The phase construction gives $q_i\geq1$. Set $x_i\defeq[q_i-z_i]_+$. The first request to each of the $q_i$ phase pages absent from $K_i^{\mathrm M}$ is a compulsory miss. Marking replaces a dummy whenever one remains, so exactly $x_i$ of these misses replace real pages.

If $x_i=0$, no additional miss occurs. Otherwise, consider the initially resident real pages not yet requested in the phase. Each compulsory miss that replaces a real page increases the number absent by one; an additional miss restores its requested page and evicts another, leaving that number unchanged. Uniform eviction preserves symmetry among these pages. Before the $j$th first request to an initially present phase page, there are $k-z_i-j+1$ such pages, at most $x_i$ of them absent. Thus the miss probability is at most $x_i/(k-z_i-j+1)$. There are at most $k-q_i$ such first requests. Reindexing and padding with nonnegative terms gives, also for $x_i=0$ and the incomplete final phase,
\[
\mathbb E[C_i^{\mathrm M}]
\leq q_i+\sum_{s=1}^{k-q_i}\frac{x_i}{x_i+s}
\leq q_i+\sum_{s=1}^{k-q_i}\frac{q_i}{q_i+s}
=q_i(1+H_k-H_{q_i})
\leq q_iH_k.
\]

\medskip
\noindent\underline{Offline comparison.}
The cache discrepancy $\chi_i\defeq|K_i^{\mathrm M}\setminus K_i^A|=|K_i^A\setminus K_i^{\mathrm M}|$ bounds the offline cache's possible advantage at phase start. At most $\chi_i$ of the $q_i$ pages absent from $K_i^{\mathrm M}$ are present in $K_i^A$, so $C_i^A\geq q_i-\chi_i$ for every phase. For a complete phase $i<P$, each of the $\chi_{i+1}$ pages in $K_{i+1}^A\setminus B_i$ is unrequested in phase $i$ and, by laziness, was already in $K_i^A$. Thus $|B_i\cap K_i^A|\leq k-\chi_{i+1}$; since $|B_i|=k$, this gives $C_i^A\geq\chi_{i+1}$. Adding the two lower bounds handles complete phases; the possibly incomplete final phase uses only the first:
\[
q_i\leq2C_i^A+\chi_i-\chi_{i+1}\quad(i<P),
\qquad q_P\leq C_P^A+\chi_P.
\]
Summing cancels the discrepancy terms. The initial caches coincide, so $\chi_1=0$ and no additive term remains:
\[
\sum_{i=1}^Pq_i\leq2\sum_{i=1}^{P-1}C_i^A+C_P^A
\leq2\sum_{i=1}^PC_i^A.
\]
Combining $\mathbb E[C_i^{\mathrm M}]\leq q_iH_k$ with the telescoping bound above proves
\[
\mathrm{Val}(\mathrm{CA}_{\bm S},\bm e_{1:n},\mathrm{Mark}^{\mathrm{fs}})
=\sum_{i=1}^P\mathbb E[C_i^{\mathrm M}]
\leq2H_k\sum_{i=1}^PC_i^A
=2H_k\,\mathrm{Opt}(\mathrm{CA}_{\bm S},\bm e_{1:n}).
\]
\medskip
\noindent\underline{Fixed-history online oracles.}
Fix $m\geq0$, a history $\mathcal I=(\bm e_{1:m},\bm a_{1:m})$, and either a finite continuation or an admissible infinite continuation $\bm e_{m+1:m+n}$. Observation~\ref{obs:k-server-start-middle-equal} identifies the continuation with caching from the resulting cache. For finite $n$, the preceding bound applies after reindexing. If $n=\infty$, set $T\defeq\max\{m,M(\bm e_{1:m}\circ\bm e_{m+1:\infty})\}<\infty$. All costs after $T$ vanish, so the Marking and offline continuation values equal their values on the finite, possibly empty continuation $\bm e_{m+1:T}$, to which the finite bound applies. Hence fresh-start Marking satisfies both online-oracle conditions with factor $2H_k$ and no additive term; $H_k\leq1+\ln k$ permits $\eta^{\mathrm{CA}}=2(\ln k+1)$.
\hfill\Halmos

\endgroup

\section{Supplementary Proofs for Section~\ref{sec:ORRA}}
\label{sec:proof-ORRA}
\noindent\textbf{Effective horizon.}
Fix a finite-support request sequence $\bm e'$. If $\bm e'\equiv\bm0$, both $M(\bm e')$ and $\widetilde M(\bm e')$ are zero. Otherwise rewards vanish after the last nonzero eligibility vector, while rejecting every earlier request leaves every resource available and makes a positive reward feasible in that final nonzero period. Hence $M(\bm e')=\widetilde M(\bm e')$. Applying this equality to every realized-prefix/predicted-suffix splice verifies the effective-horizon estimator in Section~\ref{sec:ORRA}.

For a request--action history, define the vector of next available periods by
$W_t=W_t^{\mathrm{ORRA}_n}(\bm e_{1:t},\bm a_{1:t})\in\mathbb Z_+^n$.
Set $W_0=(1,\ldots,1)$ and, for every $i\in[n]$, define
\begin{equation}
W_t(i)\defeq
\begin{cases}
t+D,&\text{if }W_{t-1}(i)\leq t,\ a_t=i,\text{ and }e_t(i)=1,\\
W_{t-1}(i),&\text{otherwise}.
\end{cases}
\label{eq:ORRA-waiting-update}
\end{equation}
Thus resource $i$ is available in period $t$ exactly when $W_{t-1}(i)\leq t$, and the period-$t$ reward is
\begin{align*}
R_t^{\mathrm{ORRA}_n}(\bm e_{1:t},\bm a_{1:t})
\defeq
\sum_{i=1}^{n}
\mathbb I(a_t=i)\,e_t(i)\,
\mathbb I\!\left(W_{t-1}(i)\leq t\right).
\end{align*}

\subsection{The Fixed-History Offline Dynamic Program}
\label{sec:proof-ORRA-fixed-history-DP}

Fix a history $\mathcal I=(\bm e_{1:h},\bm a_{1:h})$ and write a length-$T$ continuation using local indices $1,\ldots,T$, with $\bar e_t\defeq e_{h+t}$ denoting the request in local period $t$. For every $\tau\in[h]$, let $x_\tau^{\mathcal I}$ be the resource successfully assigned in calendar period $\tau$, setting $x_\tau^{\mathcal I}=0$ if no assignment succeeds. Also set $x_\tau^{\mathcal I}=0$ for $\tau\leq0$. At the start of the continuation, initialize
\[
s^{\mathcal I}
\defeq
(x_{h-D+2}^{\mathcal I},\ldots,x_h^{\mathcal I})
\in(\{0\}\cup[n])^{D-1},
\]
with $s^{\mathcal I}=()$ when $D=1$. More generally, a state $s=(s_1,\ldots,s_{D-1})$ records, from oldest to newest, the successfully assigned resource in each of the preceding $D-1$ calendar periods, using $0$ if none. Let $\operatorname{Occ}(s)$ be the set of nonzero entries of $s$, with $\operatorname{Occ}(())=\varnothing$. For $D\geq2$, set $\sigma(s,i)=(s_2,\ldots,s_{D-1},i)$; for $D=1$, set $\sigma((),i)=()$. Define
\[
V_{T+1}(s)=0,\qquad
V_t(s)=
\max_{i\in\{0\}\cup\{j\in[n]:\bar e_t(j)=1,\ j\notin\operatorname{Occ}(s)\}}
\bigl\{\mathbb I(i\neq0)+V_{t+1}(\sigma(s,i))\bigr\}.
\]

\begin{proposition}[Exact fixed-history dynamic program]
\label{prop:ORRA-fixed-history-DP}
For every length-$h$ fixed history and every length-$T$ finite continuation, the dynamic program above returns an exact offline-optimal continuation. It forms $s^{\mathcal I}$ using $O(h+D)$ history-processing steps; thereafter it uses at most $(n+1)^{D-1}$ states per continuation period and examines $O(T(n+1)^D)$ state--action pairs.
\end{proposition}
For an infinite continuation, finite support gives $M=M(\bm e_{1:h}\circ\bm e_{h+1:\infty})<\infty$. Truncation at calendar endpoint $\max\{h,M\}$ yields local length $[M-h]_+$ and reduces to the proposition.

\begingroup
\noindent\emph{Proof.}
For any request--action trajectory extending $\mathcal I$, set $x_q=x_q^{\mathcal I}$ for $q\leq h$ and extend the notation to $q>h$ by letting $x_q\in\{0\}\cup[n]$ be the resource successfully assigned in period $q$, with $x_q=0$ when no assignment succeeds. At the start of calendar period $q$, define the state tuple
\[
s_q\defeq(x_{q-D+1},\ldots,x_{q-1}),
\]
with $s_q=()$ when $D=1$. By Eq.~\eqref{eq:ORRA-waiting-update}, resource $i$ is unavailable in period $q$ if and only if it was successfully assigned in one of periods $q-D+1,\ldots,q-1$, which is equivalent to $i\in\operatorname{Occ}(s_q)$. In particular, at the first continuation period $q=h+1$, the tuple $s_q$ is exactly $s^{\mathcal I}$ defined above.

Fix a local continuation period $t\in[T]$, whose calendar index is $q=h+t$, and a state $s$. A nonzero action $i$ can succeed exactly when $\bar e_t(i)=1$ and $i\notin\operatorname{Occ}(s)$. Such an action earns one reward and makes the next state $\sigma(s,i)$. Rejection earns zero and makes the next state $\sigma(s,0)$. Any ineligible or unavailable nonzero attempt also earns zero and leaves availability unchanged, so replacing it by rejection produces the same next state and loses no reward. Therefore the action set in the displayed recursion for $V_t$ includes every successful assignment and an equivalent rejection action for every unsuccessful attempt, and the transition rule gives their corresponding successor states.

The terminal value $V_{T+1}(s)=0$ is correct for the empty continuation. Given the optimal values $V_{t+1}$, the recursion maximizes immediate reward plus the optimal value of the successor state. Backward induction therefore gives the continuation optimum $V_1(s^{\mathcal I})$; backtracking under the standing measurable selection rule returns an optimal action sequence.

Every state belongs to $(\{0\}\cup[n])^{D-1}$, so there are at most $(n+1)^{D-1}$ states per period. Each state considers at most $n+1$ actions. Hence the dynamic program examines $O(T(n+1)^D)$ state--action pairs. The initializer $s^{\mathcal I}$ is obtained by replaying the fixed history once and retaining only the last $D-1$ assignment outcomes. This proves the proposition.
\hfill\Halmos
\endgroup

\begin{lemma}[Clearing and copying]
\label{lem:ORRA-clearing-copy}
Fix $m\in\mathbb Z_+$, $N\in\{m,m+1,\ldots\}\cup\{\infty\}$, two histories $\mathcal I,\mathcal I'$ ending in period $m-1$, and a common continuation $\bm e_{m:N}$. Every feasible continuation after $\mathcal I'$ can be copied after $\mathcal I$ with a loss of at most $D-1$ reward.
\end{lemma}
\noindent\emph{Proof.}
Set $t_{\mathrm{reset}}\defeq m+D-1$. Take any feasible continuation after $\mathcal I'$ and replace each unsuccessful nonzero attempt by rejection. After $\mathcal I$, reject through period $t_{\mathrm{reset}}-1$ and copy these normalized actions from period $t_{\mathrm{reset}}$ onward. At the start of period $t_{\mathrm{reset}}$, every resource is free. Each resource's first copied nonzero action is therefore feasible; after that success, its return time agrees with the original trajectory, so induction preserves every later copied success. The construction discards rewards only from periods $m,\ldots,t_{\mathrm{reset}}-1$, at most $D-1$ periods. If $N<t_{\mathrm{reset}}$, the copied interval is empty; if $N=\infty$, finite support reduces the comparison to a finite endpoint.\hfill\Halmos

\subsection{Proof of Lemma~\ref{lem:ORRA-bounded-Lipschitz}}
\label{sec:proof-lem-ORRA-bounded-influence-Lipschitz}
Fix $m\in\mathbb Z_+$, $N\in\{m,m+1,\ldots\}\cup\{\infty\}$, two histories $\mathcal I,\mathcal I'$ ending in period $m-1$, and a common continuation $\bm e_{m:N}$. Applying Lemma~\ref{lem:ORRA-clearing-copy} to an optimizer after $\mathcal I'$ gives
\[
\mathrm{Opt}(\mathrm{ORRA}_n^{\mathcal I},\bm e_{m:N})
\geq
\mathrm{Opt}(\mathrm{ORRA}_n^{\mathcal I'},\bm e_{m:N})-(D-1).
\]
Swapping the histories gives the reverse inequality. Since $L=1$, this proves $(D-1)$-bounded influence.

For the strongly Lipschitz condition, fix $m,N\in\mathbb Z_+$ with $m\leq N$, a history through period $m-1$, requests $e_m,e'_m$, a common suffix through $N$, and fixed actions $\bm a_{m:N}$. Put $i_0\defeq a_m$. The two runs have the same availability state before period $m$. If $i_0=0$, their trajectories coincide. Hence assume $i_0\in[n]$. If resource $i_0$ is unavailable in the common state or $e_m(i_0)=e'_m(i_0)$, their trajectories again coincide. Otherwise, swap the requests if needed so that the first trajectory successfully assigns resource $i_0$ in period $m$ and the second does not. All other resources then have identical trajectories.

Let $S_t,S'_t$ count the successful uses of resource $i_0$ during periods $m,\ldots,t$. Let $W_t,W'_t$ be the vectors of next available periods for the two trajectories and set $w_t\defeq W_t(i_0)$ and $w'_t\defeq W'_t(i_0)$. Induction gives, after every $t\in\{m,\ldots,N\}$, either
\[
S_t=S'_t\ \text{ and }\ w_t\leq w'_t,
\qquad\text{or}\qquad
S_t=S'_t+1\ \text{ and }\ w_t\geq w'_t.
\]
The second alternative holds after period $m$. A later period in which both trajectories succeed or both fail preserves the applicable alternative. A one-sided success switches the alternatives: its new threshold is $t+D$, whereas the other active threshold was set no later than $t-1$ and is at most $t+D-1$. Hence $S_N-S'_N\in\{0,1\}$. Rewards from every other resource coincide, so the cumulative rewards differ by at most one. Swapping $e_m,e'_m$ gives the absolute-value bound; with $L=1$ and $\delta_{\mathrm{req}}(e,e')=\mathbb I(e\neq e')$, this proves ORRA is $(1,1)$-strongly Lipschitz. \hfill\Halmos

\subsection{Proof of Lemma~\ref{lem:ORRA-online-oracle}}
\label{sec:proof-lem-ORRA-online-oracle}

Let $\eta=\eta^{\mathrm{ORRA}}=0.589$. Theorem~1 of \citet{delong2024online} implies that, from the all-free state, fresh PRR earns expected reward at least $\eta$ times the finite-horizon offline optimum: its guarantee is against an LP upper bound and hence against our offline optimum. Unsuccessful attempts in our formulation can be replaced by rejection without changing reward or state. Thus the models have the same optimum, and the base guarantee proves the initial-policy clause of Definition~\ref{def:eta-online-oracle}. For $h=0$, the restarted policy equals the initial policy, so this guarantee also implies the restarted clause.

Now fix $h\geq1$, a history $\mathcal I=(\bm e_{1:h},\bm a_{1:h})$, and a finite endpoint $N\geq h+1$, and put $t_{\mathrm{reset}}\defeq h+D$. If $N<t_{\mathrm{reset}}$, then $\bm\pi_h^{\mathcal I}$ earns zero and the continuation optimum is at most $N-h\leq D-1$, so the required additive bound holds. If $N\geq t_{\mathrm{reset}}$, PRR$^*$ clears the inherited state and starts fresh PRR at $t_{\mathrm{reset}}$. Let $\mathrm{Opt}_{\mathrm{free}}(\bm e_{t_{\mathrm{reset}}:N})$ denote the offline optimum from the all-free state at $t_{\mathrm{reset}}$. Discarding the first $D-1$ periods of an optimal continuation loses at most $D-1$ reward, and the copying construction in Lemma~\ref{lem:ORRA-clearing-copy} preserves every remaining success when the normalized suffix is run from the all-free state. Thus $\mathrm{Opt}_{\mathrm{free}}(\bm e_{t_{\mathrm{reset}}:N})\geq \mathrm{Opt}(\mathrm{ORRA}_n^{\mathcal I},\bm e_{h+1:N})-(D-1)$, and the base guarantee gives
\[
\begin{aligned}
\mathrm{Val}(\mathrm{ORRA}_n^{\mathcal I},\bm e_{h+1:N},\bm\pi_h^{\mathcal I}) 
&\geq\eta\,\mathrm{Opt}_{\mathrm{free}}(\bm e_{t_{\mathrm{reset}}:N})\\
&\geq\eta\bigl(\mathrm{Opt}(\mathrm{ORRA}_n^{\mathcal I},\bm e_{h+1:N})-(D-1)\bigr)\\
&\geq\eta\,\mathrm{Opt}(\mathrm{ORRA}_n^{\mathcal I},\bm e_{h+1:N})-(D-1).
\end{aligned}
\]
The last inequality uses $\eta\leq1$. Since $L=1$ and $f=D-1$, this is the additive loss permitted by Definition~\ref{def:eta-online-oracle}. For an infinite endpoint, finite support makes both continuation values equal their truncations through the effective horizon; if that horizon is at most $h$, both continuation values are zero. Hence the same conclusion holds for infinite endpoints, and PRR$^*$ is an $\eta^{\mathrm{ORRA}}$-online oracle.\hfill\Halmos

\section{Comparison for Caching under a Common Request Prediction}
\label{sec:ec-related-comparison}
\label{sec:ec-request-advice}

\noindent\underline{Shared scenario.}
We compare exact-oracle AdaSwitch, Blum--Burch, and DART on the same caching problem, with cache capacity $k\geq1$, a common initial cache, realized requests $\bm e_{1:\infty}$, and a fixed prediction $\bm e^*_{1:\infty}$. Both request sequences have finite support. Each method is evaluated by its expected total number of cache misses.

All three methods use fresh-start Marking as the baseline. Each combiner initializes separate simulated copies of Marking and the advice policy from the common initial cache and runs them continuously. AdaSwitch initializes a fresh copy of Marking from its actual cache whenever it returns to conservative mode.

Predictive planning in AdaSwitch and in the combiners' advice policy (constructed below) uses the same exact fixed-history offline oracle and effective-horizon estimator. Write $\mathrm{Opt}>0$ for the offline optimum on the realized requests and $C_B$ for the expected cost of running Marking without switching. Lemma~\ref{lem:strict-marking-bound} gives $C_B\leq\eta\mathrm{Opt}$ with $\eta=2(\ln k+1)$.

\noindent\underline{From page loads to cache misses.}
As in Section~\ref{sec:proof-marking-strict-online-oracle}, we represent empty slots and redundant copies by distinct dummy pages absent from both request sequences. We apply the Blum--Burch and DART guarantees using page-load costs: moving between $k$-page cache states $S,T$ costs $|T\setminus S|\leq k$, giving the additive terms involving $k$ below. Page loads equal cache misses for Marking and the advice policy, but a combiner may also load pages to match a selected policy's simulated cache.

To transfer the page-load guarantees to cache misses, we simulate each combiner using a virtual cache and maintain a separate actual cache that serves requests. Both caches start with the common initial contents. The virtual cache represents the combiner's own state; its component policies continue to maintain their separate simulated caches.

After the combiner serves a request in the virtual cache, we leave the actual cache unchanged if it already contains the requested page. Otherwise, we load that page, first evicting a page absent from the virtual cache if the actual cache is full. Such a page exists because both caches have capacity $k$, and the virtual cache contains the requested page that the full actual cache lacks.

The actual cache incurs no more misses than the number of page loads into the virtual cache~\citep[Section~5]{christianson2023optimal}. To see this, count the real pages present in the virtual cache but absent from the actual cache. This count starts at zero, increases by at most one per page load into the virtual cache, and decreases by one per actual miss. Since the count remains nonnegative, total actual misses cannot exceed total page loads into the virtual cache. Thus, the bounds below apply to cache misses for both the converted combiners and AdaSwitch.

\noindent\underline{Advice policy.}
Each period, the advice policy for Blum--Burch and DART calls the shared exact offline oracle to compute an optimal plan from its own current cache, using the observed request and the original predictions for later periods. It executes only the first action and replans in the next period. The shared oracle selects optimal caching plans that fill empty slots before eviction.

Let $M=M(\bm e_{1:\infty})$ be the realized effective horizon and put $\bar k=\max\{k,2\}$. Write $\varphi_M^*=\sum_{t=1}^{M}\mathbb I(e_t\ne e_t^*)$ for the number of request mismatches through $M$, including those involving the no-request symbol $\bot$. The expected total cost $C_P$ of running this advice policy alone satisfies
$
C_P\leq\mathrm{Opt}+4\varphi_M^*+\bar k.
$
This follows from Lemma~\ref{lem:cost-predictive-follow-finite} with $m=0$, $n=M$, $L=1$, $b=2$, and $c=\bar k$. Under perfect predictions, each reoptimization uses the true remaining requests, so $C_P=\mathrm{Opt}$ by Observation~\ref{obs:sequential-exact-optimality}.

\noindent\underline{Resulting guarantees.}
Each method satisfies both cost bounds in its row of Table~\ref{tab:switching-frameworks}, so its expected total cost is bounded by their minimum. The parameter $0<\delta<1/4$ controls the combiners' multiplicative overhead relative to advice, and $\epsilon>0$ is AdaSwitch's robustness slack. The AdaSwitch bounds use
$
A_\epsilon=2+(\eta-1)\left(\frac{10\eta}{\epsilon}+11\right), 
B_\epsilon=8+\frac{10(\eta-1)}{\epsilon}.
$
Here $A_\epsilon\bar k$ is the additive term independent of prediction error, and $B_\epsilon$ is the coefficient of the request-mismatch count. For DART, Appendix~C.1 of \citet{christianson2023optimal} yields the explicit robustness coefficient $R_\delta\defeq4+(1+8/\delta)\exp(1+4/\delta)$.

\begin{table}[h]
\centering
\caption{Expected cost bounds in the shared caching scenario.}
\label{tab:switching-frameworks}
\small
\setlength{\tabcolsep}{4pt}
\renewcommand{\arraystretch}{1.25}
\begin{tabular}{@{}
>{\raggedright\arraybackslash}m{3.4cm}
>{\centering\arraybackslash}m{6.9cm}
>{\centering\arraybackslash}m{5.1cm}
@{}}
\hline
Method & Worst-case bound & Prediction-dependent bound \\
\hline
Blum--Burch & $(1+\delta)C_B+\mathcal O(k/\delta)$ & $(1+\delta)C_P+\mathcal O(k/\delta)$ \\
\hline
DART & $\begin{gathered}\min\bigl\{R_\delta C_B,\\ (4+4/\delta)C_B+(4/\delta+2)k\bigr\}\end{gathered}$ & $(1+\delta)C_P$ \\
\hline
Exact-oracle AdaSwitch & $(\eta+\epsilon)\mathrm{Opt}$ & $\mathrm{Opt}+A_\epsilon\bar k+B_\epsilon\varphi_M^*$ \\
\hline
\end{tabular}
\par\smallskip
\begin{minipage}{\linewidth}
\footnotesize
Sources: Blum--Burch, \citet[Theorem~2.2]{wei2020better} and \citet[Theorem~19]{antoniadis2023online}; DART, \citet[Theorems~3.1 and~5.1; Appendices~C.1 and~E.1]{christianson2023optimal}; AdaSwitch, Theorem~\ref{thm:cache-adaswitch-ca}. Additive terms are independent of the request sequence length.
\end{minipage}
\end{table}

\noindent\underline{Comparison.}
Substituting the advice bound into Table~\ref{tab:switching-frameworks} gives expected cost at most $(1+\delta)(\mathrm{Opt}+4\varphi_M^*+\bar k)$ for DART, with an additional $\mathcal O(k/\delta)$ term for Blum--Burch. Thus, all three prediction-dependent bounds can be expressed in terms of the same request-mismatch count \(\varphi_M^*\). We next compare their robustness and perfect-prediction guarantees.

\emph{Robustness guarantees.} With $\epsilon=\eta\delta$, AdaSwitch guarantees expected cost at most $(1+\delta)\eta\mathrm{Opt}$. After substituting $C_B\leq\eta\mathrm{Opt}$, the cited Blum--Burch bound has the same leading factor and an additional $\mathcal O(k/\delta)$ term. Under the same substitution, even the smaller leading robustness factor in the cited DART bounds is $4/\delta>16$ times AdaSwitch's for $0<\delta<1/4$.

\emph{Perfect-prediction guarantees.} Fix $k$ and $\delta$, keeping $\epsilon=\eta\delta$. Under perfect predictions, $C_P=\mathrm{Opt}$, so DART guarantees a cost ratio of at most $1+\delta$ on every instance. AdaSwitch retains robustness factor $(1+\delta)\eta$, while its cost-ratio bound $1+A_\epsilon\bar k/\mathrm{Opt}$ approaches $1$ as $\mathrm{Opt}$ grows. This bound is below DART's displayed bound $1+\delta$ when $\mathrm{Opt}>A_\epsilon\bar k/\delta$.

\section{Additional OLTQwP and CAwP Experiments}
\label{sec:additional-experiments}

This section records the full protocols for the main-text experiments and supplements them with two arrival-horizon studies and a burst-and-gap demand study for AdaSwitch-OLTQ, and a prediction-error stress test for AdaSwitch-CA.

\subsection{OLTQwP: Experiment Designs and Results}
\label{sec:additional-oltq-experiments}

\noindent\underline{Common setup and uncertainty.}
All four OLTQwP designs below compare AdaSwitch-OLTQ with Q-FRACwP and the prediction-free Q-FRAC and Immediate service benchmarks.

Both the realized and predicted sequences are zero after the arrival horizon $T$, and all policies and the exact optimum are evaluated through $T+\ell-1$. For robustness target $r$, Q-FRACwP receives $r$ and AdaSwitch-OLTQ uses $\epsilon_r=\eta^{\mathrm{OLTQ}}-r$. AdaSwitch's reward threshold for entering predictive mode is $10\ell^2/\epsilon_r$, and its prediction-error threshold for returning to conservative mode is $2\ell/\eta^{\mathrm{OLTQ}}$.

Immediate service earns $\ell\sum_{t=1}^{T}\mathbb I(e_t>0)$ by serving one newly arriving request immediately in each nonempty period and rejecting all other copies. Its worst-case factor on the full OLTQ model is $2/(\ell+1)$. Indeed, each nonempty arrival cohort contributes at most $\ell(\ell+1)/2$ to the offline optimum, and an isolated $\ell$-request burst attains this ratio. Thus, Immediate service need not meet the robustness target $r$.

Each parameter setting uses 200 independent trials. Within a trial, all policies share the realized instance, and the predictive policies share its prediction. The robustness sweep also evaluates all target values on the same instance. Rewards are divided instancewise by the exact realized-sequence optimum before averaging across trials; uncertainty is reported using pointwise two-sided $95\%$ Student $t$ confidence intervals.

\noindent\underline{Independent demand and robustness sweep.}
The robustness sweep and two arrival-horizon studies use a zero-inclusive capped geometric model, motivated by the experiments of \citet{huo2026online}. Independently draw $G_1,\ldots,G_T$ with $\Pr(G_t=g)=p(1-p)^g$ for $g\in\mathbb Z_{\geq0}$, set $e_t=\min\{G_t,\ell\}$, and fix $p=0.1$.

The main-text robustness sweep uses $\ell\in\{20,30,40\}$, $T=15000$, and perfect predictions. AdaSwitch-OLTQ and Q-FRACwP share the guaranteed robustness grid $\mathcal R=\{0.050,0.075,\ldots,0.600\}$. Figure~\ref{fig:empirical-consistency-different-robustness} reports the results.

\noindent\underline{Arrival-horizon designs.}
The two studies use the independent-demand model above with $\ell=20$, $r=0.425$, and $T\in\{2000,2500,\ldots,20000\}$. They use separate trial sets. Within each study and trial, all horizons use prefixes of the same arrival sequence of length $20000$. The perfect-prediction study sets $e_t^*=e_t$ for every $t\leq T$.

The fixed-budget study holds expected total absolute prediction error at $B=200$. Starting from $e_t^*=e_t$ for every $t\leq T$, independently corrupt each predicted count with probability $q_T=B/T$. A corrupted count moves to an adjacent count in $\{0,\ldots,\ell\}$, choosing either neighbor equiprobably in the interior and moving inward at a boundary. Within each trial, we reuse the random draws for corruption decisions and directions across horizons, with each horizon using its own corruption probability $q_T$.

\begin{figure}[htbp]
    \centering
    \includegraphics[width=0.60\textwidth]{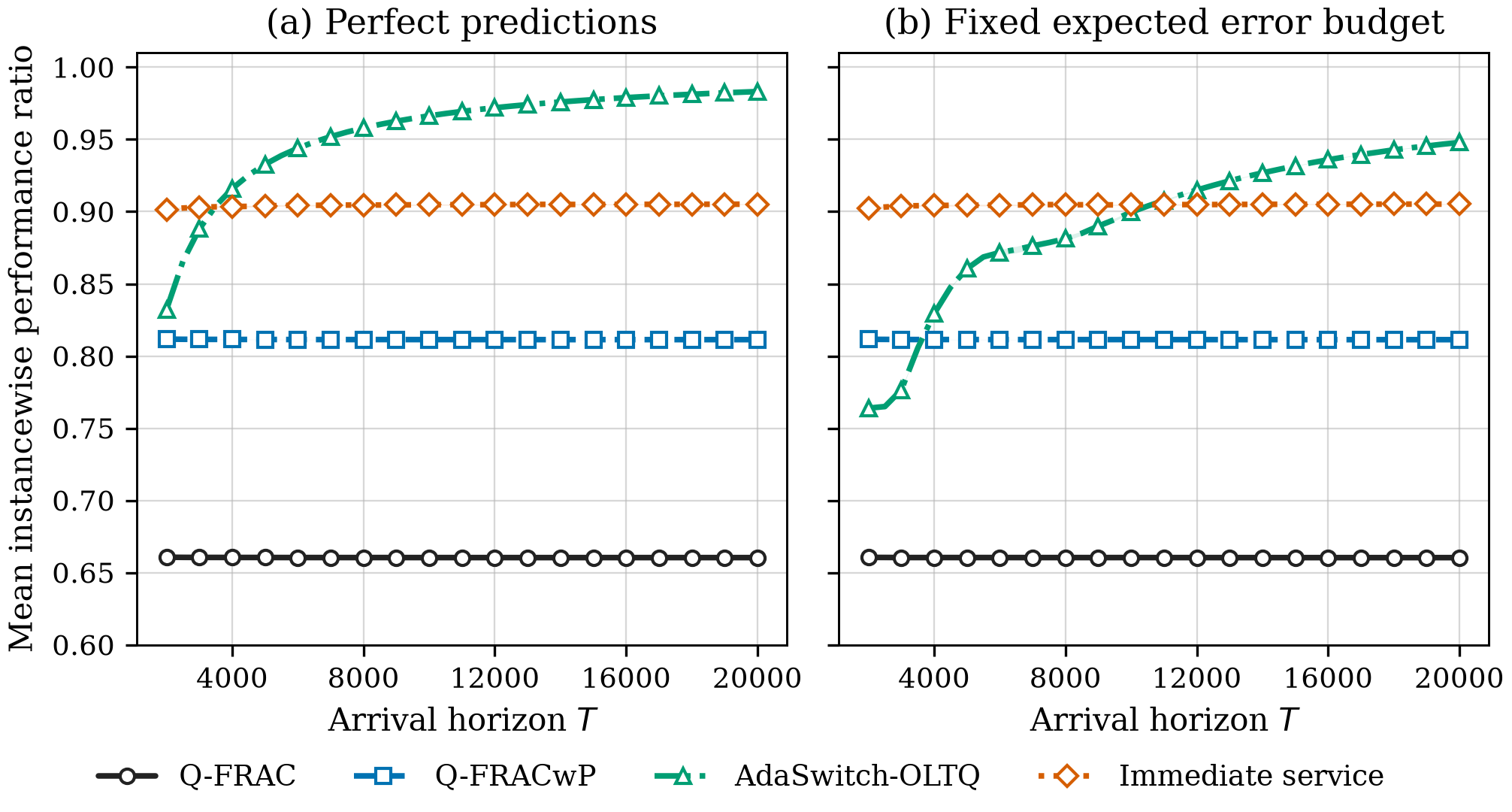}
    \caption{OLTQwP performance versus arrival horizon $T$. Both panels use $p=0.1$, $\ell=20$, and $r=0.425$. Higher ratios are better.}
    \label{fig:oltq-arrival-horizon}
\end{figure}

\noindent\underline{Arrival-horizon results.} Under perfect predictions in Figure~\ref{fig:oltq-arrival-horizon}(a), AdaSwitch-OLTQ's mean reward ratio increases toward $1$ as $T$ grows, while the comparators remain essentially flat. With a fixed expected error budget in panel~(b), AdaSwitch-OLTQ initially trails Immediate service but exceeds it in mean at all tested $T\geq11000$, reaching $0.948$ versus $0.905$ at $T=20000$. Its mean reward ratio also exceeds that of Q-FRACwP at longer tested horizons.

The recorded mode-switching histories help explain these patterns: initial conservative operation delays predictive gains, and prediction errors can trigger additional conservative epochs. In the fixed-budget study, as $T$ increases, the fraction of processing periods in predictive mode rises and the corruption probability $q_T=B/T$ falls.

\noindent\underline{Burst-and-gap design.}
We next consider short arrival bursts separated by gaps of random duration to examine how policies serve previously accepted requests during periods with no new arrivals. Fix $p=0.1$, $\ell=20$, $r=0.425$, and $T=20000$. Each demand cycle consists of an active burst lasting $U_j$ periods, drawn uniformly from $\{1,2,3\}$, followed by a gap of $L_j$ periods with no arrivals, where
$
\Pr(L_j=k)=\frac1g\left(1-\frac1g\right)^{k-1}, k\geq1, g\in\{5,6,\ldots,20\}.
$
Thus $g$ is the mean gap length. To start the demand process in stationarity, sample the initial burst--gap pair from the product of the distributions above, weighting each pair by its total duration. Then choose the position of period $1$ uniformly within that cycle. Subsequent cycles use independent burst and gap durations drawn from the distributions above. The queue starts empty.

Set $J_t=1$ during bursts and $J_t=0$ during gaps. Independently of burst and gap durations, draw potential counts $X_t=\min\{H_t,\ell\}$ in every period, where the $H_t$ are independent and $\Pr(H_t=h)=p(1-p)^{h-1}$ for $h\geq1$. Actual arrivals are $e_t=J_tX_t$; potential counts drawn during gap periods are used only to construct timing predictions below. Since bursts last two periods on average, the stationary probability of no arrivals is $g/(g+2)$. Thus, increasing $g$ both lengthens gaps and reduces the mean arrival rate.

\noindent\underline{Burst-and-gap predictions.}
We use four synthetic ex ante prediction models: perfect predictions, count errors, and two timing-error models with $d=2$ and $d=5$. Count errors use the corruption rule above with fixed probability $q=0.05$ per period.

For timing errors, consider the maximal runs of periods with $J_t=1$ within $1,\ldots,T$. Independently assign each run a forward or backward direction with equal probability, reusing the same directions for $d=2$ and $d=5$. Shift each run by $d$ periods in its assigned direction and clip it to $\{1,\ldots,T\}$. Set $\widehat J_t=1$ on the union of these shifted runs and $0$ elsewhere. Keep the potential counts $X_t$ fixed at their original periods and set $e_t^*=\widehat J_tX_t$. Counts agree wherever actual and predicted activity overlap, while predicted gaps can appear, disappear, or change length.

In all four models, predictive policies receive only the resulting predicted sequence as forecast input. Total absolute error is not held fixed across models.

\begin{figure}[htbp]
    \centering
    \includegraphics[width=0.6\textwidth]{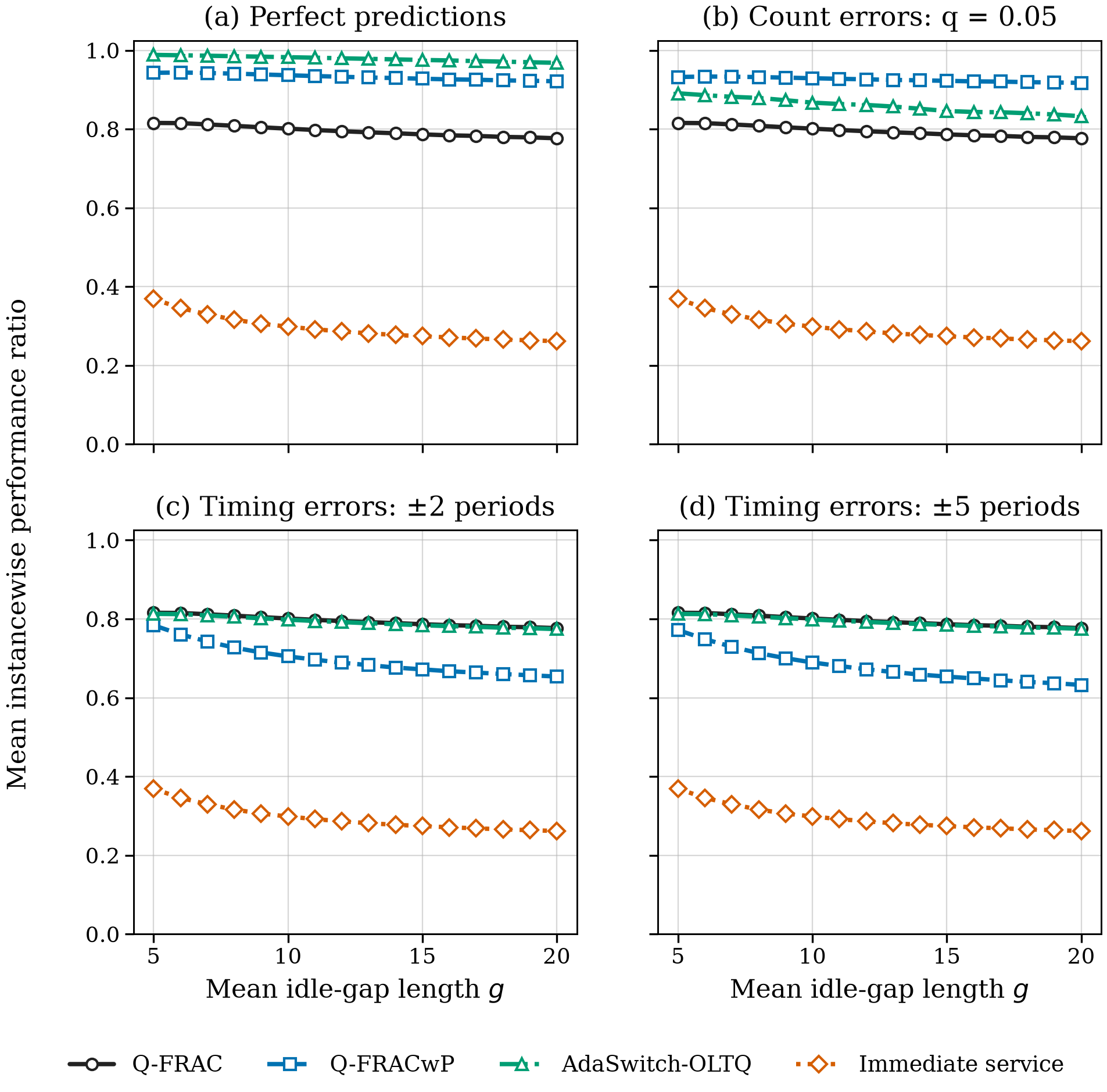}
    \caption{OLTQwP performance under short bursts and random gaps. All panels use $T=20000$, $p=0.1$, $\ell=20$, and $r=0.425$. Higher ratios are better.}
    \label{fig:oltq-burst-gap-forecasts}
\end{figure}

\noindent\underline{Burst-and-gap results.}
Under perfect predictions in Figure~\ref{fig:oltq-burst-gap-forecasts}(a), AdaSwitch-OLTQ's mean reward ratio is close to $1$ and highest at every tested $g$, with a substantial advantage over Immediate service. Reward diagnostics show that AdaSwitch-OLTQ's reward during zero-arrival periods within the arrival window is close to that of the offline schedule, indicating effective use of idle capacity. The recorded switching histories show predominantly predictive operation.

With count errors in panel~(b), Q-FRACwP has a higher mean reward ratio than AdaSwitch-OLTQ throughout the tested gap range. Under timing errors in panels~(c,d), AdaSwitch-OLTQ's mean reward ratio remains close to that of Q-FRAC and above that of Q-FRACwP. The recorded switching histories in these runs show short predictive epochs and predominantly conservative operation.

\subsection{CAwP: Experiment Design and Prediction-Error Stress Tests}
\label{sec:additional-caching-experiments}

\noindent\underline{Comparators.}
AdaSwitch-CA is compared with Marking~\citep{fiat1991competitive}, Predictive Marker~\citep{lykouris2021competitive}, and Predicted Belady. Marking is the prediction-free baseline. All predictive policies use the same predicted sequence. For Predictive Marker, we convert it into next-arrival forecasts: on arrival of actual page $e_t$, the policy stores $\min\{s>t:e_s^*=e_t\}$, or $\infty$ if that page is absent from the predicted suffix. On an actual miss with a full cache, Predicted Belady recomputes each resident page's next predicted occurrence strictly after the current period and evicts the page with the latest occurrence, treating absence as $\infty$ and breaking ties by least recent use. It is optimal under perfect predictions~\citep{belady1966study,sleator1985amortized}.

\noindent\underline{Request and prediction models.} All realized and predicted sequences are padded by $\bot$ after period $T$. All request models use a common $100$-page universe. In the uniform model, realized requests are drawn independently and uniformly from this universe. In each Zipf model, realized requests are drawn independently from the same universe, with the probability of rank $j$ proportional to $j^{-0.8}$ or $j^{-1.1}$. The fourth request model is a fixed cycle over a subset of $k+1$ pages.

Predictions are generated by one of six models: perfect prediction; independent per-entry corruption with probability $10\%$; independent per-entry corruption with probability $30\%$; a $50$-period delay; reversal within consecutive $100$-position blocks; or a cycle over $20$ disjoint prediction-only pages. In a corrupted prediction, each selected entry is replaced uniformly by one of the other $99$ pages in the common universe. The per-entry corruption probability governs request replacements; it does not quantify the resulting next-arrival errors. In the delayed prediction, $e_t^*=e_1$ for $t\leq50$ and $e_t^*=e_{t-50}$ for $51\leq t\leq T$. In the disjoint-cycle prediction, predictions cycle over $20$ pages outside the common $100$-page universe.

\noindent\underline{Implementation and uncertainty.} All methods and the exact optimum start with $k$ empty slots. AdaSwitch-CA uses the fresh-start Marking oracle of Lemma~\ref{lem:strict-marking-bound}, with $b=2$, $c=\bar{k}\defeq\max\{k,2\}$, and, outside the robustness sweep, $\epsilon=2(\ln k+1)$.

Each experiment uses 200 independent trials. Within each scenario of a trial, all methods are evaluated on the same instance. For each instance, we divide each method's number of cache misses by the exact offline optimum. When a figure combines multiple scenarios, we average these ratios within each trial. We report means across trials, together with nominal pointwise two-sided $95\%$ confidence intervals based on the Student $t$ distribution.

\noindent\underline{Experiment grids.} The main-text robustness sweep (Figure~\ref{fig:cache-consistency-robustness}) uses perfect predictions, fixes $k=10$, and varies $\epsilon\in\{0.5,1.25,\ldots,8\}$. Within each trial, we average ratios over the four combinations of $T\in\{5000,10000\}$ and two request models: uniform and Zipf with exponent $0.8$. The horizontal coordinate is AdaSwitch-CA's guaranteed robustness upper bound $2(\ln k+1)+\epsilon$; the comparator policies are independent of $\epsilon$.

The main-text horizon study (Figure~\ref{fig:cache-opt-scale}) uses all combinations of $M=T\in\{500,1450,\ldots,10000\}$, $k\in\{5,10,20\}$, perfect and $10\%$-corrupted predictions, and the same two request models. Each point averages the ratios from the two request models within a trial.

\begin{figure}[htbp]
    \centering
    \includegraphics[width=0.5\textwidth]{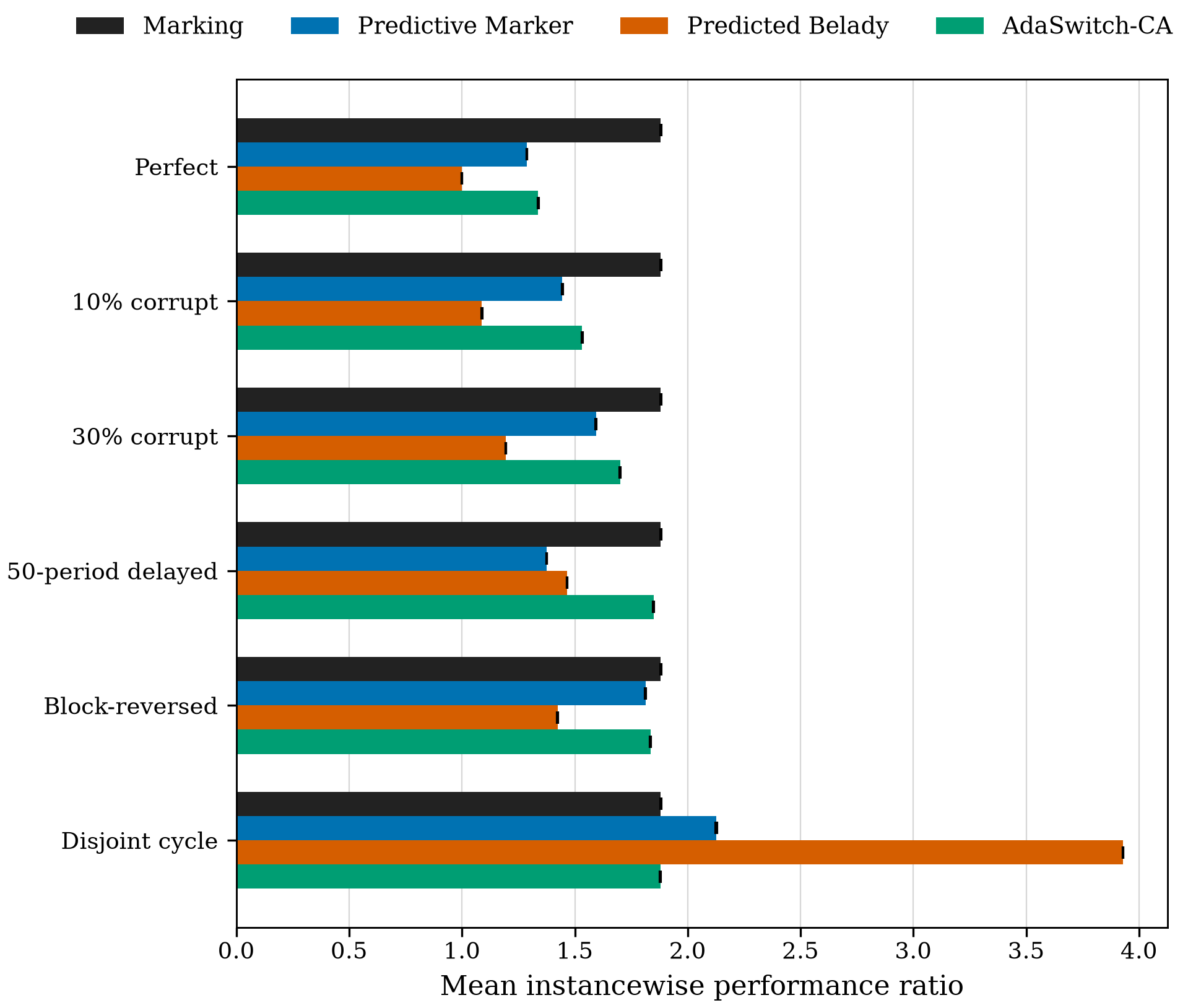}
    \caption{CAwP performance under six prediction models. Lower ratios are better.}
    \label{fig:cache-prediction-errors}
\end{figure}

The prediction-error stress test (Figure~\ref{fig:cache-prediction-errors}) fixes $T=5000$ and uses all six prediction models. For each prediction model, we average ratios within each trial over the twelve combinations of four request models and $k\in\{5,10,20\}$.

\noindent\underline{Results.} In the prediction-error stress test shown in Figure~\ref{fig:cache-prediction-errors}, Predictive Marker has a lower mean cost ratio than AdaSwitch-CA under the delayed prediction. Under the disjoint-cycle prediction, AdaSwitch-CA and Marking have similar mean cost ratios, both below those of Predictive Marker and Predicted Belady.

\end{document}